%% file: hcdbnqjtcjmvfxrwcqhvxzfkhynqrfnr 2/body.tex
\title[Analogue film artefact removal]%
      {Simulating analogue film damage to analyse and improve\\artefact restoration on high-resolution scans}

\author[D. Ivanova \& J. Williamson \& P. Henderson]
{\parbox{\textwidth}{\centering D. Ivanova$^{1}$\orcid{0000-0002-3710-7413}
        and J. Williamson$^{1}$\orcid{0000-0001-8085-7853} 
        and P. Henderson$^{1}$\orcid{0000-0002-5198-7445}
        }
        \\
{\parbox{\textwidth}{\centering $^1$University of Glasgow, UK\\
      }
}
}

\begin{document}


\maketitle
\begin{abstract}
Digital scans of analogue photographic film typically contain artefacts such as dust and scratches.
Automated removal of these is an important part of preservation and dissemination of photographs of historical and cultural importance.
While state-of-the-art deep learning models have shown impressive results in general image inpainting and denoising, film artefact removal is an understudied problem.
It has particularly challenging requirements, due to the complex nature of analogue damage, the high resolution of film scans, and potential ambiguities in the restoration.
There are no publicly available high-quality datasets of real-world analogue film damage for training and evaluation, making quantitative studies impossible.
We address the lack of ground-truth data for evaluation by collecting a dataset of 4K damaged analogue film scans paired with manually-restored versions produced by a human expert, allowing quantitative evaluation of restoration performance. We have made the dataset available at \URL{https://doi.org/10.6084/m9.figshare.21803304}.
We construct a larger synthetic dataset of damaged images with paired clean versions using a statistical model of artefact shape and occurrence learnt from real, heavily-damaged images.
We carefully validate the realism of the simulated damage via a human perceptual study, showing that even expert users find our synthetic damage indistinguishable from real. In addition, we demonstrate that training with our synthetically damaged dataset leads to improved artefact segmentation performance when compared to previously proposed synthetic analogue damage overlays. The synthetically damaged dataset can be found at \URL{https://doi.org/10.6084/m9.figshare.21815844}, and the annotated authentic artefacts along with the resulting statistical damage model at \URL{https://github.com/daniela997/FilmDamageSimulator}.
Finally, we use these datasets to train and analyse the performance of eight state-of-the-art image restoration methods on high-resolution scans. We compare both methods which directly perform the restoration task on scans with artefacts, and methods which require a damage mask to be provided for the inpainting of artefacts. We modify the methods to process the inputs in a patch-wise fashion to operate on original high resolution film scans.


\end{abstract}  

\begin{figure*}
  \begin{subfigure}[t]{.325\textwidth}
    \centering
        \begin{tikzpicture}[spy using outlines={circle,blue,magnification=3,size=1.8cm, connect spies}]
        \node {\includegraphics[width=\linewidth]{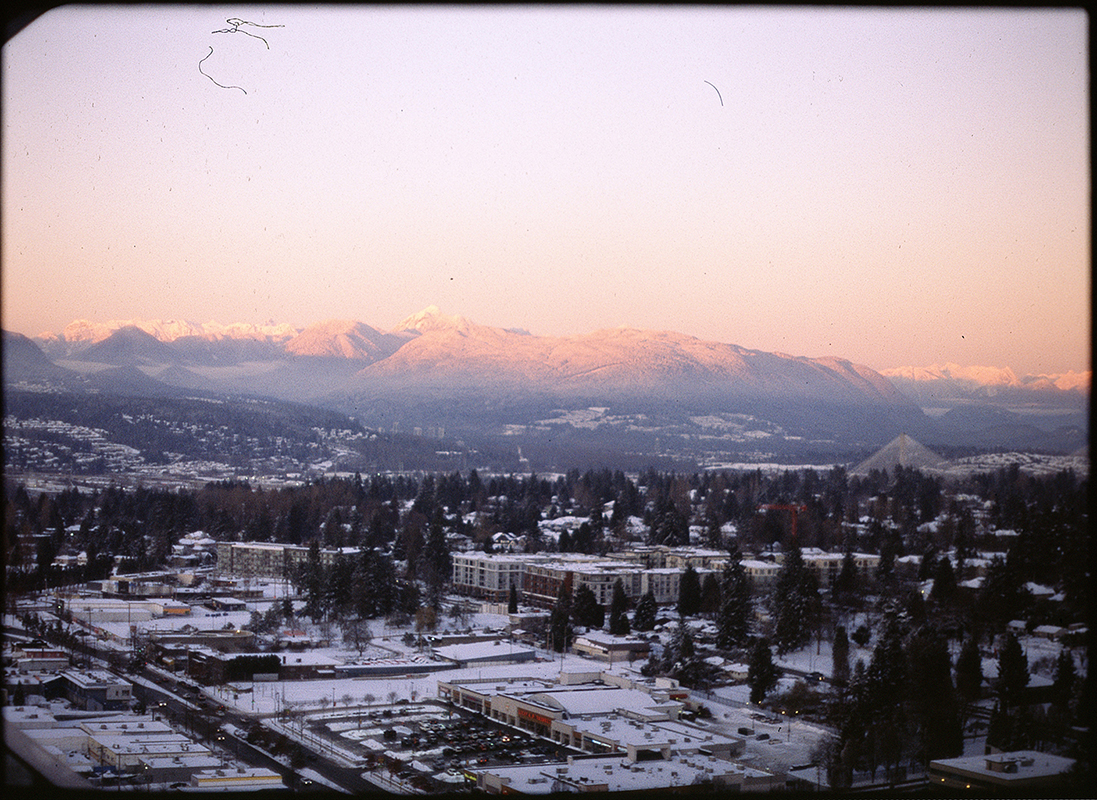}};
        \spy on (-1.63,1.75) in node [left] at (1.8,1.20);
        \end{tikzpicture}
    \caption{\textbf{Input}: 4K film scan with authentic damage (hairs and dirt)}
  \end{subfigure}
    \hfill
  \begin{subfigure}[t]{.325\textwidth}
    \centering
        \begin{tikzpicture}[spy using outlines={circle,yellow,magnification=3,size=1.8cm, connect spies}]
        \node {\includegraphics[width=\linewidth]{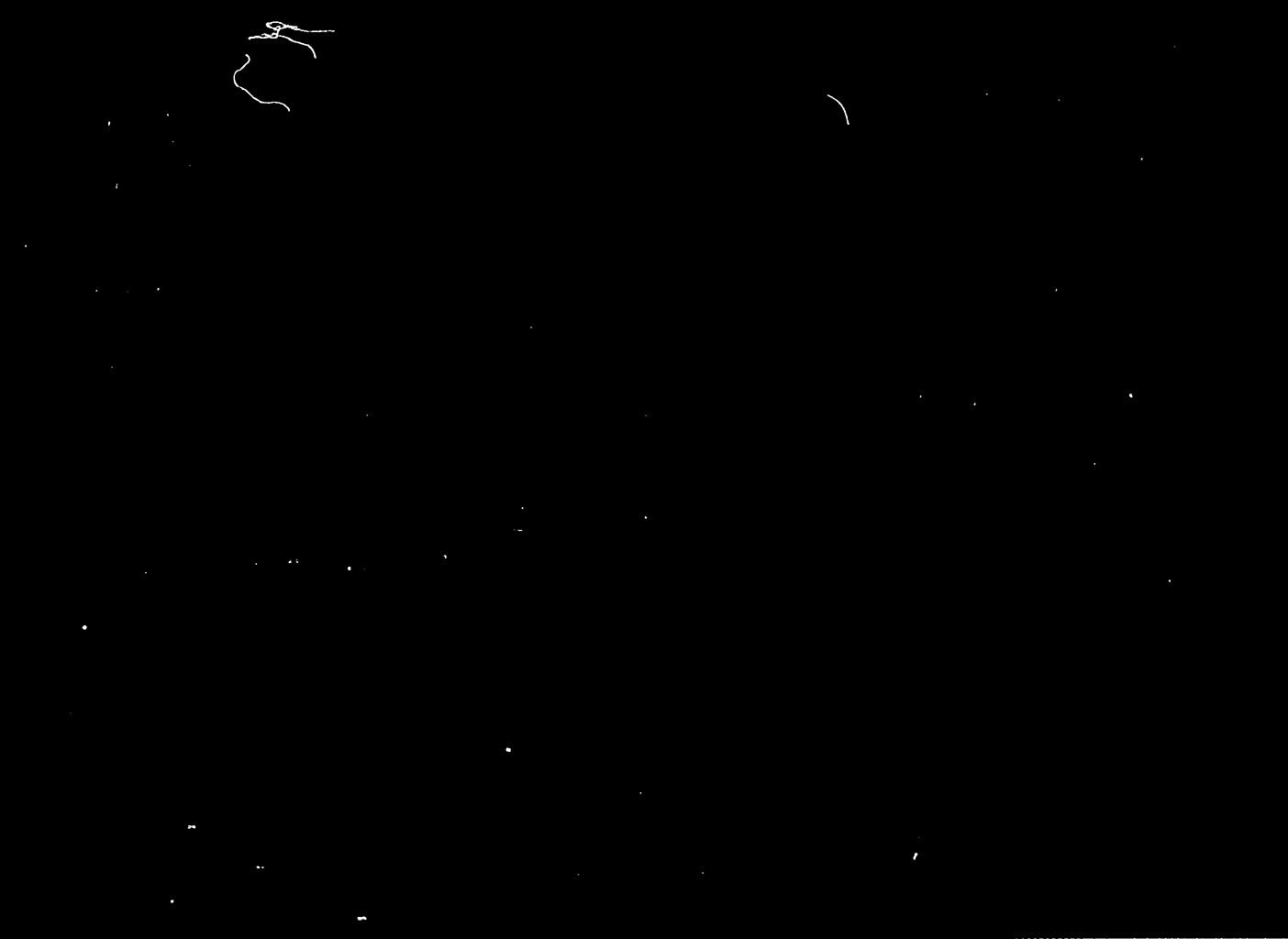}};
        \spy on (-1.63,1.75) in node [left] at (1.8,1.20);
        \end{tikzpicture}
    \caption{\textbf{Artefact Segmentation}: prediction from our U-Net trained on synthetically damaged data.}
  \end{subfigure}
    \hfill
  \begin{subfigure}[t]{.325\textwidth}
    \centering
        \begin{tikzpicture}[spy using outlines={circle,blue,magnification=3,size=1.8cm, connect spies}]
        \node {\includegraphics[width=\linewidth]{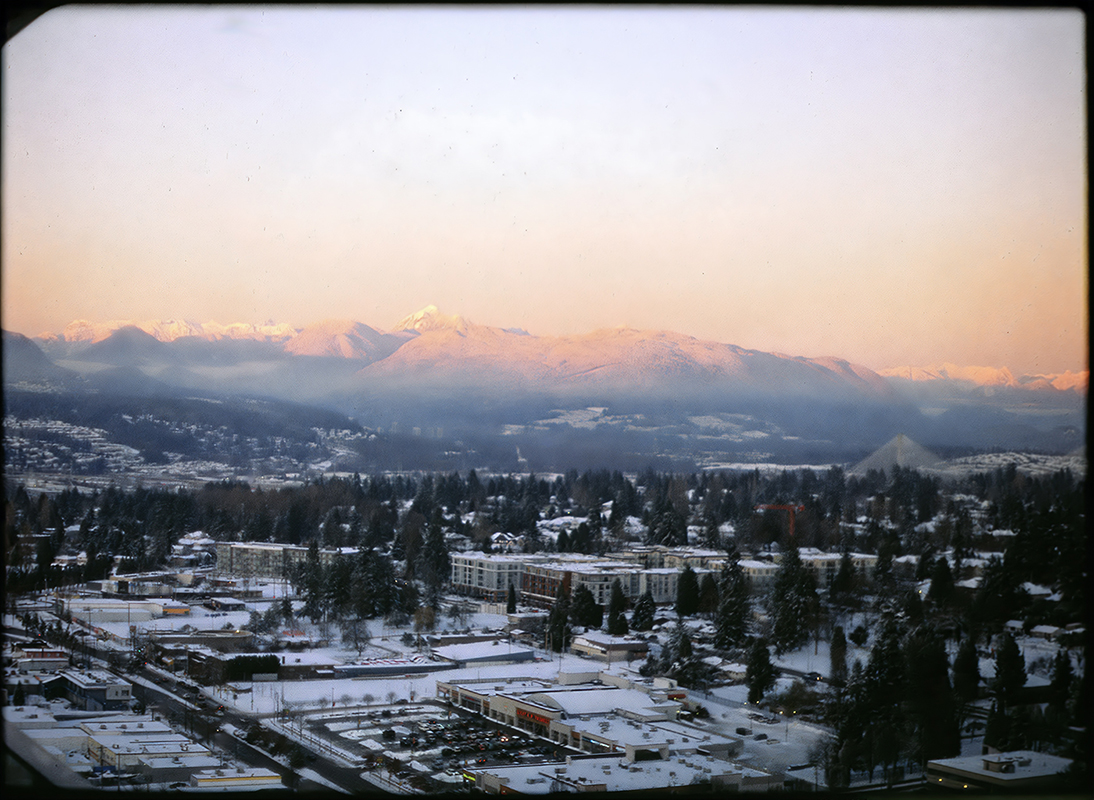}};
        \spy on (-1.63,1.75) in node [left] at (1.8,1.20);
        \end{tikzpicture}
    \caption{\textbf{Restoration by BOPB~\cite{wan2020bringing}}: using our predicted segmentation.}
  \end{subfigure}
  
   \medskip
  
  \begin{subfigure}[t]{.325\textwidth}
    \centering
        \begin{tikzpicture}[spy using outlines={circle,blue,magnification=3,size=1.8cm, connect spies}]
        \node {\includegraphics[width=\linewidth]{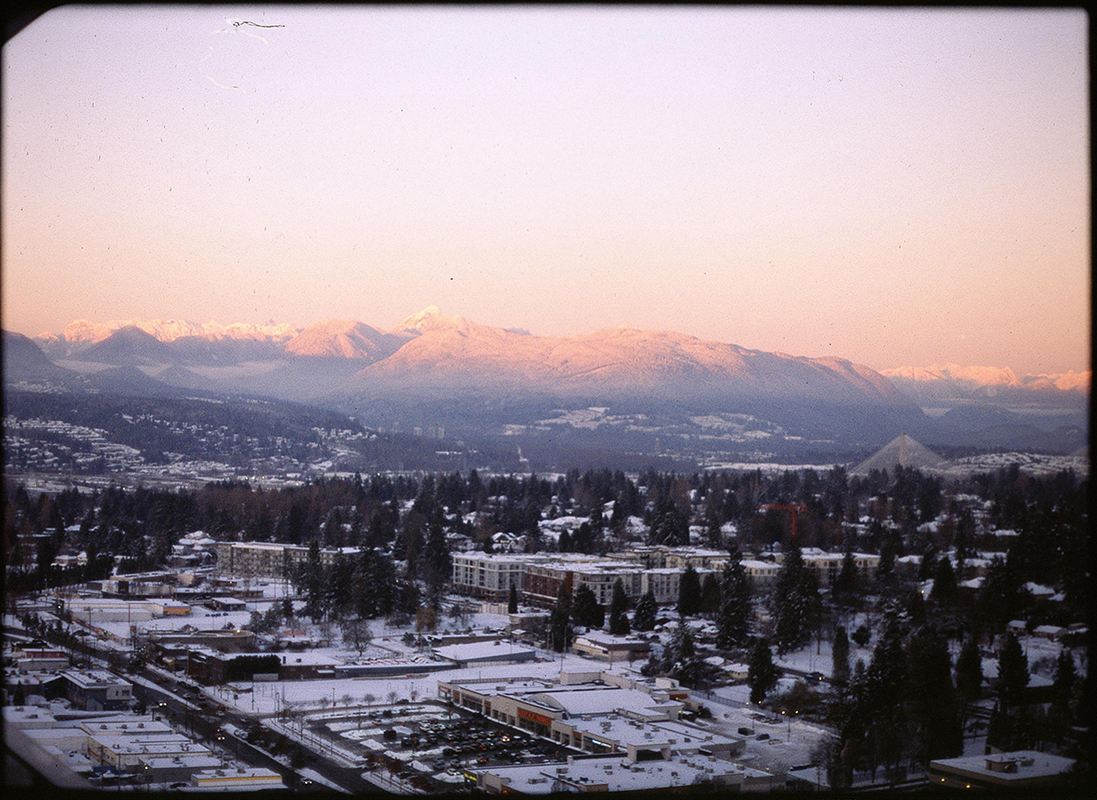}};
        \spy on (-1.63,1.75) in node [left] at (1.8,1.20);
        \end{tikzpicture}
    \caption{\textbf{Restoration by U-Net~\cite{visapp22}}: retrained on our synthetic damage.}
  \end{subfigure}
    \hfill
  \begin{subfigure}[t]{.325\textwidth}
    \centering
        \begin{tikzpicture}[spy using outlines={circle,blue,magnification=3,size=1.8cm, connect spies}]
        \node {\includegraphics[width=\linewidth]{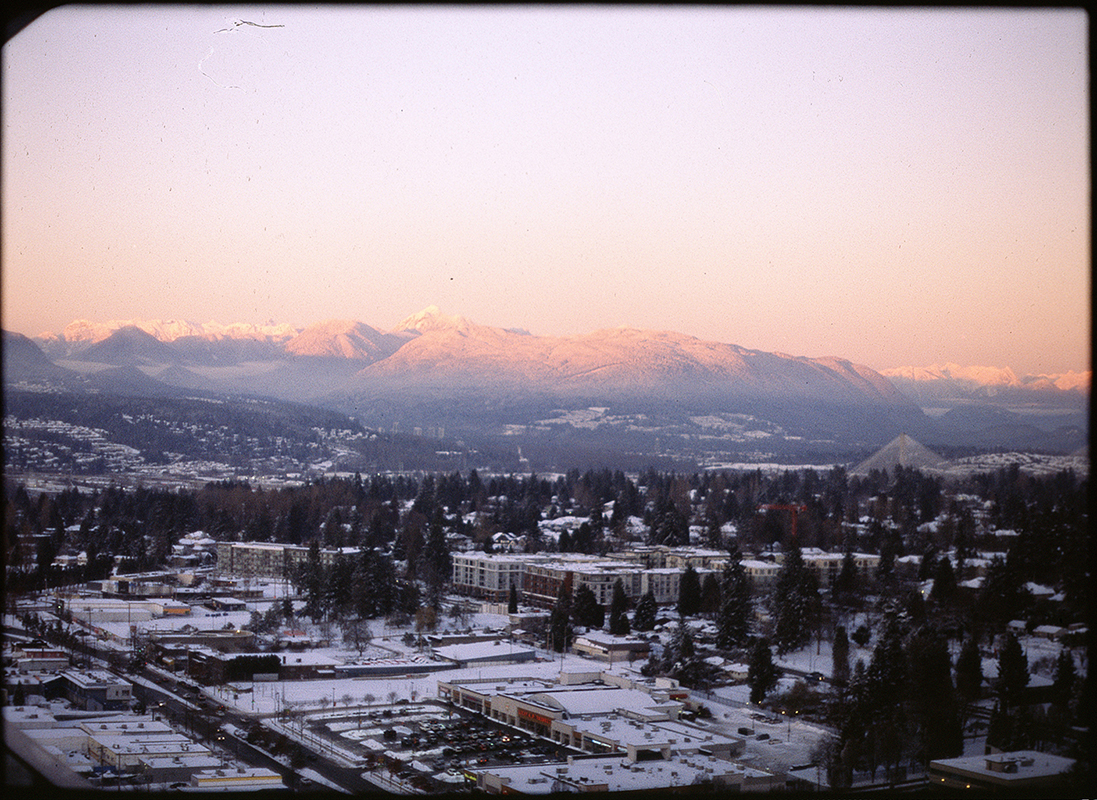}};
        \spy on (-1.63,1.75) in node [left] at (1.8,1.20);
        \end{tikzpicture}
    \caption{\textbf{Restoration by LaMa~\cite{suvorov2022resolution}}: best performing model, using our predicted segmentation.}
  \end{subfigure}
    \hfill
  \begin{subfigure}[t]{.325\textwidth}
    \centering
        \begin{tikzpicture}[spy using outlines={circle,blue,magnification=3,size=1.8cm, connect spies}]
        \node {\includegraphics[width=\linewidth]{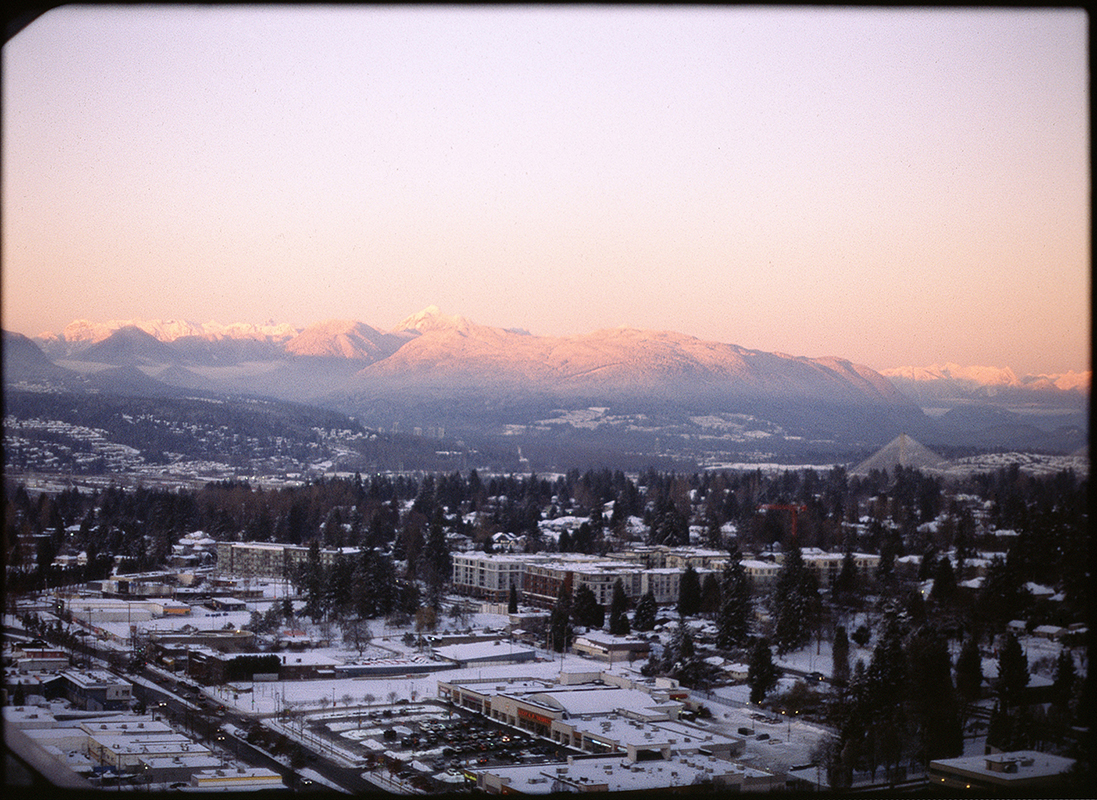}};
        \spy on (-1.63,1.75) in node [left] at (1.8,1.20);
        \end{tikzpicture}
    \caption{\textbf{Ground Truth}: manually restored in Photoshop by a human expert.}
  \end{subfigure}
  \caption{Input and ground truth from our authentic artefact damage dataset, along with restorations from some of the models we evaluated, presented at full resolution. Ours is the first public dataset allowing accurate quantitative evaluation of film damage restoration.}
  \label{fig:qualitative-comparison}
\end{figure*}

\section{Introduction}
Photographs captured on film constitute a major part of our cultural heritage and historical record, and many photographers continue to shoot on film. While it has appealing imaging qualities, film emulsion is highly susceptible to various kinds of mechanical damage: scratches, dust, hairs and dirt which mar the image when scanned and persist in darkroom and digital prints, as well as in cinematographic release prints. 

Isolating dust and scratches from natural image features in film scans and inpainting the damage are not difficult tasks for humans, but can be extremely time-consuming. \textbf{Automated film restoration} aims to localise and inpaint artefacts at a quality level comparable or exceeding that of human experts.

This is challenging: mechanical artefacts can have complex shapes and are non-uniformly distributed. They must be localised accurately when performing restoration, as it is crucial that restoration is applied only on the affected areas. It is also crucial to preserve the desirable qualities of analogue film, such as grain or characteristic colour grading, avoiding over-smoothing or distorted colour distributions.  

Automated film restoration can be broken down into two sub-tasks: artefact localisation (segmentation), and inpainting. Inpainting \cite{liu2018image,suvorov2022resolution} and segmentation \cite{ronneberger2015u,badrinarayanan2017segnet,minaee-segmentation-review-22} are well-studied in the literature, but relatively few models have been published that are suitable for professional-quality analogue film artefact restoration. Many existing general inpainting approaches are unsuitable as they operate at much lower resolutions than required for film scans (typically at least 4K), and rely on a mask being provided. Traditional solutions employed in commercial film scanners, such as Kodak's Digital ICE \cite{sanz-85,stavely-film-99,gann-film-08}, use a separate infrared illumination process to create an inpainting mask and apply simple nearest neighbour inpainting. This requires specialist hardware, only works on a limited number of emulsions, and cannot detect every form of artefact. We focus instead on a purely image-based automated film restoration process which works for all emulsion and damage types without additional imaging.

Large datasets with realistic film damage to train and evaluate on are not easily available \cite{chambah2006} and most ML-based film restoration instead relies on na\"ively generated synthetic images to approximate analogue artefacts\cite{iizuka2019deepremaster, mironicua2020generative,wan2020bringing,visapp22}; systems with for which authentic damaged images \textit{have} been collected have typically not made their data publicly available \cite{mironicua2020generative,wan2020bringing}. 
Unlike tasks such as JPEG artefact removal, super-resolution or colourisation, there are no robust simulation models of analogue film damage to generate high-quality synthetic training samples.
There is also no consensus as to what makes a “good” restoration, and therefore how a film restoration model should be evaluated, especially without human-restored ground-truth scans \cite{chambah2005,chambah2019}. 

\subsection{Contributions}

We make three contributions that address these issues:

\noindent\textbf{1. We curate and release a dataset of real analogue film scans (Section~\ref{sec:gt-dataset})}; the dataset consists of 4K scans of a variety of positive and negative colour emulsions, with damage artefacts such as dust, scratches, dirt and hairs present, paired with corresponding professionally manually restored versions. This is the first public dataset enabling quantitative evaluation of restoration models. 

\noindent\textbf{2. We construct a detailed statistical model of analogue film damage, allowing us to create highly-realistic synthetic data for training (Section~\ref{sec:modelling-damage}).}
To estimate the parameters of this statistical model, we collect another dataset of scanned empty film emulsion which has been scratched and left to get dusty; each damage artefact in the scans is manually annotated, resulting in a set of over 12,000 unique artefacts.
We extract statistics governing the shape, size and spatial density of different types of artefacts, which allows us to generate realistic synthetic analogue damage overlays to train restoration models.
We validate the resulting statistical model through a human perceptual study. In addition, we release a dataset of analogue image scans paired with their synthetically damaged versions, which we use to fine-tune or re-train models. 

\noindent\textbf{3. We use these datasets to perform a detailed empirical comparison of seven state-of-the-art models on film artefact restoration (Section~\ref{sec:comparison})}.
We include specialised film artefact restoration models, and general models that can be adapted for this purpose.
We perform evaluation on our dataset of images with real analogue damage and manually restored ground truths. We address the low-resolution limitations of existing methods that make them unsuitable for professional-grade restoration and extend processing to analogue film scans at 4K resolution by applying them patch-wise. Our synthetically damaged dataset allows us to train a segmentation network to supply damage masks to inpainting models that explicitly require them.

\begin{figure}
  \centering \includegraphics[width=1\linewidth]{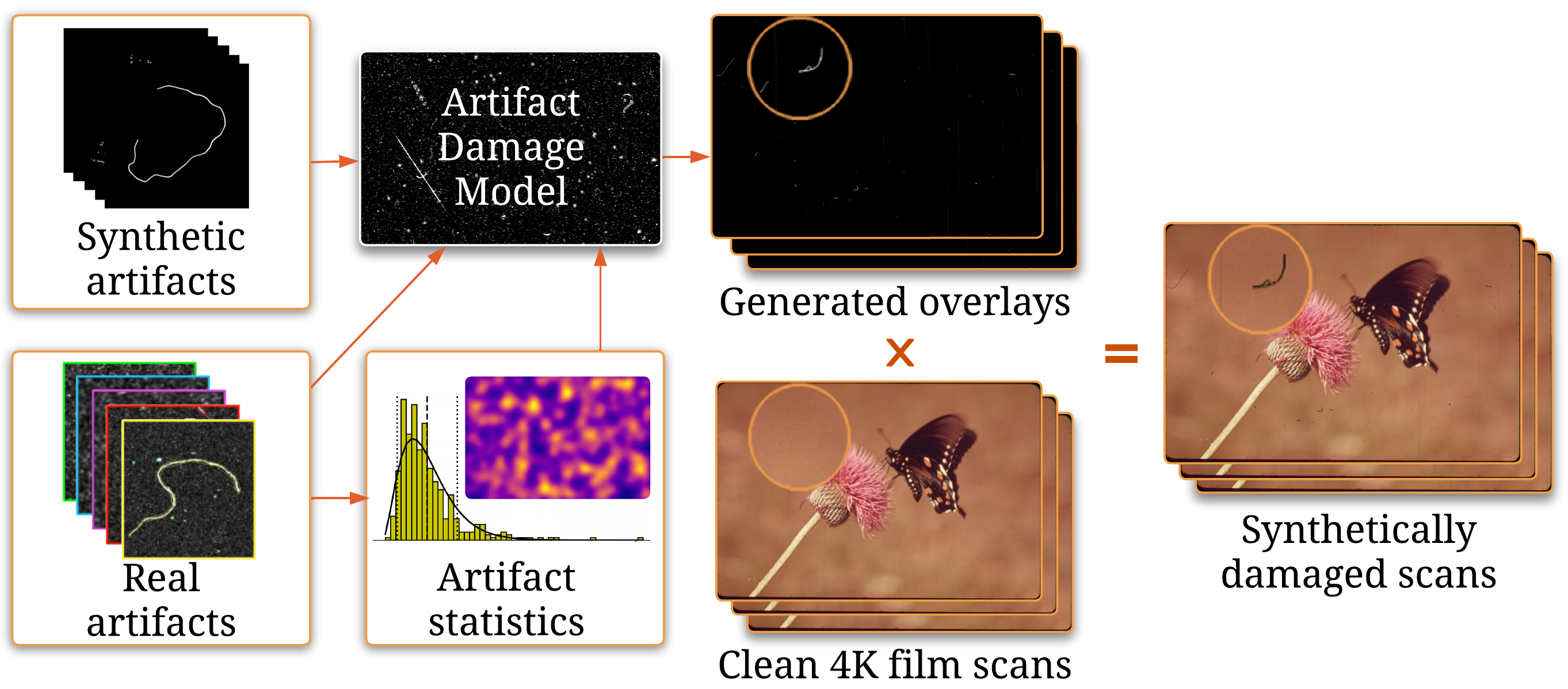}
  \caption{Overview of our proposed approach to generating realistic damage overlays showing analogue film artefacts (Section~\ref{sec:modelling-damage}). We extract statistics from 12~000 manually-annotated artefacts, and use these to build a probabilistic model of artefact damage that can be applied to clean film scans. The resulting damage overlays are highly realistic, and suitable for training image restoration models.}
  \label{fig1:overview-artefact-simulation}
\end{figure}

\section{Background and related work}

In this section we discuss definitions of film damage in the literature, and review existing machine learning approaches to film artefact damage restoration. Additionally, we consider state-of-the-art machine learning solutions to similar tasks which can be adapted to the film artefact restoration problems, and the challenges involved in doing so. 

\subsection{Film damage}

Analogue film damage is a loose term covering a variety of degradation types. Chambah \cite{chambah2019} roughly categorises analogue damage into two groups: chemical degradations and mechanical degradations. Chemical degradations, such as colour dye fading, contrast saturation and vinegar syndrome, are usually spatially homogenous \cite{chambah2019, wan2020bringing}; grain is also spatially homogenous, but while some works consider it to be another type of degradation \cite{wan2020bringing}, others point to it as an example of the ill-defined difference between film damage and artistically valued properties of the medium \cite{chambah2019, visapp22}. On the other hand, mechanical degradations, for example abrasions to the emulsion (scratches) or dust specks or hairs stuck to the emulsion, are not spatially uniform \cite{chambah2019, wan2020bringing}. 

As film is the source record for printing images, artifacts which affect the photochemical emulsion will be carried over to any prints made from it \cite{chambah2019}. Therefore, damaged film scans and the resulting damaged reproductions can be restored via the same techniques. However, further damage inflicted on the print itself, such as tears and folds, is independent of the emulsion. While some works consider film and print degradations interchangeable \cite{wan2020bringing}, it is notable that they have distinct properties in terms of scale and spatial distributions with respect to the medium in which they originally occur. Furthermore, print degradations are not unique to prints of images captured on film, and as such are beyond the scope of this work.

Removal of any type of degradation can be reframed as an image-to-image translation problem, where some degradation operation has been applied to the original, non-damaged image, and needs to be inverted, similar to denoising \cite{pmlr-v80-lehtinen18a, zamir2021multi, Zamir_2022_CVPR, liang2021swinir, chen2022simple}, superresolution \cite{johnson2016perceptual, wang2018esrgan, liang2021swinir}, colorisation \cite{zhang2016colorful, isola2017image, kumar2021colorization, saharia2022palette}. For these tasks, the conventional approach is to generate training data from clean images by applying a transform which is identical to the degradation process, e.g.~with colourisation, grayscale versions of the training data images can be easily derived from color images. Yu et al. \cite{yu2018crafting} propose a reinforcement learning approach to modelling more complex degradations as a dynamic mix of simpler types of damage, however, it is important to note that the most common degradations in the image denoising, restoration and in-painting tasks addressed in the literature happen entirely in the digital domain. Film damage, and especially mechanical film damage, in contrast, is a product of the physical properties of the film emulsion \cite{chambah2019}, and is translated into the digital domain during the film scanning process. Due to their analogue nature, lack of spatial uniformity, variability in opacity, and randomness of shape and size, analogue artefacts such as scratches, dust specks and hairs are more challenging to model, and by extension, digitally simulate and restore \cite{chambah2019, wan2020bringing, mironicua2020generative, visapp22}.

Film damage has been simulated na\"ively by directly compositing a small set of full frame damage textures over clean images \cite{iizuka2019deepremaster, wan2020bringing, mironicua2020generative, visapp22}. Although transforms such as resizing, rotating, flipping and randomly cropping the damaged textures are employed in order to introduce more diversity in the synthetic data, to our knowledge, no approach in the literature varies the shape, location, and/or rotation of individual artefacts on the emulsion, nor their size with respect to the contents of the image and the film frame. 

\subsection{Inpainting}

Inpainting is a task in which missing regions in the image, indicated by a binary mask, are filled in by estimation based on neighbouring pixel and global image context. The task lends itself to being reframed as part of the analogue artefact restoration problem: image areas obscured by the artefacts need to be suitably inpainted.
However, there are some challenges: artefacts have arbitrary shape, size and location, which, in addition, are not known a priori.
Recent advances in deep learning have enabled inpainting of larger image regions by generating semantically consistent content \cite{suvorov2022resolution}. The task has been tackled by various families of models, such as VAEs \cite{ham18, peng2021generating}, GANs \cite{yu2018generative}, and diffusion models \cite{saharia2022palette, rombach2021highresolution, lugmayr2022repaint}. While earlier approaches are constrained in the mask shape applied for inpainting, some recent models consider arbitrary masks.
Another related task is single-image de-raining, which has been tackled by CNN-based approaches \cite{ren2019progressive}, and more recently, by transformer-based ones, such as Restormer \cite{Zamir_2022_CVPR}, which is currently state-of-the-art.
Blind image inpainting is an extension of the inpainting task, where the mask indicating areas to be inpainted is not provided \cite{wang20eccv,cai17tvc}. Hertz et al. \cite{hertz2019blind} propose a deep learning approach which attempts to address this problem in the context of watermark removal, by also predicting a mask to separate out the areas to be inpainted from the ares which are to be preserved. In line with this method, state-of-the-art inpainting approaches can be adapted to blind image inpainting via an additional segmentation network tasked with predicting masks for the inpainting network to use. 
\subsection{Processing of high-resolution data}
Modern camera-equipped devices are able to capture images at very high resolution (i.e. mega- and even gigapixels). The exact definition of “high resolution” varies between applications \cite{bakhtiarnia2022efficient}; in film scanning, it is common to scan images at 3000DPI and at least 4K resolution. Many of the image-based tasks that have drawn attention in the machine learning community in the last few years do not require for the image input to be processed at full resolution -- e.g.~classification models are conventionally trained on cropped and downsampled images of size $224\times224$ pixels to meet computational limitations. Talebi \& Milanfar \cite{talebi2021learning} even investigate learning the resizing operator on the input to improve network performance while maintaining the constraint for lower resolution image input. On the contrary, few state-of-the-art image restoration approaches, e.g.~Restormer \cite{Zamir_2022_CVPR} and LaMa \cite{suvorov2022resolution} claim the ability to process high resolution image data.
When applying machine learning approaches to problems in fields such as medical imaging, processing large medical image scans at their original resolutions is crucial. The standard approach is to split the images into patches, have the model process them, and stitch them back together \cite{pielawski2020introducing}. Processing film scans poses a very similar problem.

\subsection{Deep learning for analogue film restoration}

Due to the challenges in modelling film damage and collecting relevant data, there are few approaches in the literature which set out to solve the specific problem of film artefact restoration. Strubel et al. \cite{strubel:hal-02369128} train a SegNet model to remove dust and scratches, and provide a limited dataset of grayscale image scans for training and evaluation. Mironica \cite{mironicua2020generative} proposes a GAN-based approach to film artefact restoration; to generate training data, they use a set of 100 synthetic overlays applied over a set of 2500 clean film scans, which are randomly cropped to patches of $128\times128$ pixels. The qualitative results in the paper demonstrate some success in inpainting smaller artefacts, at the expense of overly smoothing grain. Similarly, Wan et al. \cite{wan2020bringing} address the problem of limited training data by applying synthetic damage overlays onto digital images, along with synthetic grain to mimic film scans and prints; they train a segmentation U-Net to predict damage masks, and further jointly train two VAEs to translate images between two latent spaces, corresponding to the domains of damaged and restored photos, respectively. The model is trained on cropped patches of size $256\times256$ pixels. Since the training data is derived from digital images, the restorations produced by this approach have the effect of overly smoothing grain and shifting color in actual film scans from the test set, i.e.~the network itself introduces loss of information to the input, including novel artefacts. A follow-up work by the same authors extends the approach to analogue video sequences \cite{wan2022bringing}. Similarly, DeepRemaster~\cite{iizuka2019deepremaster} also applies pre-rendered damage overlays to sequence frames to model analogue damage. Finally, Ivanova et al. \cite{visapp22} propose a U-Net restoration network trained with a perceptual loss, which tackles the artefact detection and restoration tasks simultaneously. As with previous works, film damage is again simulated by applying a set of overlays, modified by various simple transforms, over clean analogue film scans. While this restoration network is qualitatively and quantitatively shown to outperform the method of Wan et al., it is limited in the resolution of inputs which it can be applied to, due to being trained on images downsampled and cropped to $256\times256$ pixels.

\begin{figure}[hbtp]
\vspace*{-2pt}
\centering
\subcaptionbox{\textbf{Dirt} artefacts are irregularly shaped and of varying size.\label{fig2.1:dirt}}{%
\includegraphics[width=1.\linewidth]{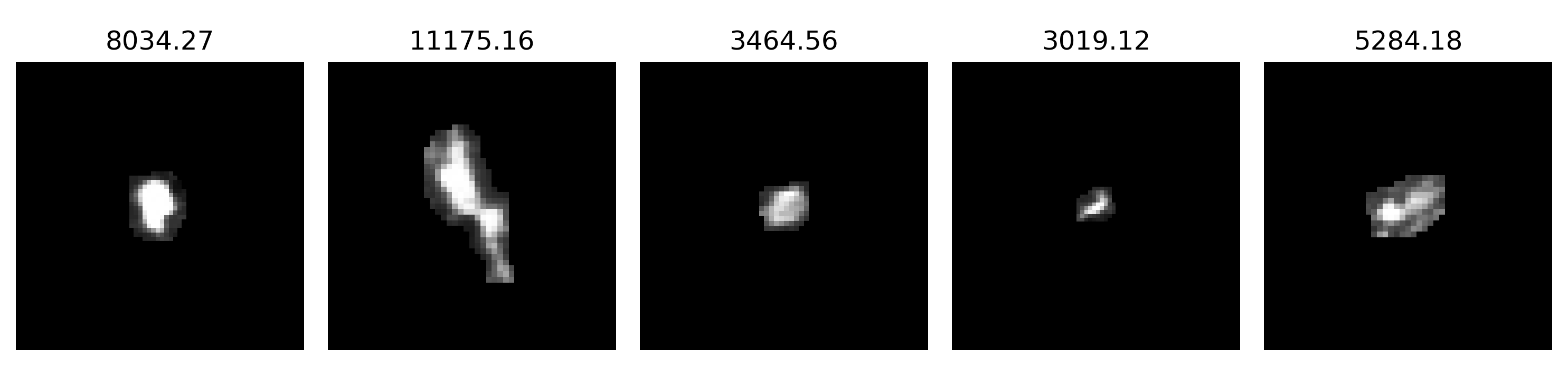}\vspace{-6pt}%
}\hfill
\subcaptionbox{\textbf{Dust} specks are round and relatively small in size.\label{fig2.2:dust}}{%
\includegraphics[width=1.\linewidth]{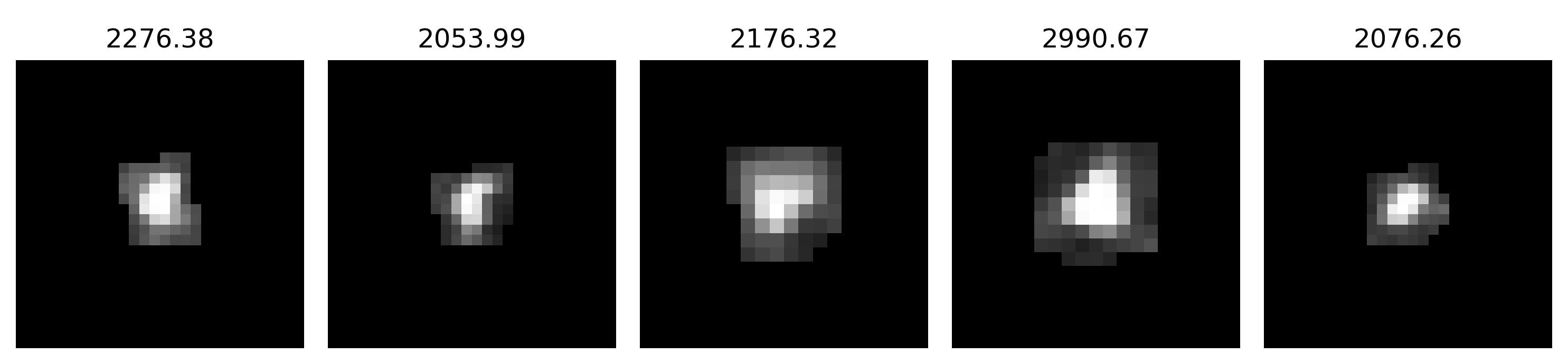}\vspace{-6pt}%
}\hfill
\subcaptionbox{\textbf{Long hairs} are thin, overall large in size and of varying shape and extent.\label{fig2.3:long-hairs}}{%
\includegraphics[width=1.\linewidth]{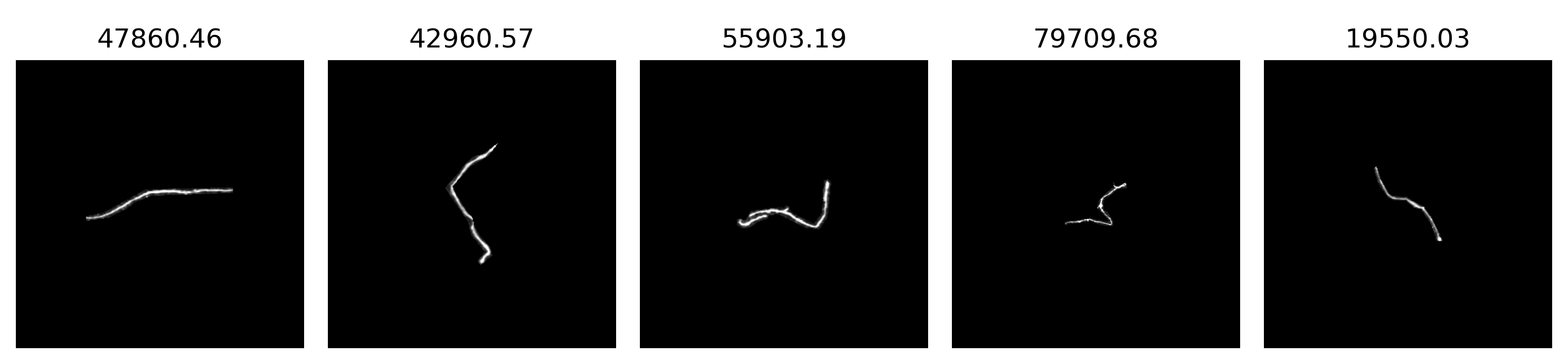}\vspace{-6pt}%
}\hfill
\subcaptionbox{\textbf{Short hairs} are similar to long hairs but smaller in length (and size).\label{fig2.4:short-hairs}}{%
\includegraphics[width=1.\linewidth]{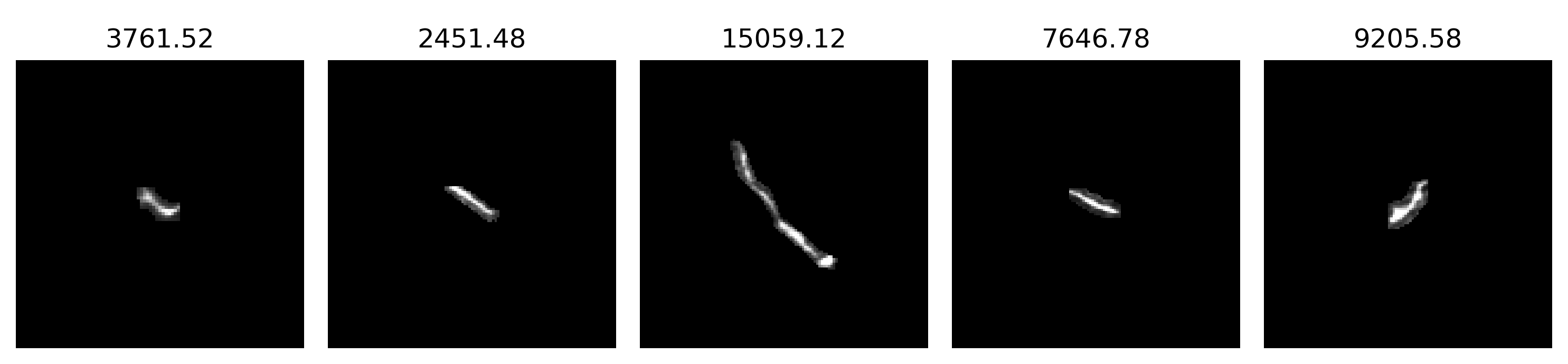}\vspace{-6pt}%
}\hfill
\subcaptionbox{\textbf{Scratches} are comparable to long hairs in being thin and long, but without significant curvature, and often even larger in size. \label{fig2.5:scratches}}{%
\includegraphics[width=1.\linewidth]{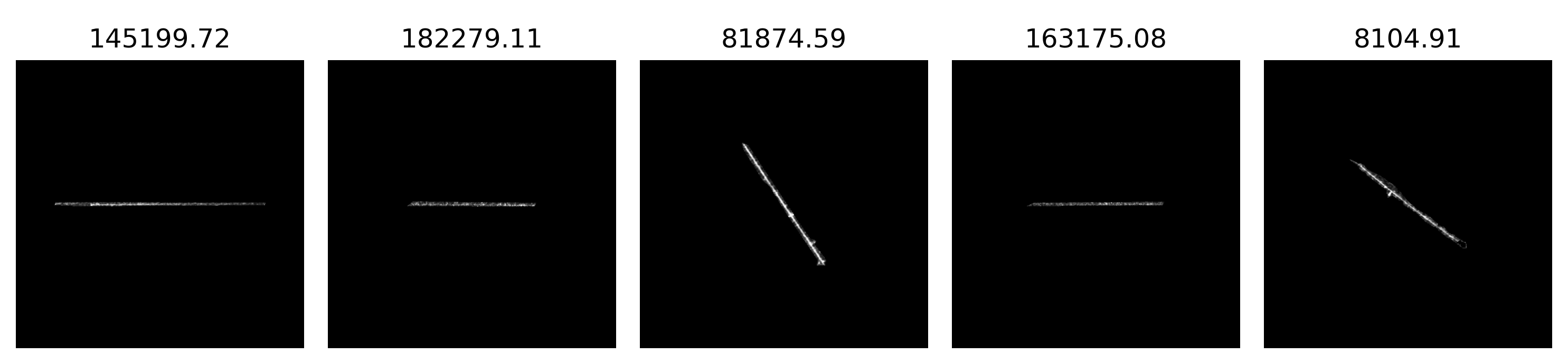}\vspace{-6pt}%
}
\caption{Examples of extracted and padded artefacts. The number above each is its area in $\mu m^2$ (square microns).}
\label{fig2:artefact-examples}
\vspace{-12pt}
\end{figure}

\section{Evaluation data for damage restoration}
\label{sec:gt-dataset}

We curate a dataset of 35mm film scans at 4K resolution, with varying degree of authentic artefact damage in the form of dust, scratches, hairs and dirt. The dataset includes 44 images of various film emulsions, both positive (slide) and negative. The content of the images is also diverse, including landscapes, architecture shots and still lifes. The images have been shot, developed and scanned by Dmitri Tcherbadji of Analog.Cafe, and are used with his kind permission. In addition, each damaged image in the set has been paired with a ground-truth restoration via manual inpainting of the artefacts in Photoshop by the same expert.
Original and manually-restored versions of an example photograph from the dataset are shown in Figure~\ref{fig:qualitative-comparison}.
To our knowledge, this is the only public dataset of high quality damaged film scans paired with expert restoration ground truths. We have made the dataset available at \URL{www.doi.org/10.6084/m9.figshare.21803304}.

\begin{figure}[hbtp]
\centering
\vspace*{-2pt}
\begin{tikzpicture}[zoomboxarray, zoomboxes below, zoomboxarray inner gap=0.1cm, zoomboxarray columns=3, zoomboxarray rows=2]
    \node [image node] { \includegraphics[width=1.\linewidth]{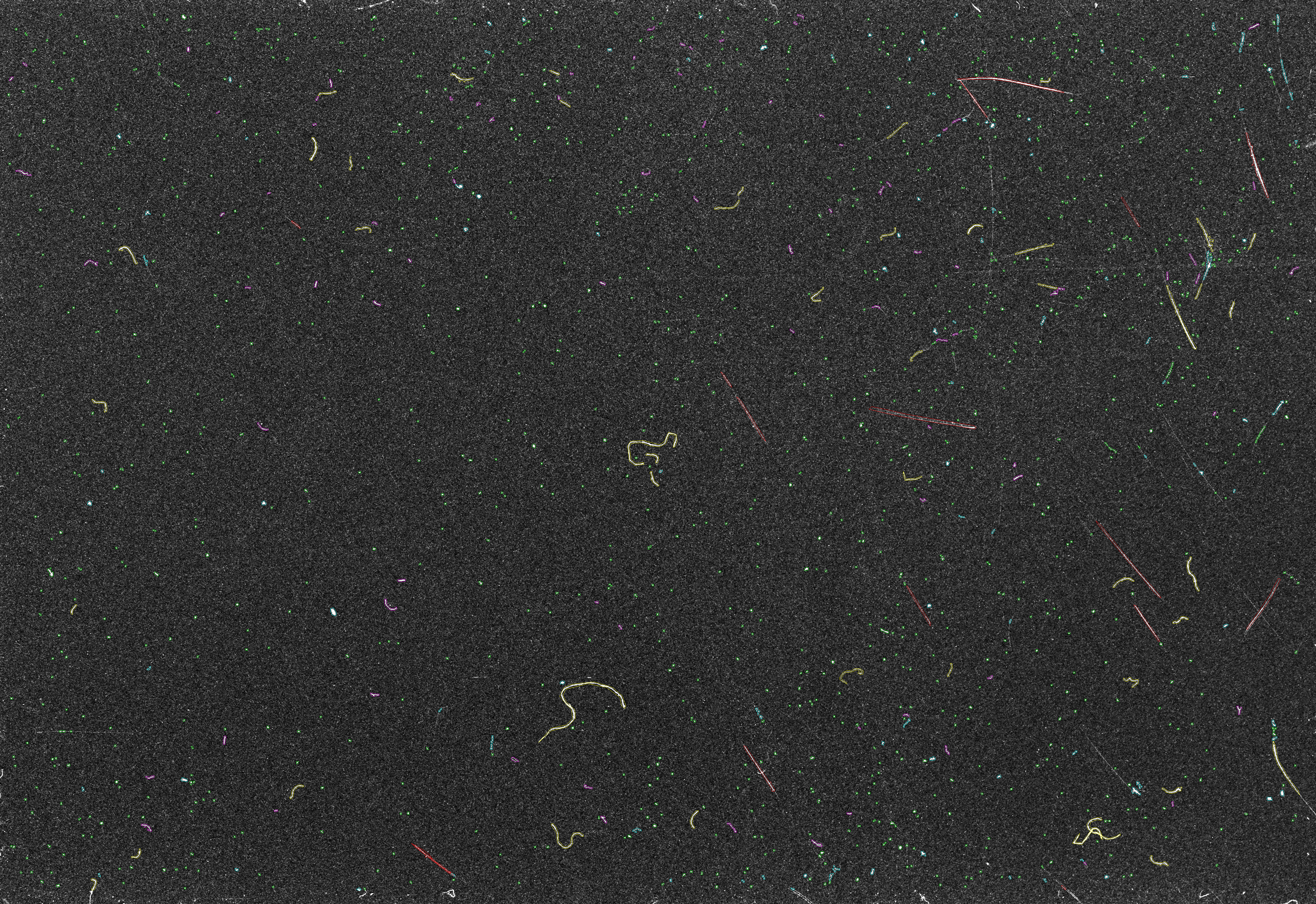} };
    \zoombox[magnification=9, color code=green]{0.482,0.78}
    \zoombox[magnification=12, color code=cyan]{0.917,0.705}
    \zoombox[magnification=15, color code=cyan]{0.254,0.324}
    \zoombox[magnification=4, color code=yellow]{0.445,0.22}
    \zoombox[magnification=16, color code=Magenta3]{0.671,0.79}
    \zoombox[magnification=7, color code=red]{0.87,0.31} 
\end{tikzpicture}
\caption{An example of an annotated scan. Full resolution (4944 by 3396 pixels) is shown at the top, along with zoomed-in regions around examples of each type of artefact below. Overlaid annotations for dust (green), dirt (blue), long hairs (yellow), short hairs (magenta), and scratches (red).}
\label{fig1:example-annotated-scan}
\vspace{-6pt}
\end{figure}

\section{Modelling analogue film damage}\label{sec:modelling-damage}
In this section, we describe our novel approach to generating synthetic analogue damage.
Our overall approach is as follows:
\begin{enumerate}
    \item Annotate, classify and extract over 12~000 individual real analogue film artefacts from heavily-damaged high resolution scans (Section \ref{subsec:capturing-real-artefacts}). 
    \item 
    Calculate statistics of the 
    extracted artefacts, such as size, count, and spatial density (Section \ref{subsec:analysing-artefact-stats}).
    \item Build a probabilistic model to generate new artefact damage overlays using a combination of the extracted artefacts and synthetic ones, parameterised by the recorded statistics (Section \ref{subsec:generating-artefact-overlays}).
\end{enumerate}

We demonstrate the realism of damage overlays generated by our model, via a perceptual user study (Section~\ref{subsec:validating-generated-damage}).
We further validate the damage by evaluating it in the context of artefact segmentation (Section~\ref{subsec:segmentation-comparison}) and artefact restoration.
In both cases, the models trained with data synthesised via our approach outperform the alternatives.
The full damage synthesis pipeline, including the annotated analogue damage scans, is available at \URL{www.github.com/daniela997/FilmDamageSimulator}.

\paragraph{Artefact types.}
Prior works have classified damage as dust, scratches or hairs \cite{chambah2019, wan2020bringing}.
We extend this taxonomy by splitting hairs into two classes, short and long, due to their large variation in size; we also define an additional class, dirt, to capture artefacts of irregular shape which are larger than dust specks, but are neither hairs nor scratches.

\begin{figure}[hbtp]
    \vspace{-8pt}
    \centering
        \includegraphics[width=0.75\linewidth,]{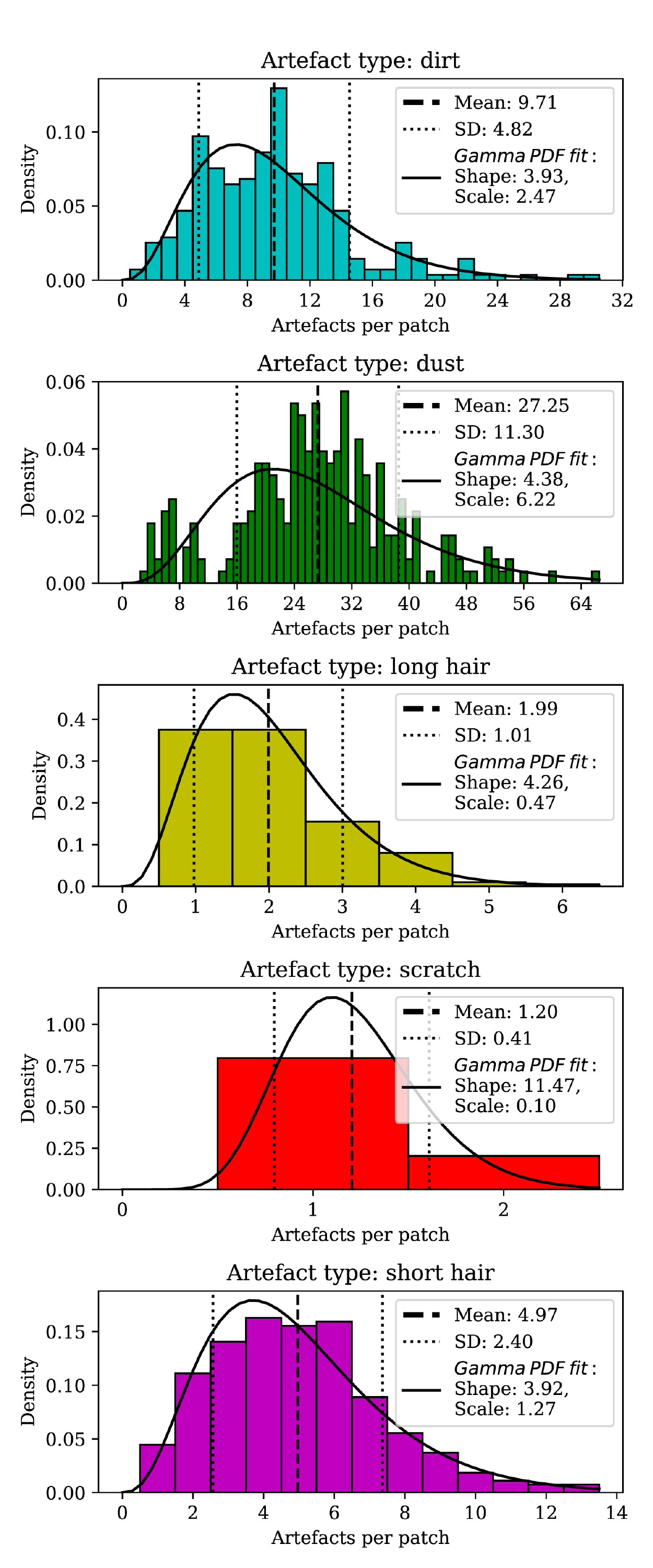}
    \caption{Number of artefacts per $256\times256$ pixel patch for each artefact type across all scans, and fitted Gamma distributions.
    }
    \label{fig3:artefact-counts-per-patch}
    \vspace{-6pt}
\end{figure}

\subsection{Capturing real analogue film artefacts}\label{subsec:capturing-real-artefacts}

To gather samples of real-life dust specks, scratches, hairs and dirt, we scan 10 heavily damaged, empty 35mm film frames of Lomochrome Color Negative 400 ISO film using a Plustek OpticFilm 8100 dedicated film scanner and SilverFast 9. The resulting scans are saved at 4K resolution.

We manually annotate individual artefacts in the scans with bounding polygons, and classify each as dirt, dust, long hair, short hair or scratch (see Figure~\ref{fig1:example-annotated-scan} for examples).
We calculate the area of each polygon, and convert these to physical units of square microns based on the ratio between the scanned frame’s size in pixels, and its size in millimeters -- 35mm on the long edge and 24mm on the short edge.
Finally, we extract each artefact, zero padded to square, to create a bank of isolated artefacts to sample from when generating new overlays; examples for each class are visualised in Figure~\ref{fig2:artefact-examples}.
In total we have annotated 12135 artefacts across the 10 scanned frames. 

\subsection{Analysing artefact statistics}\label{subsec:analysing-artefact-stats}

For each artefact class, we collect several statistics to ensure our generated overlays match the distribution of real film artefact damage.
We measure individual artefacts (area in $\mu m^2$, square microns), as well as their distribution over the entire frame (counts, spatial frequency). Artefact counts and sizes are summarised in Table~\ref{tab:counts-sizes-summary}; Figure~\ref{fig5:size-distributions} displays the full distributions of sizes for each class.

\begin{table}[hbtp]
\begin{tabular}{@{}llll@{}}
\toprule
\textbf{Artefact type} & \textbf{Count} & \textbf{Avg. area ($\mu$$m^2$)} & \textbf{Std. dev. area ($\mu$$m^2$)} \\ \midrule
\textit{dirt}                                                     & 2700           & 8194                        & 10304                                            \\
\textit{dust}                                                     & 7631           & 3344                        & 1637                                           \\
\textit{long hair}     
        & 398            & 53501               & 35524                                                    \\
\textit{short hair}                                              & 1341           & 16365                       & 11735                                            \\
\textit{scratch}                                                & 65             & 229660                      &521345                                            \\ \bottomrule
\end{tabular}
\caption{Statistics of annotated artefacts -- we report the total count and average area for each artefact class. See Figure~\ref{fig5:size-distributions} for the full distributions of areas.}
\label{tab:counts-sizes-summary}
\vspace{-6pt}
\end{table}

\paragraph{Counts.}
For all 10 scans, we observe strong class imbalance in favor of dust and dirt, with scratches being very scarce.
Moreover, the artefacts’ spatial distribution is not uniform; we therefore split each scan into $256\times256$ pixel patches (padding as required) and record the artefact counts for each class in each patch. The resulting distributions are shown in Figure \ref{fig3:artefact-counts-per-patch}.

\begin{figure}[hbtp]
\centering
\subcaptionbox{
\label{fig4.1:spatial-freq-short-hair}}{%
\includegraphics[width=0.48\linewidth,trim={3em 2em 8em 3em},clip]{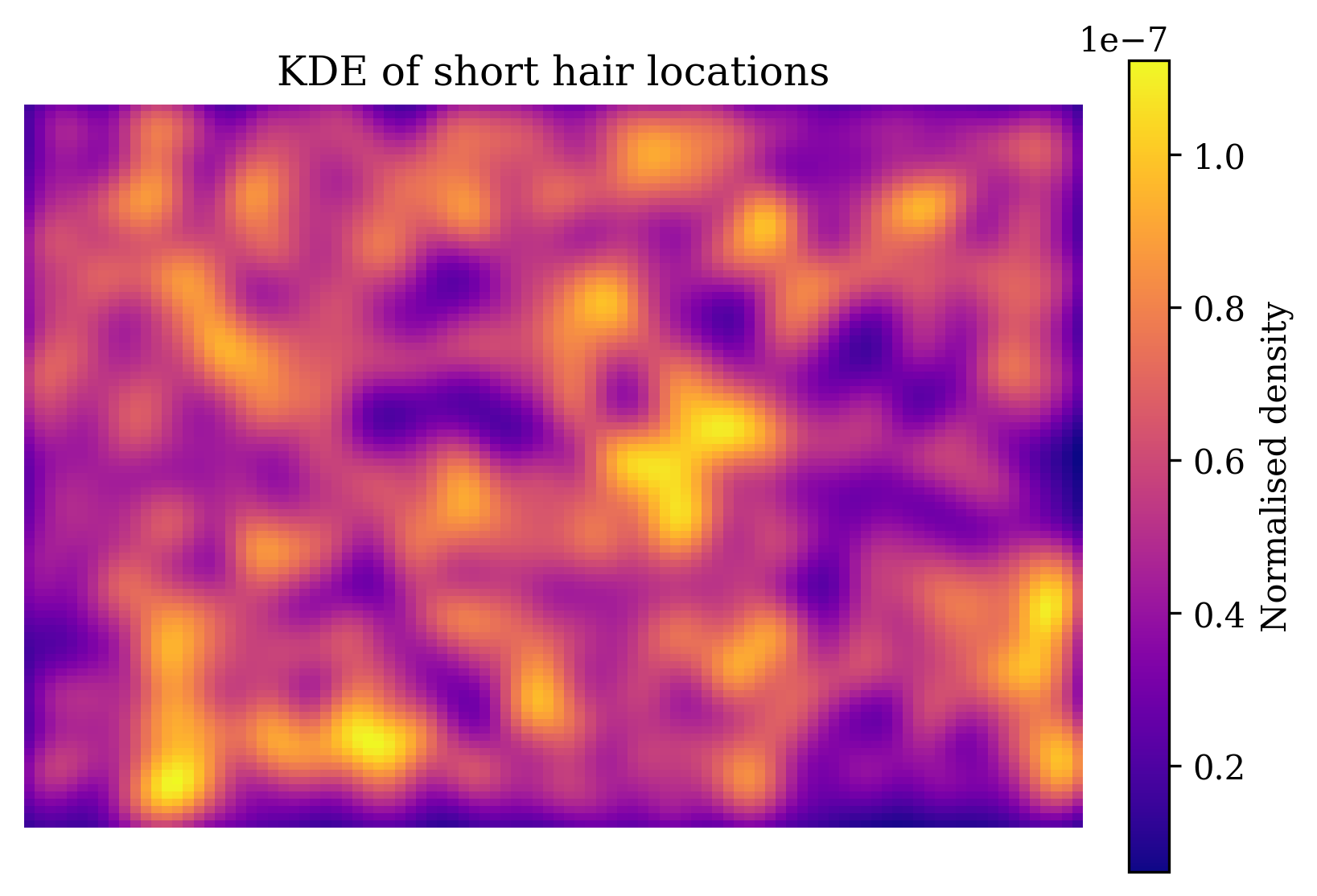}%
}
\subcaptionbox{
\label{fig4.2:perlin-noise}}{%
\includegraphics[width=0.48\linewidth,trim={3em 2em 8em 3em},clip]{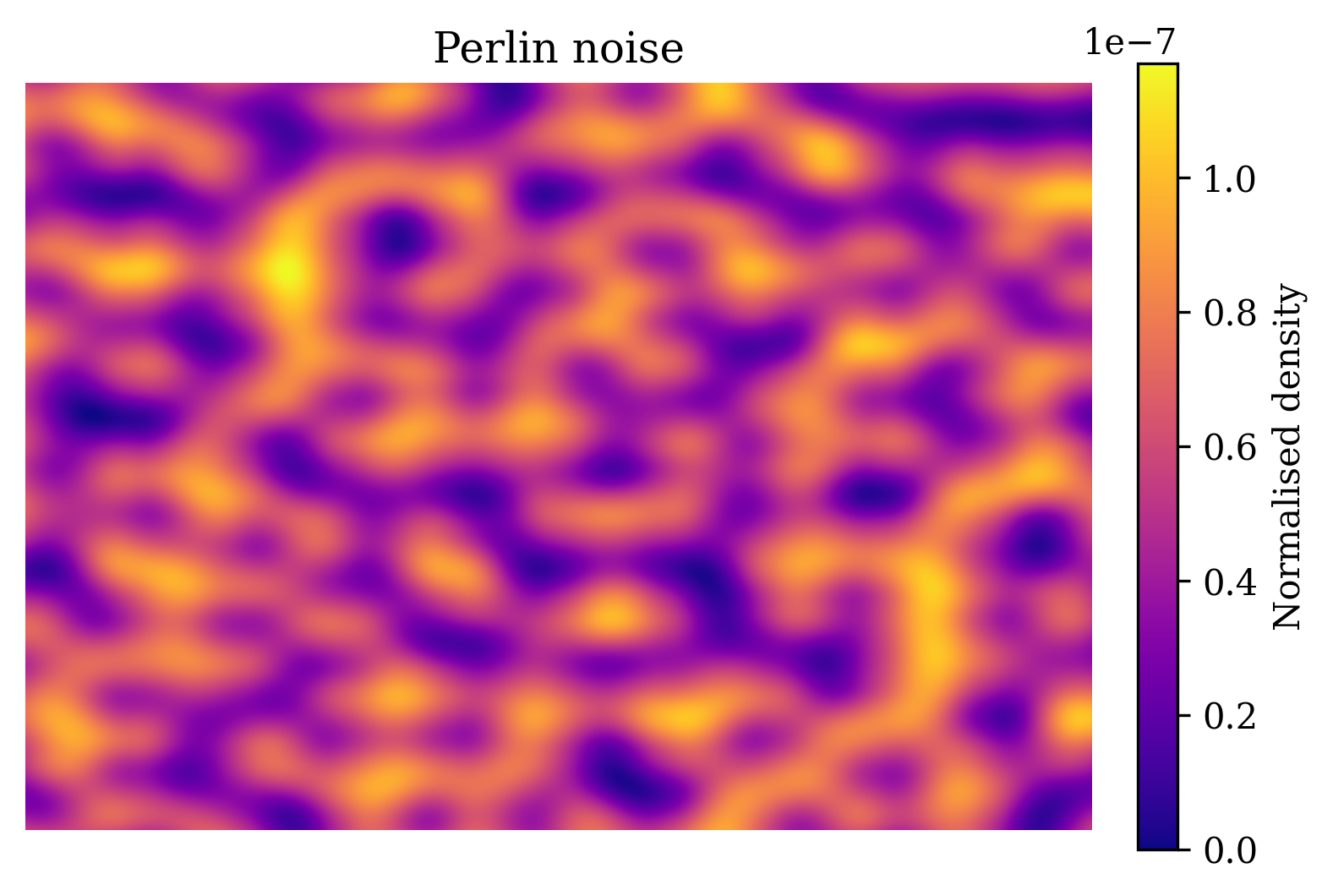}
}
\vspace{-6pt}
\caption{Visual comparison between the true spatial distribution of short hairs (by kernel density estimation on annotation centroids) \textbf{(a)}, and Perlin noise \textbf{(b)}, which we use as an approximation when sampling new overlays. Each plot shows the distribution over a full 35mm frame, with brighter colors corresponding to higher density.
}
\label{fig4:spatial-freq}
\end{figure}

\paragraph{Spatial frequencies.}
We visualise the spatial distribution of artefacts via kernel density plots, for an example class in Figure~\ref{fig4.1:spatial-freq-short-hair} and for all classes in the supplementary material (Figure S5).
While there is no obvious pattern to the distributions of artefact occurrences, we can observe that four artefact types are more frequent near the frame’s upper and lower left corners; this could be related to the direction in which the film strip is advanced inside the camera and/or the scanner.

\begin{figure}[hbtp]
    \centering
    \vspace{-10pt}
    \includegraphics[width=0.85\linewidth,]{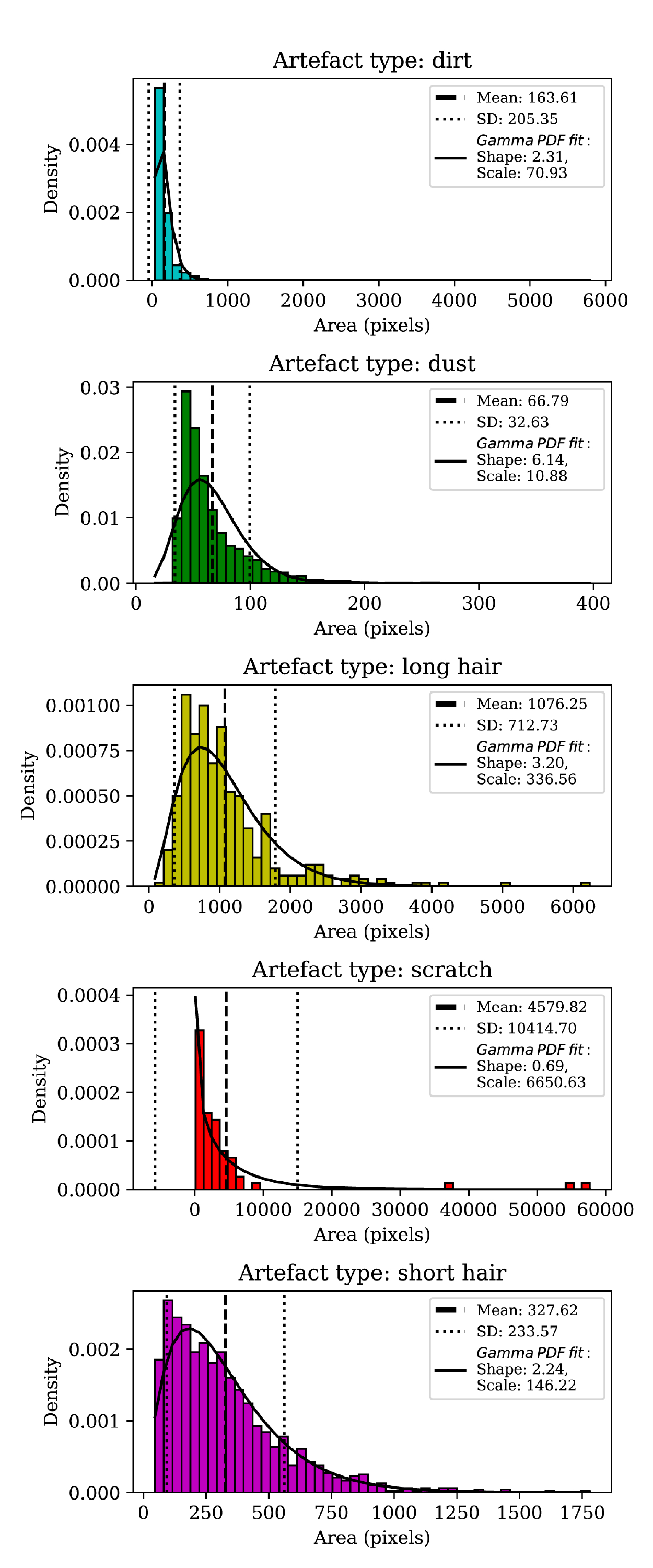}
    \caption{Artefact size histograms in pixels, and fitted Gamma distributions. While average sizes for each type of artefact vary, the sizes of artefacts are distributed in a similar way. }
    \label{fig5:size-distributions}
    \vspace{-6pt}
\end{figure}

\begin{table*}[]
\centering
\begin{tabular}{@{}llllllllllllll@{}}
\toprule
                                      &                                                  & \multicolumn{12}{c}{\textbf{Resolutions}}                                                                                                                                                                             \\ \cmidrule{3-14}
\multirow{2}{*}{\textbf{Familiarity}} & \multirow{2}{*}{\textbf{Participants}} & \multicolumn{2}{c}{$\mathbf{128\times128}$} & \multicolumn{2}{c}{$\mathbf{256\times256}$} & \multicolumn{2}{c}{$\mathbf{512\times512}$} & \multicolumn{2}{c}{$\mathbf{1024\times1024}$} & \multicolumn{2}{c}{$\mathbf{2048\times2048}$} & \multicolumn{2}{c}{\textbf{Overall}} \\
\cmidrule(r){3-4}\cmidrule(lr){5-6}\cmidrule(lr){7-8}\cmidrule(lr){9-10}\cmidrule(lr){11-12}\cmidrule(l){13-14}
                                      &                                                  & \textbf{Mean}  & \textbf{Std}  & \textbf{Mean}  & \textbf{Std}  & \textbf{Mean}  & \textbf{Std}  & \textbf{Mean}   & \textbf{Std}  & \textbf{Mean}   & \textbf{Std}  & \textbf{Mean}    & \textbf{Std}    \\
\textbf{Not Familiar}                 & 49                                               & 52.4           & 12.7            & 52.6           & 11.1            & 50.6           & 9.6             & 52.5            & 12.4            & 50.3            & 13.2            & 51.5             & 6.9               \\
\textbf{Somewhat Familiar}            & 81                                               & 52.9           & 10.1            & 54.1           & 12.0            & 51.7           & 10.0            & 54.4            & 13.6            & 51.1            & 15.9            & 52.9             & 7.0               \\
\textbf{Very Familiar}                & 121                                              & 54.9           & 11.1            & 53.5           & 11.3            & 50.4           & 12.2            & 49.9            & 12.1            & 48.1            & 16.1            & 51.4             & 7.0               \\
\textbf{Overall}                      & 251                                              & 53.8           & 11.2            & 53.5           & 11.5            & 50.8           & 11.0            & 51.9            & 12.7            & 49.6            & 15.5            & 51.9             & 7.0               \\ \bottomrule
\end{tabular}%
\caption{Summary of participant scores (in \%) based on responses to Survey 1. A score of $50\%$ is optimal since it indicates that our synthetic damage is on average indistinguishable from real damage.}
\label{tab:survey1-sumary}
\end{table*}

\subsection{Synthetic artefacts.}
In addition to the artefacts extracted from the 10 scans, we also use a set of 6100 synthetic dust, scratch, hair, lint, and dirt artefacts, which were manually drawn in Photoshop and kindly shared by Stefan Ringelschwandtner of Mononodes.
We do this to further increase the diversity of rarer artefact classes.
Every artefact in this set is of size $400\times400$ pixels, and therefore must be rescaled to match the observed distribution of areas for real artefacts of the corresponding class.
As the scratch class is particularly under-represented, we also programatically generate additional scratches matching the appearance of those found in real scans.

\subsection{Generating damage overlays}\label{subsec:generating-artefact-overlays}

We develop a probabilistic model based on the measured artefact properties, which allows us to generate new synthetic full-frame damage overlays with a realistic distribution of artefacts.
Our generative process is as follows:
\begin{enumerate}
    \item 
\textbf{Sample the numbers of artefacts for each class:}
For each artefact class, we sample a target count from a Gamma distribution fit by maximum likelihood estimation to the empirical counts in Figure~\ref{fig3:artefact-counts-per-patch} and round to the nearest integer. 
As the recorded counts are per $256\times256$ pixel patch, we scale the sampled counts depending on the target overlay resolution.

\item \textbf{Sample the artefact sizes:}
Similarly, we sample artefact sizes from Gamma distributions fitted to the observed artefact sizes in Figure~\ref{fig5:size-distributions}. The sampled sizes are further rescaled according to the target overlay size, using the relevant pixel-to-micron conversion ratio.

\item \textbf{Sample the artefact appearances:} 
For each artefact, we randomly choose its appearance from among the real and synthetic artefacts of the relevant class.
We ensure that the size of each (in pixels) is not divergent from the target sizes sampled in the previous step, in order to avoid excessive upsampling of small artefacts in classes with high variance in area (e.g.~dirt).

\item \textbf{Sample locations and rotations:}
We approximate the spatial densities discussed earlier with Perlin noise \cite{perlin}; we justify this choice by a visual comparison with the true artefact density aggregated over all scans (Figure~\ref{fig4:spatial-freq}).
We sample the location of each artefact independently from this noise distribution; we also sample a rotation from the uniform distribution on $[-\pi,\,\pi]$.

\item \textbf{Compose the final overlay:}
Given the artefact appearances, sizes, locations and rotations, we alpha-composite them into the frame at the required output resolution. The overlay can be used to simulate damage to film negatives (artefacts are white), or to developed slides (artefacts are black).

\end{enumerate}

\begin{figure*}
  \subcaptionbox{}{\includegraphics[width=0.43\linewidth]{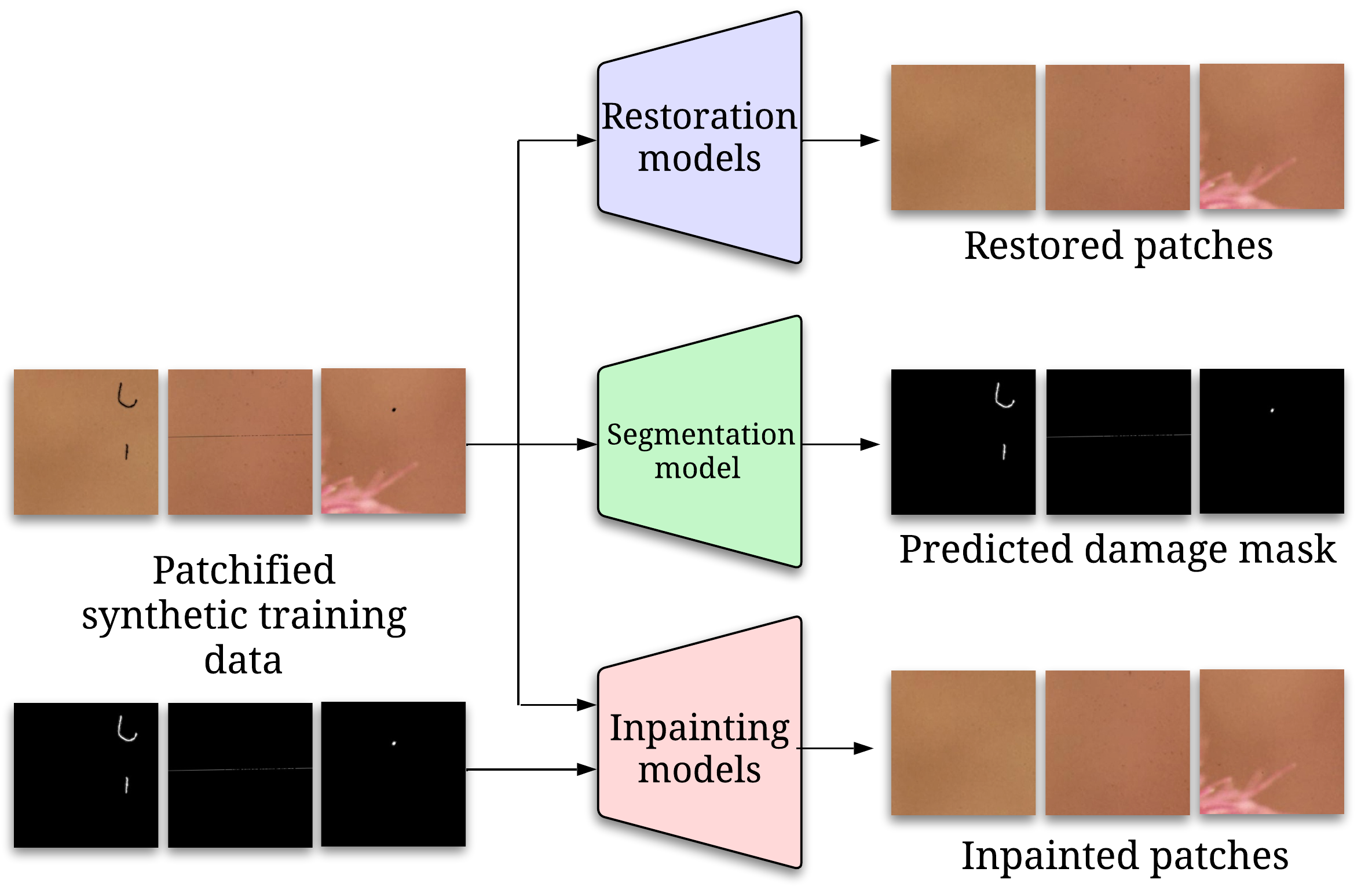}\vspace{2pt}}
  \subcaptionbox{}{\includegraphics[width=0.55\linewidth]{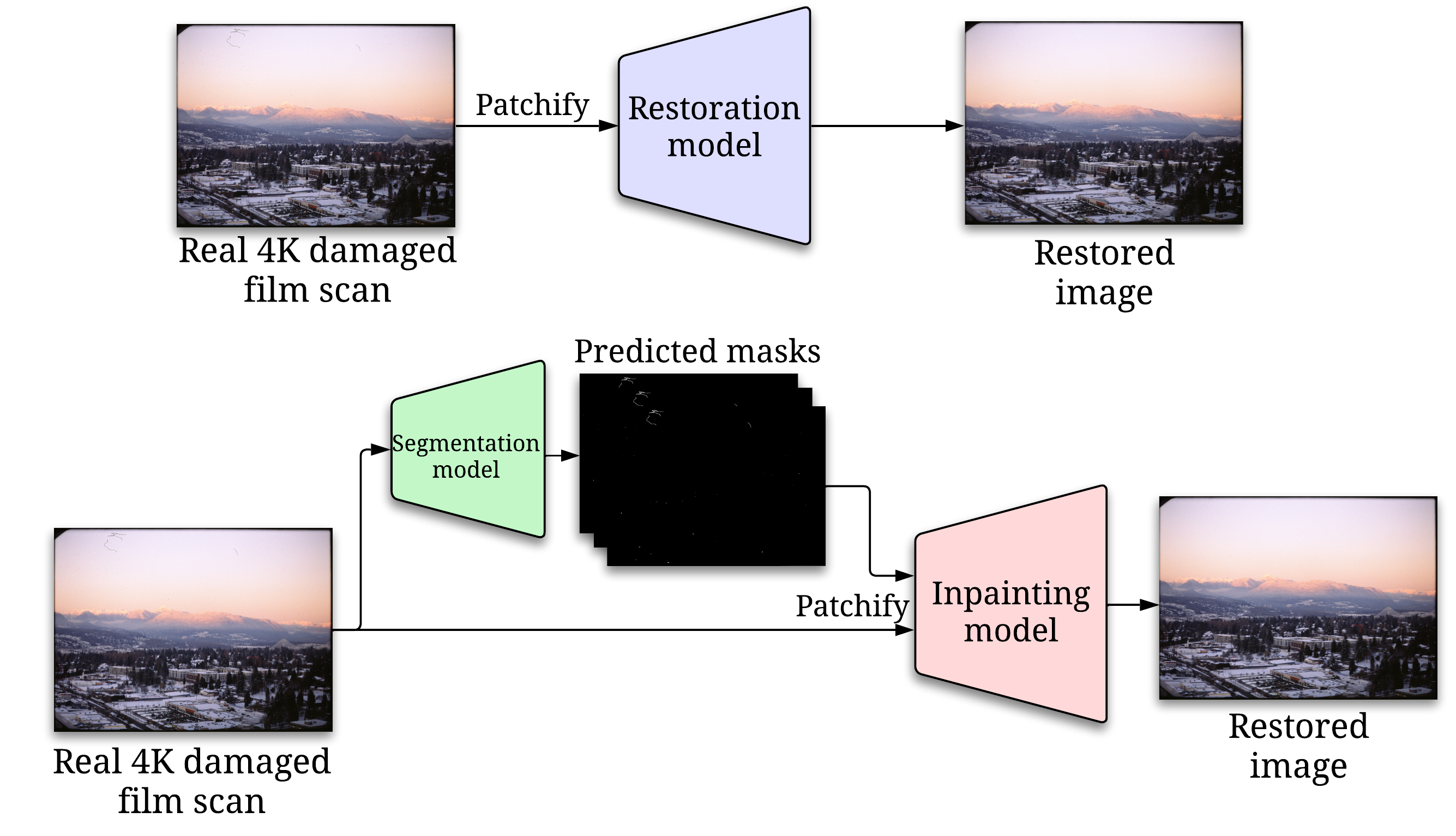}\vspace{-8pt}}
  \vspace{-8pt}
  \caption{Overview of (a) training and (b) evaluation workflows for restoration, segmentation and inpainting models. We retrain existing restoration models on our synthetic data where relevant, and train an additional segmentation model to adapt state-of-the-art inpainting methods to the film artefact restoration task.}
  \label{fig1:overview-evaluation}
\end{figure*}

\subsection{Human validation of generated artefact damage}\label{subsec:validating-generated-damage}

We conduct two user perceptual surveys to verify that our damage overlay generation process yields realistic results. We discuss the study design and summarise the results below.
In both cases, we also ask each participant to self-report their familiarity level with analogue film and its associated artefacts, as `Not familiar at all', `Somewhat familiar' or `Very familiar'. 

\paragraph{Real vs.~synthetic damage.}
In the first survey, we compare real analogue damage to the synthetic damage produced by our model. The generated damage is applied to the ground truth restored images from our test set. We pair the synthetically damaged image scans with their real damaged versions. Each participant was shown 100 such pairs (see Figure S1 in the supplementary material for examples): 50 simulating negative film damage (i.e., white artefacts), and 50 simulating positive (slide) film damage (i.e. black artefacts), and is asked to choose one image per pair which they believe shows real analogue film damage. The images are cropped to five different target resolutions to reflect the relationship between artefacts and image features of varying size. 

We collected 251 responses to this first survey. We calculate the percentage of pairs for which each participant chose the example with real artefacts. 

We define the `score' of a user as being the overall fraction of pairs for which they correctly selected the real damage.
A score of 50\% is ideal, as it would mean that the user is unable to distinguish real from synthetic damage better than chance. Participant score breakdown is summarsed in Table~\ref{tab:survey1-sumary}.
We find that unfamiliar participants scored on average 51.51\% with standard deviation of 6.88\%, moderately familiar participants scored on average 51.93\% with standard deviation of 7.05\%, and very familiar participants scored on average 51.43\% with standard deviation of 7.0\%.
Thus, all groups of users found our synthetically damaged images to be indistinguishable from original damaged scans, and there was no statistically-significant variation among the groups.

\paragraph{Ours vs. Ivanova et al.~\cite{visapp22}.}
In the second survey, participants are shown 30 pairs of film scans from the Documerica photographic collection \cite{documerica}. In each pair, one version of the image is damaged using the earlier approach of Ivanova et al.~\cite{visapp22}, whereas the other image is damaged using our proposed approach.
Participants were asked which image in the pair showed more realistic damage.

We collected 78 responses for the second survey. 
In this survey, unfamiliar participants preferred our damage on average in 63.88\% of the examples with standard deviation of 23.17\%, moderately familiar participants preferred our damage in 73.33\% of the examples with standard deviation of 21.13\%, and very familiar participants preferred our damage in 75.58\% of the examples with standard deviation of 21.71\%. These results indicate that all groups found our synthetic damage to be more realistic than that proposed in the prior work \cite{visapp22}; expert participants favored our damage the most out of the three groups.

\begin{figure}[ht]\centering
\begin{tikzpicture}[zoomboxarray, zoomboxes below, zoomboxarray inner gap=0.5cm, zoomboxarray columns=1, zoomboxarray rows=1]
    \node [image node] { \includegraphics[width=0.32\linewidth]{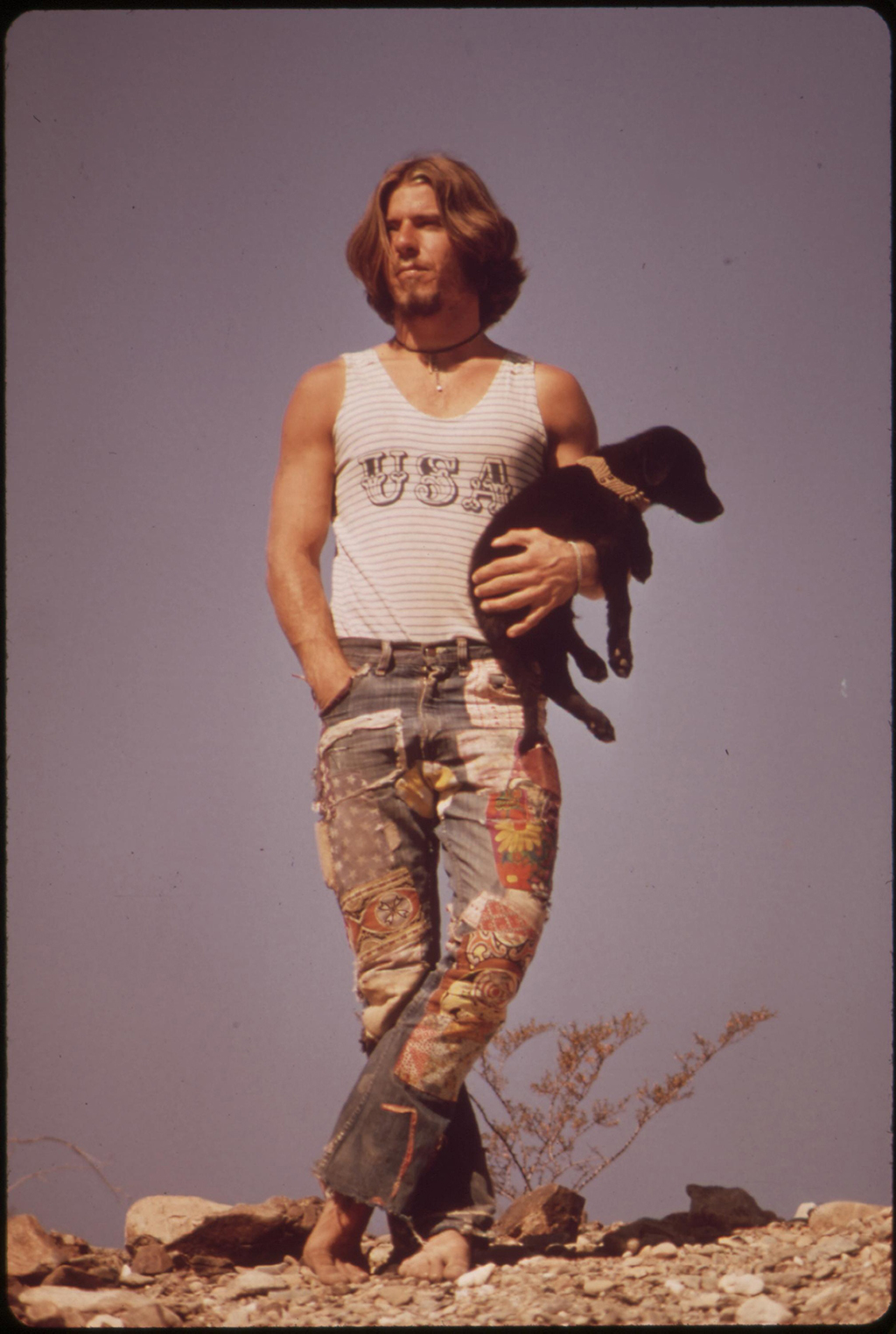} };
    \zoombox[magnification=5]{0.74,0.63}
\end{tikzpicture}
\begin{tikzpicture}[zoomboxarray, zoomboxes below, zoomboxarray inner gap=0.5cm, zoomboxarray columns=1, zoomboxarray rows=1]
    \node [image node] { \includegraphics[width=0.32\linewidth]{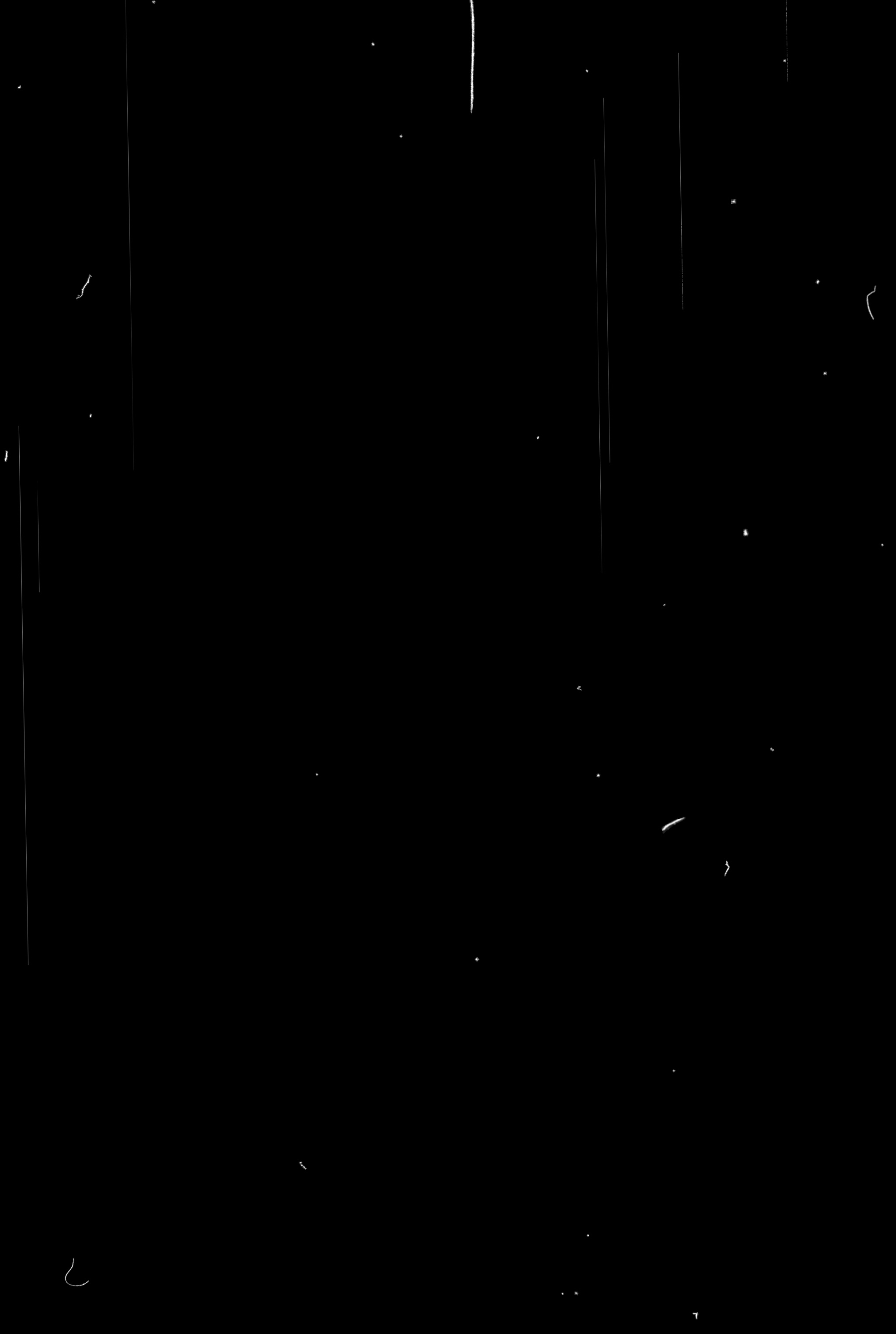} };
    \zoombox[magnification=5]{0.74,0.63}
\end{tikzpicture}
\begin{tikzpicture}[zoomboxarray, zoomboxes below, zoomboxarray inner gap=0.5cm, zoomboxarray columns=1, zoomboxarray rows=1]
    \node [image node] { \includegraphics[width=0.32\linewidth]{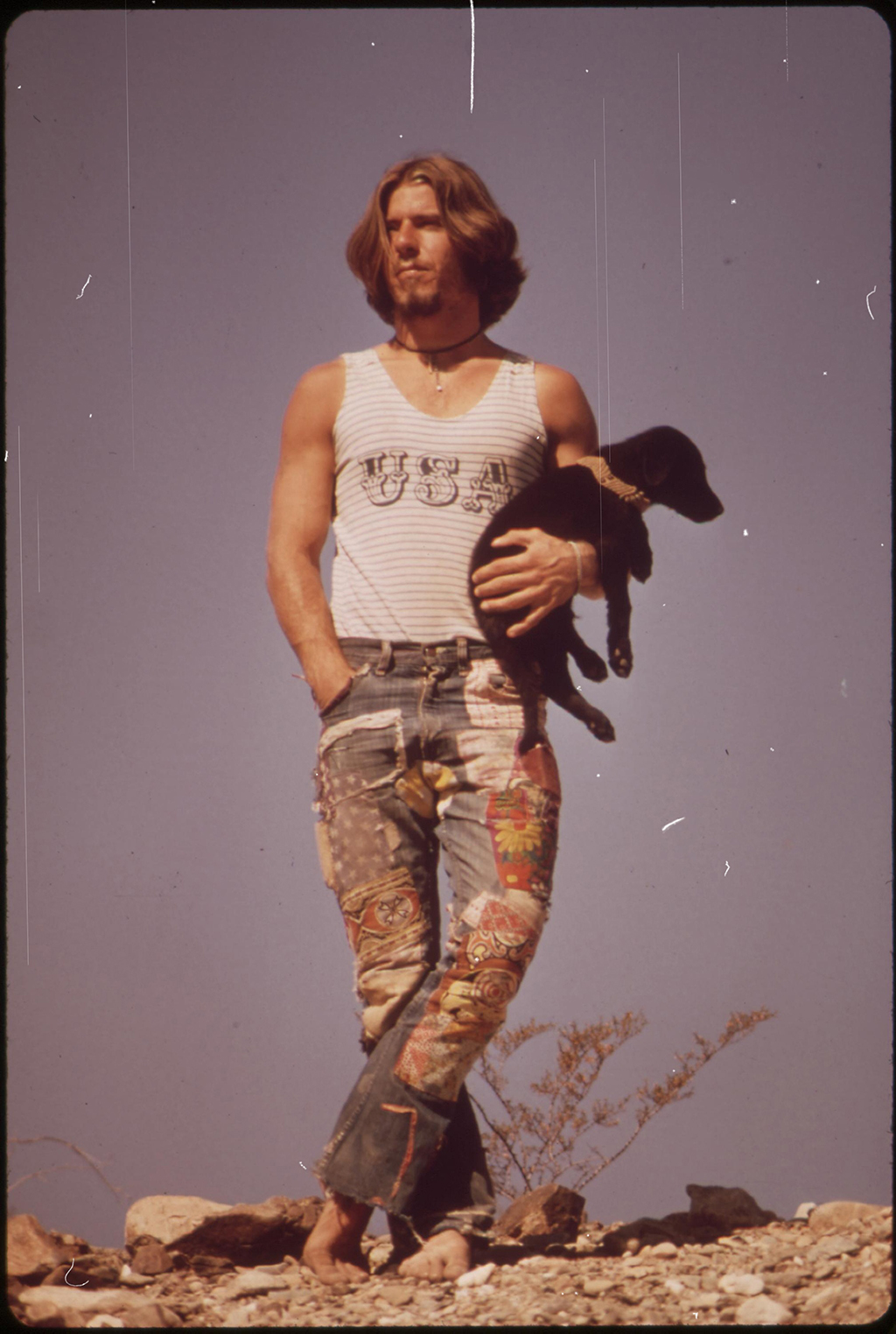} };
    \zoombox[magnification=5]{0.74,0.63}
\end{tikzpicture}
\caption{High resolution image, generated damage overlay and resulting synthetically damaged image.}
\label{fig10:damaged-documerica-example}
\end{figure}
\vspace{-10pt}
\section{Experiments}
\label{sec:comparison}

Equipped with our damage simulator (Section~\ref{sec:modelling-damage}) and our expertly curated test set of real analogue damage with hand-restored ground-truths (Section~\ref{sec:gt-dataset}), we perform two sets of experiments. 
In Section~\ref{subsec:segmentation-comparison}, we directly compare our damage synthesis pipeline with those from two prior works \cite{iizuka2019deepremaster,wan2020bringing}. In 
Section~\ref{subsec:restoration-comparison}, we evaluate several approaches on our final goal of film restoration. The following sections discuss the chosen approaches (and how we adapt them to high resolution data), describe our experimental setup, and report the results of our experiments.

\paragraph{Evaluation metrics for restoration.}
Since our evaluation set has ground-truth restored images (Section~\ref{sec:gt-dataset}), we can evaluate restoration quality directly using standard image similarity metrics.
Specifically, we use peak signal-to-noise ratio (PSNR), structural similarity (SSIM) \cite{ssim}, and the learnt perceptual metric LPIPS \cite{zhang18cvpr}. These are calculated over full frames (not individual patches), and we report the average over the evaluation set. 
\paragraph{Evaluation metrics for segmentation.}
We obtain approximate ground-truth segmentations of the authentically damaged dataset by subtracting the damaged images from the restored ones and binarising. We use the standard image segmentation metrics intersection-over-union (IoU) and F1 score.
Since these metrics are sensitive to exact pixel-value overlap, which in turn is influenced by the choice of binarisation threshold, we test several thresholds and choose the one which maximises the scores obtained by the baselines. 
We also borrow a popular point-cloud comparison metric, the earth-mover's distance (EMD)~\cite{achlioptas2018learning, fan2017point}, to provide an additional measure which is less sensitive to exact overlap, but more accurately compares proximity of long, narrow features such as hairs and scratches.

\setlength{\tabcolsep}{0.3pt}
\begin{figure*}[hbtp]
  \centering
  \begin{tabular}[c]{ccccc}
   \rotatebox[origin=c]{90}{\makecell{\textbf{Input}: 4K film scan \\ with authentic damage.}} &
    \begin{subfigure}[c]{0.31\textwidth}
        \begin{tikzpicture}[spy using outlines={circle,blue,magnification=3,size=1.8cm, connect spies}]
        \node {\includegraphics[width=\linewidth]{figures/damaged_input.jpg}};
        \spy on (-1.55,1.65) in node [left] at (1.8,1.1);
        \end{tikzpicture}
    \end{subfigure}&
    \begin{subfigure}[c]{0.31\textwidth}
        \begin{tikzpicture}[spy using outlines={circle,blue,magnification=3,size=1.8cm, connect spies}]
        \node {\includegraphics[width=\linewidth]{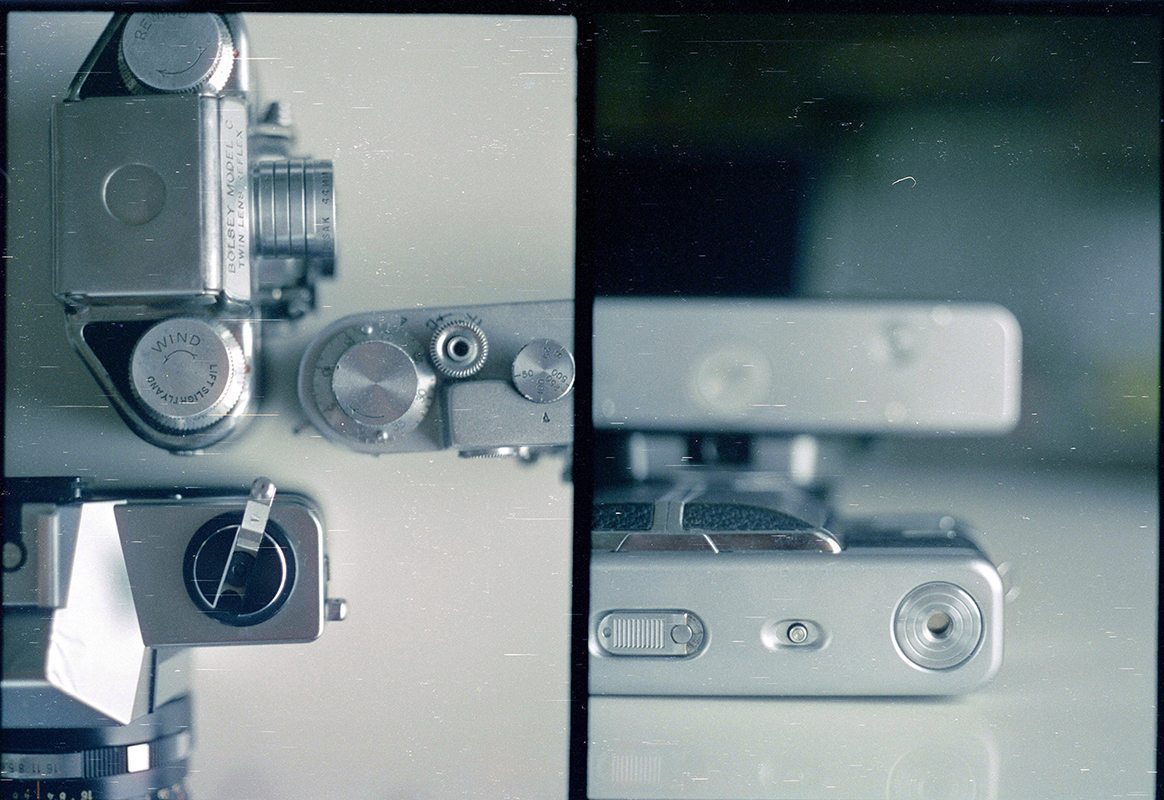}};
        \spy on (-1.75,-0.4) in node [left] at (1.8,1.0);
        \end{tikzpicture}
    \end{subfigure}&
    \begin{subfigure}[c]{0.31\textwidth}
        \begin{tikzpicture}[spy using outlines={circle,blue,magnification=3,size=1.8cm, connect spies}]
        \node {\includegraphics[width=\linewidth]{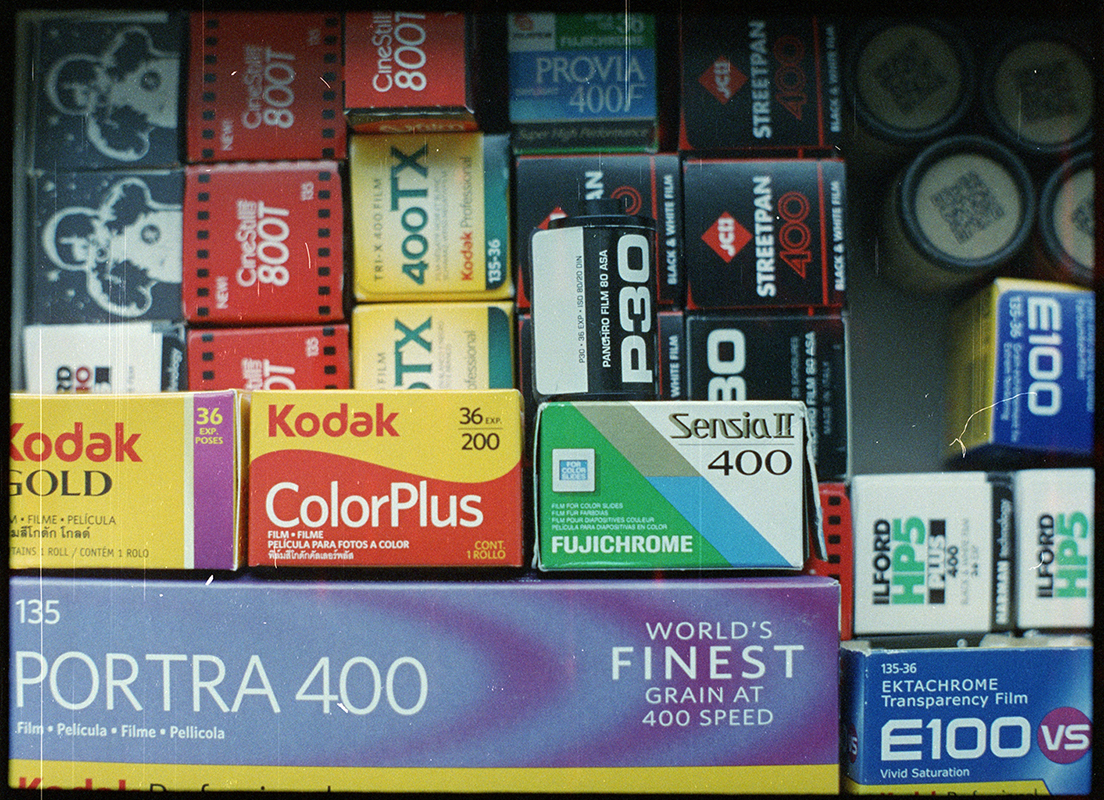}};
        \spy on (2.15,-0.1) in node [left] at (1.8,1.1);
        \end{tikzpicture}
    \end{subfigure}\\
    
    \rotatebox[origin=c]{90}{\makecell{\textbf{Segmentation} from U-Net\\ trained on our synthetically\\ damaged data.}} &
    \begin{subfigure}[c]{0.31\textwidth}
        \begin{tikzpicture}[spy using outlines={circle,yellow,magnification=3,size=1.8cm, connect spies}]
        \node {\includegraphics[width=\linewidth]{figures/our_mask.png}};
        \spy on (-1.55,1.65) in node [left] at (1.8,1.1);
        \end{tikzpicture}
    \end{subfigure}&
    \begin{subfigure}[c]{0.31\textwidth}
        \begin{tikzpicture}[spy using outlines={circle,yellow,magnification=3,size=1.8cm, connect spies}]
        \node {\includegraphics[width=\linewidth]{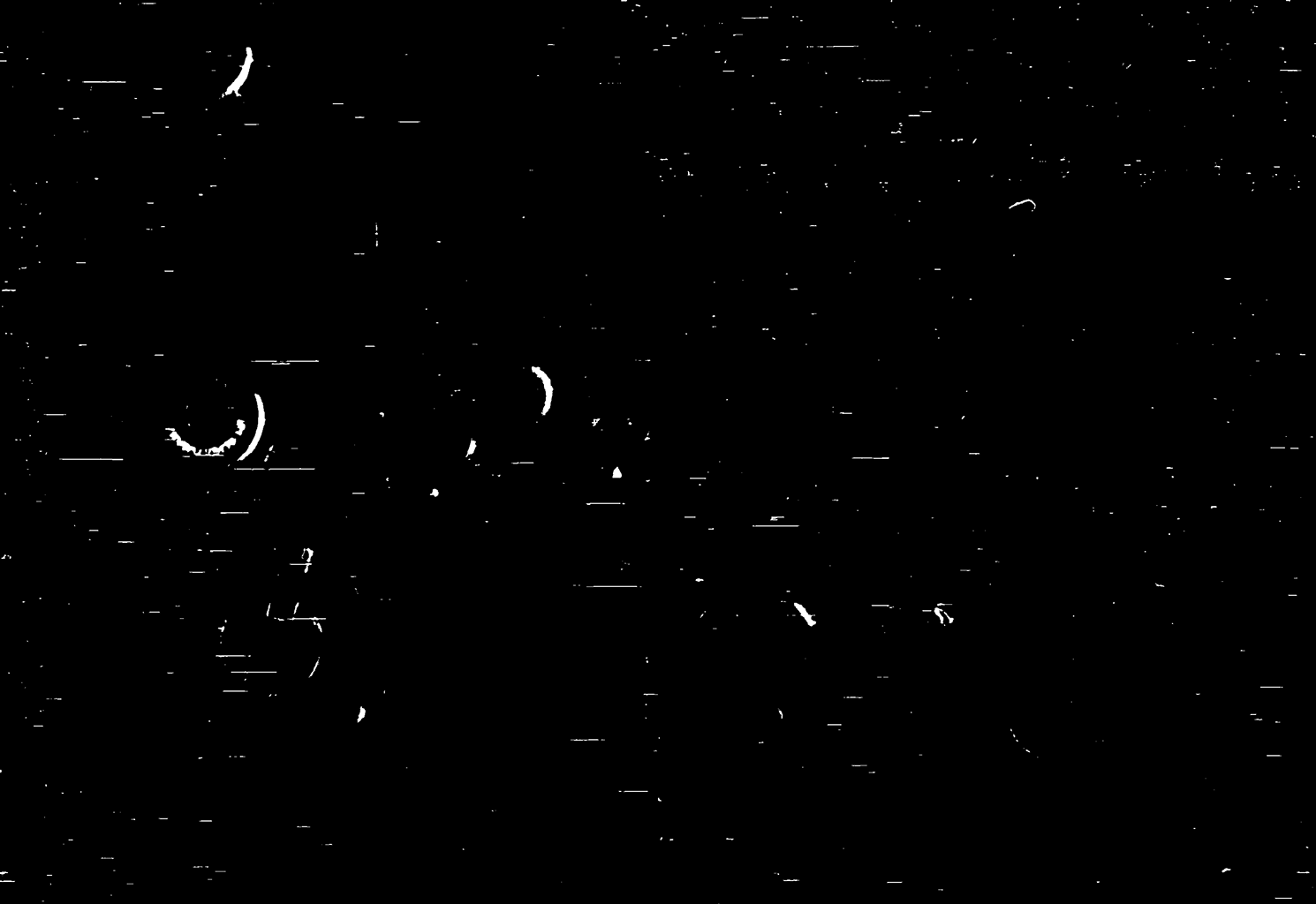}};
        \spy on (-1.75,-0.4) in node [left] at (1.8,1.0);
        \end{tikzpicture}
    \end{subfigure}&
    \begin{subfigure}[c]{0.31\textwidth}
        \begin{tikzpicture}[spy using outlines={circle,yellow,magnification=3,size=1.8cm, connect spies}]
        \node {\includegraphics[width=\linewidth]{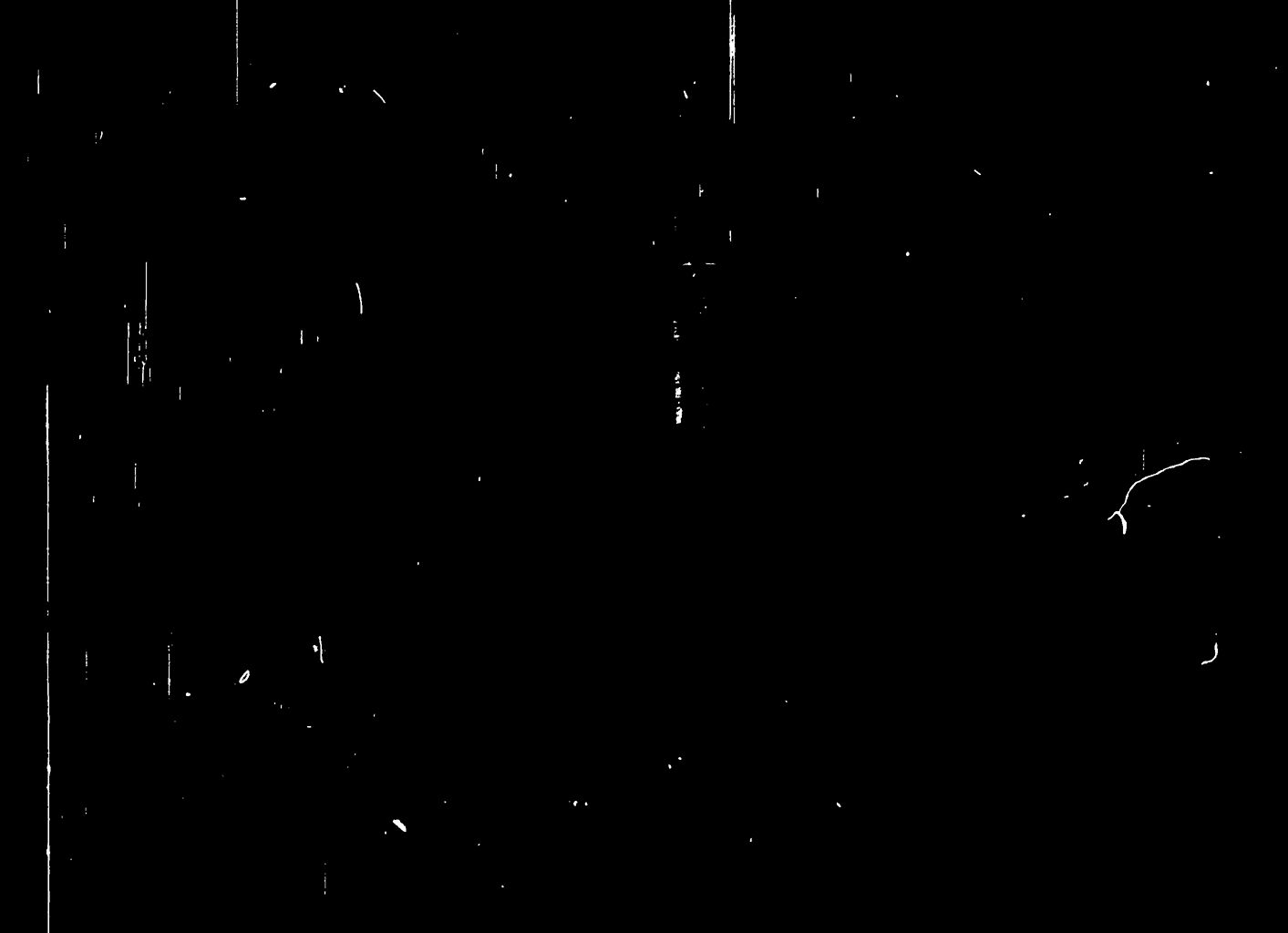}};
        \spy on (2.15,-0.1) in node [left] at (1.8,1.1);
        \end{tikzpicture}
    \end{subfigure}\\

    \rotatebox[origin=c]{90}{\makecell{\textbf{Segmentation} from U-Net\\ trained on damage overlays\\ by DeepRemaster~\cite{iizuka2019deepremaster}.}} &
    \begin{subfigure}[c]{0.31\textwidth}
        \begin{tikzpicture}[spy using outlines={circle,yellow,magnification=3,size=1.8cm, connect spies}]
        \node {\includegraphics[width=\linewidth]{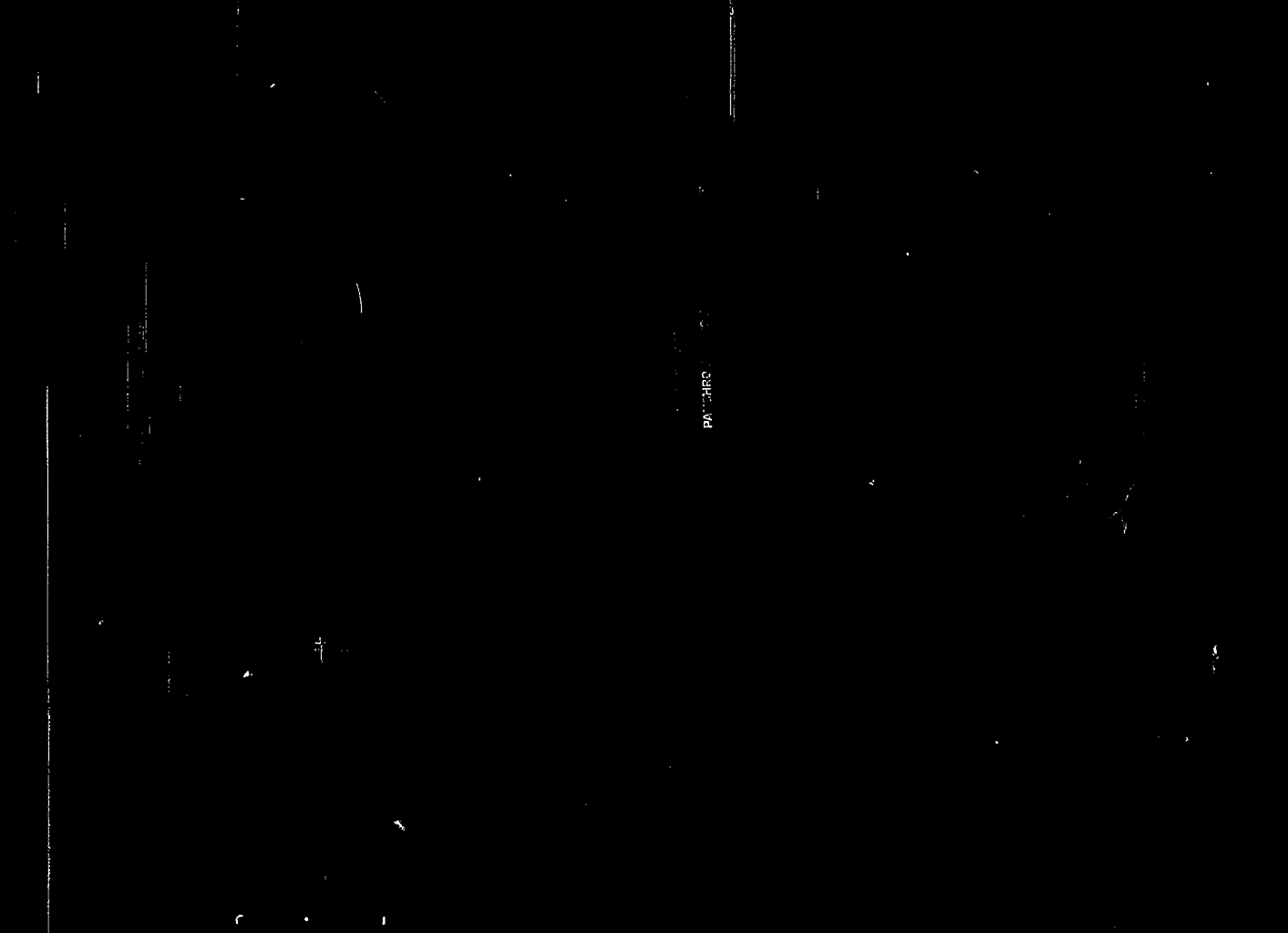}};
        \spy on (-1.55,1.65) in node [left] at (1.8,1.1);
        \end{tikzpicture}
    \end{subfigure}&
    \begin{subfigure}[c]{0.31\textwidth}
        \begin{tikzpicture}[spy using outlines={circle,yellow,magnification=3,size=1.8cm, connect spies}]
        \node {\includegraphics[width=\linewidth]{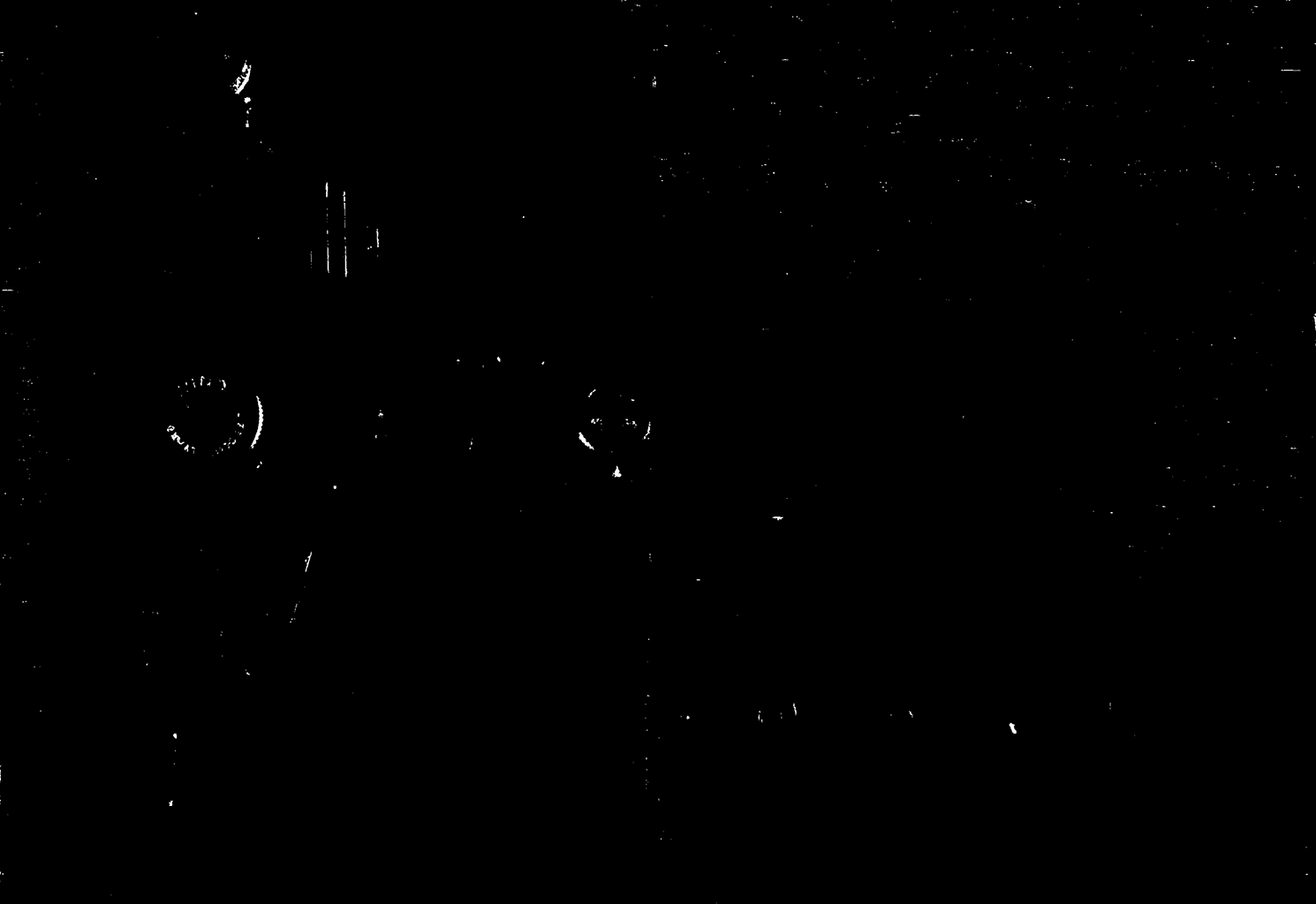}};
        \spy on (-1.75,-0.4) in node [left] at (1.8,1.0);
        \end{tikzpicture}
    \end{subfigure}&
    \begin{subfigure}[c]{0.31\textwidth}
        \begin{tikzpicture}[spy using outlines={circle,yellow,magnification=3,size=1.8cm, connect spies}]
        \node {\includegraphics[width=\linewidth]{figures/segmentation-deepremaster-cinestill800_half_1.png}};
        \spy on (2.15,-0.1) in node [left] at (1.8,1.1);
        \end{tikzpicture}
    \end{subfigure}\\

    \rotatebox[origin=c]{90}{\makecell{\textbf{Segmentation} from artefact\\ detection module of\\ BOPB~\cite{wan2020bringing}.}} &
    \begin{subfigure}[c]{0.31\textwidth}
    \centering
        \begin{tikzpicture}[spy using outlines={circle,yellow,magnification=3,size=1.8cm, connect spies}]
        \node {\includegraphics[width=\linewidth]{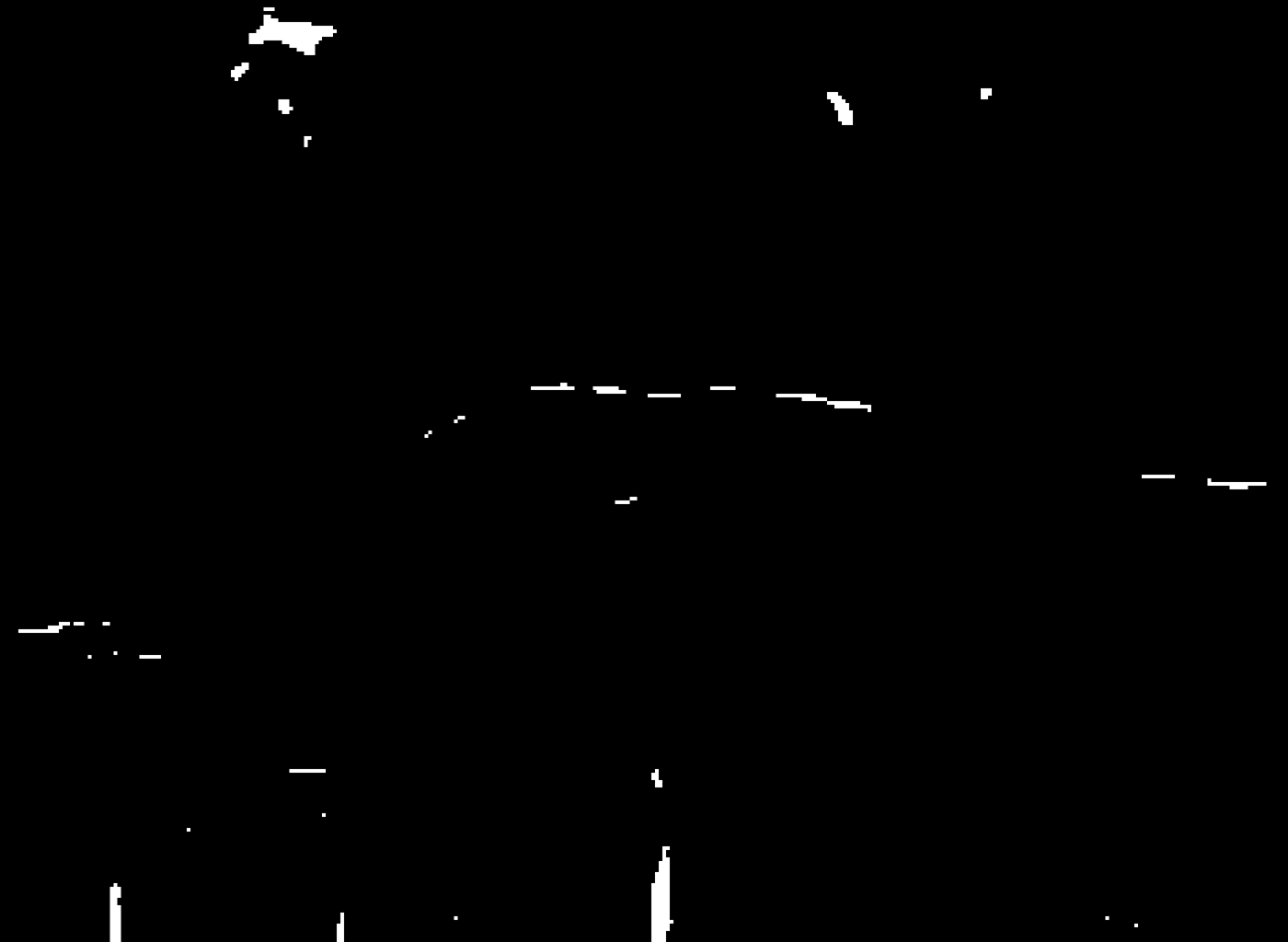}};
        \spy on (-1.55,1.65) in node [left] at (1.8,1.1);
        \end{tikzpicture}
    \end{subfigure}&
    \begin{subfigure}[c]{0.31\textwidth}
        \begin{tikzpicture}[spy using outlines={circle,yellow,magnification=3,size=1.8cm, connect spies}]
        \node {\includegraphics[width=\linewidth]{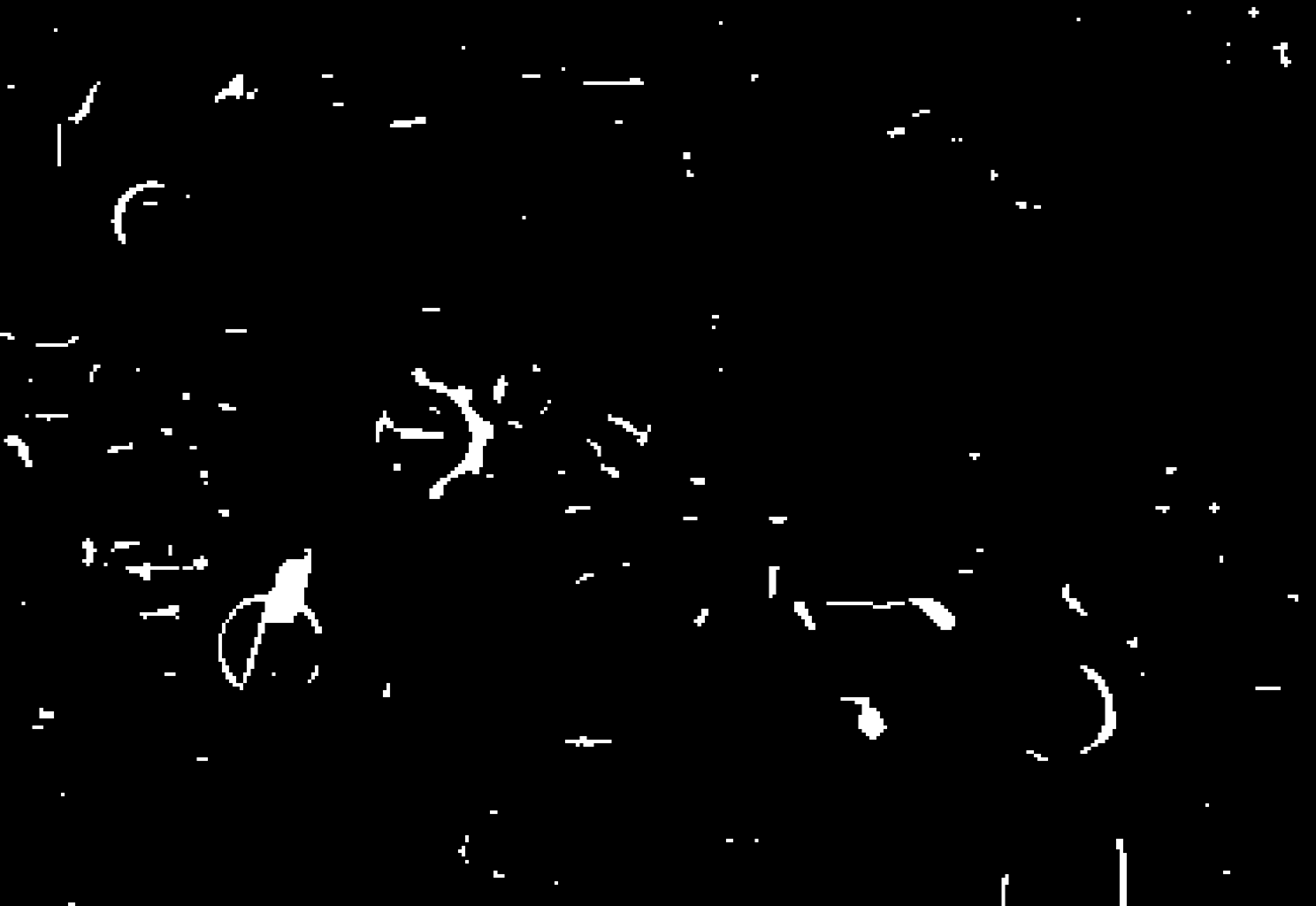}};
        \spy on (-1.75,-0.4) in node [left] at (1.8,1.0);
        \end{tikzpicture}
    \end{subfigure}&
    \begin{subfigure}[c]{0.31\textwidth}
        \begin{tikzpicture}[spy using outlines={circle,yellow,magnification=3,size=1.8cm, connect spies}]
        \node {\includegraphics[width=\linewidth]{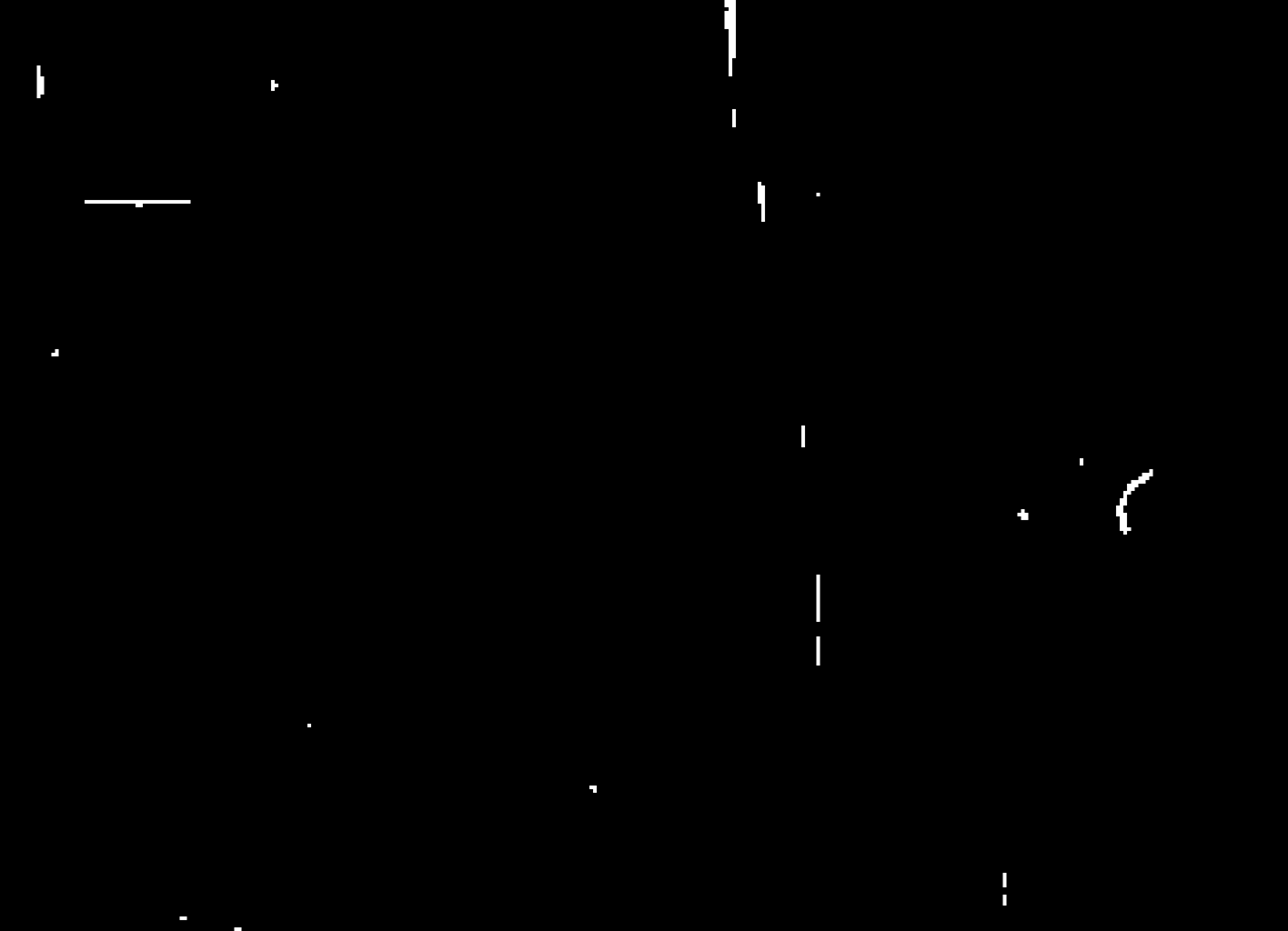}};
        \spy on (2.15,-0.1) in node [left] at (1.8,1.1);
        \end{tikzpicture}
    \end{subfigure}\\

    \rotatebox[origin=c]{90}{\makecell{\textbf{Approximate ground truth}:\\ binarised difference of damaged\\ and manually restored scans.}} &
    \begin{subfigure}[c]{0.31\textwidth}
    \centering
        \begin{tikzpicture}[spy using outlines={circle,yellow,magnification=3,size=1.8cm, connect spies}]
        \node {\includegraphics[width=\linewidth]{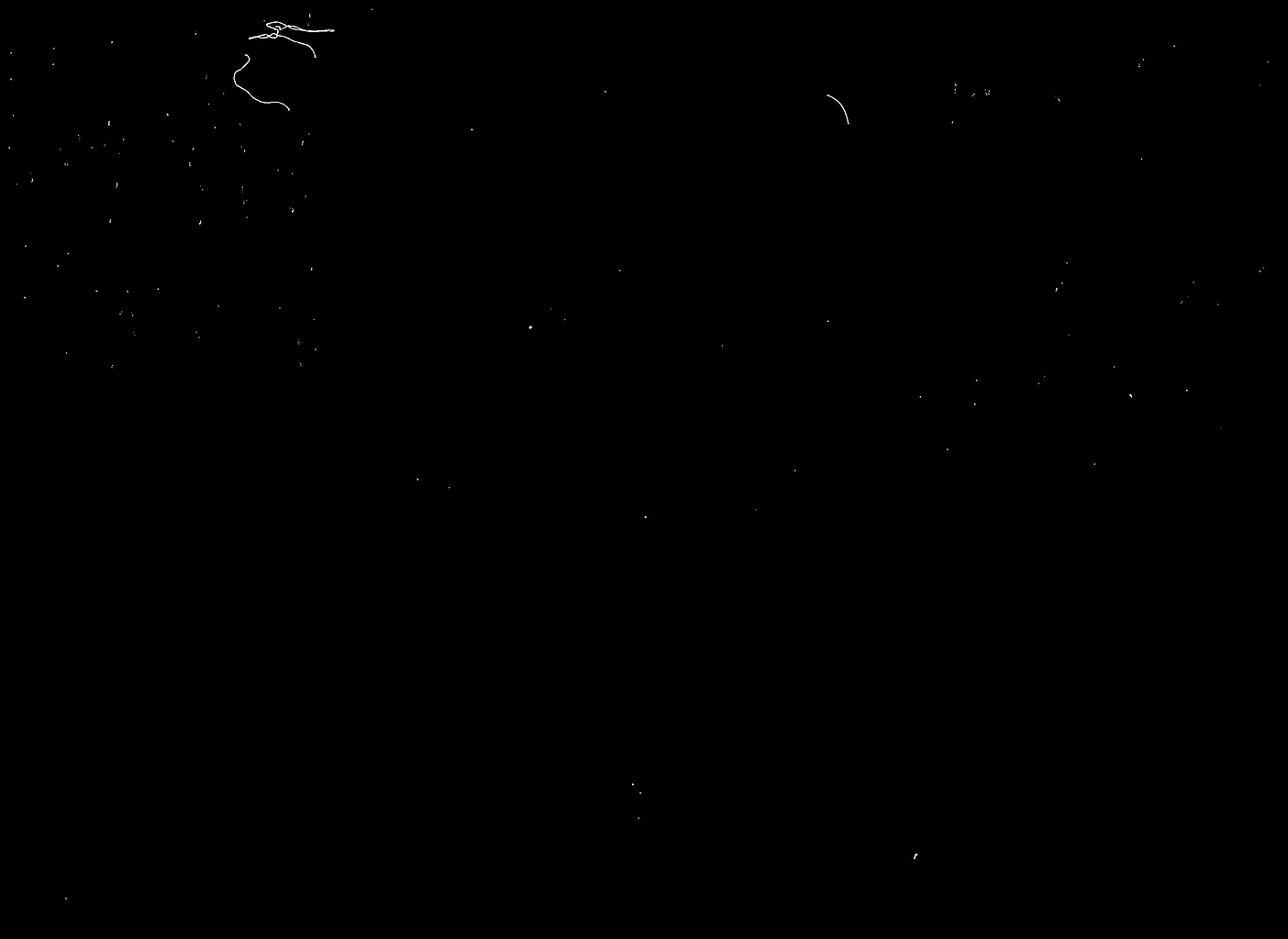}};
        \spy on (-1.55,1.65) in node [left] at (1.8,1.1);
        \end{tikzpicture}
    \end{subfigure}&
    \begin{subfigure}[c]{0.31\textwidth}
        \begin{tikzpicture}[spy using outlines={circle,yellow,magnification=3,size=1.8cm, connect spies}]
        \node {\includegraphics[width=\linewidth]{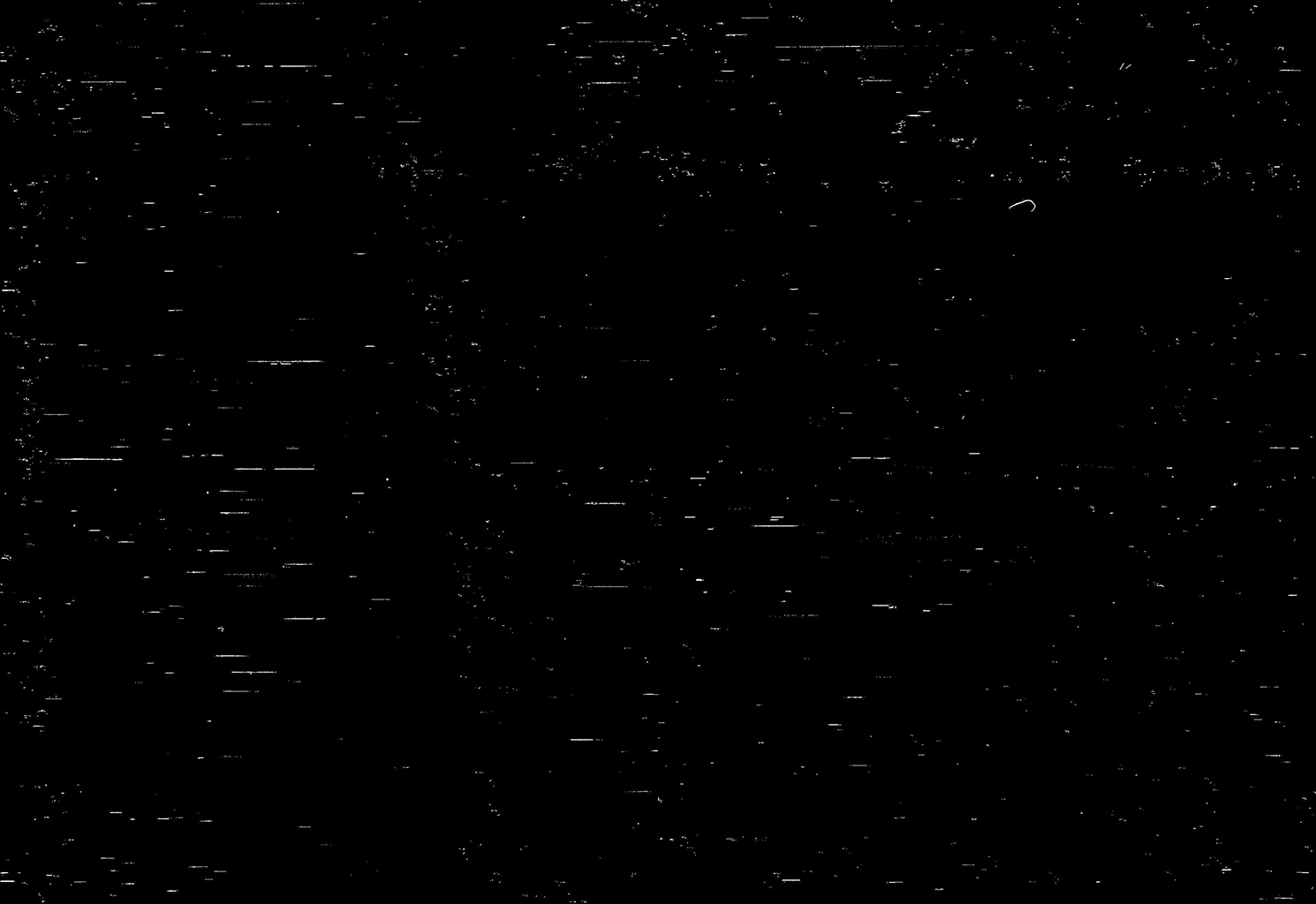}};
        \spy on (-1.75,-0.4) in node [left] at (1.8,1.0);
        \end{tikzpicture}
    \end{subfigure}&
    \begin{subfigure}[c]{0.31\textwidth}
        \begin{tikzpicture}[spy using outlines={circle,yellow,magnification=3,size=1.8cm, connect spies}]
        \node {\includegraphics[width=\linewidth]{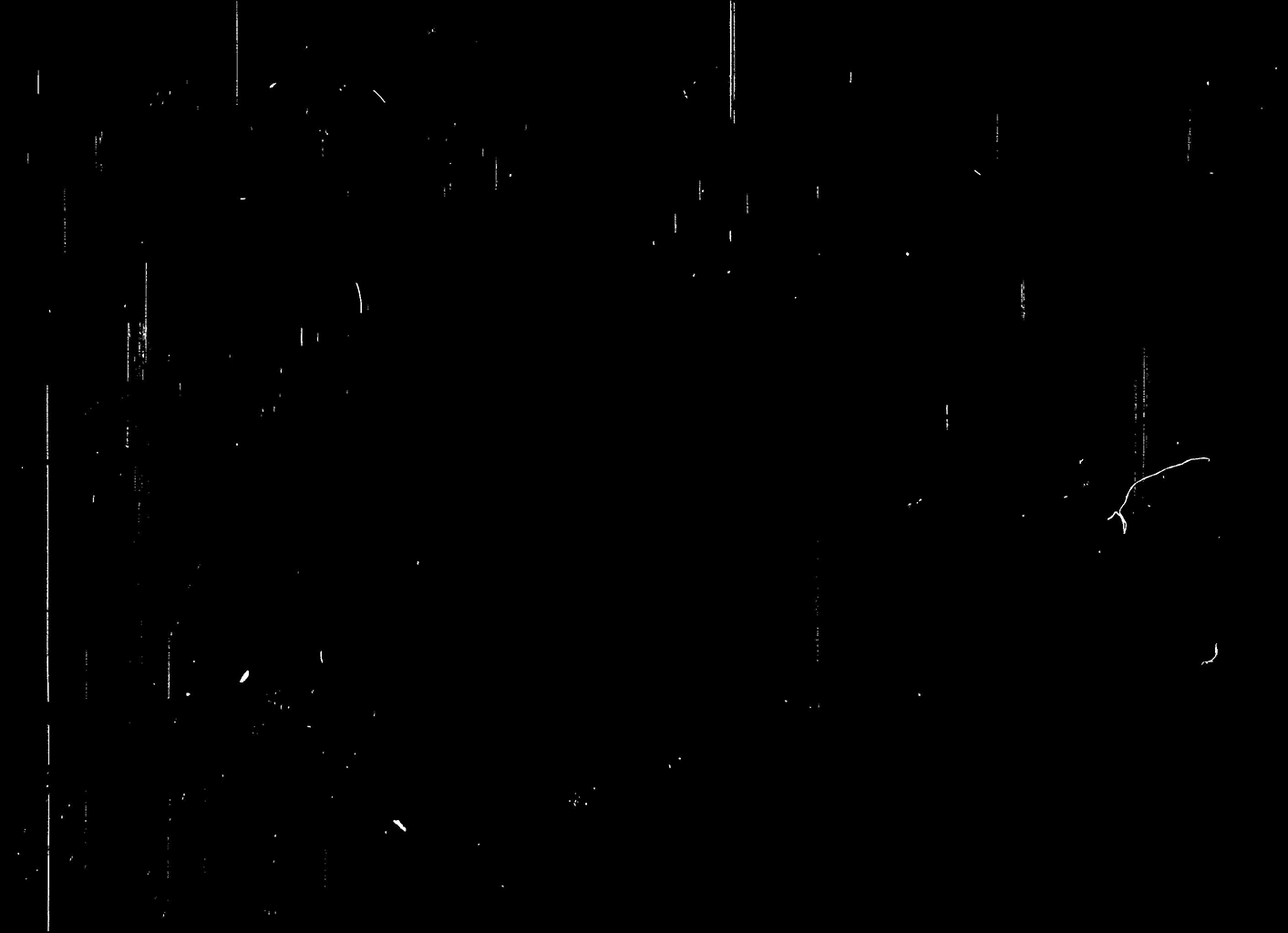}};
        \spy on (2.15,-0.1) in node [left] at (1.8,1.1);
        \end{tikzpicture}
    \end{subfigure}\\
    
  \end{tabular}    
  \caption{Qualitative comparison of segmentations of artefacts from our authentic damage dataset. }
  \label{fig:qualitative-comparison-segmentations}
\end{figure*}
\setlength{\tabcolsep}{6pt}

\paragraph{Synthetically damaged training data.}\label{subsec:synthetically-damaged-dataset}
We generate synthetic damage overlays using our proposed approach (Section~\ref{sec:modelling-damage}) for 6228 clean 4K image scans from the Documerica collection \cite{documerica} (see Figure~\ref{fig10:damaged-documerica-example} for an example).
We render the synthetic damage either as negative or positive artefacts (corresponding to damage on film negatives or developed slides respectively), with equal probability for each image.
The overlays are also binarised to produce segmentation maps indicating where damage was added.
We split the set of 6228 images into training and validation sets in the ratio of 9:1.
For the segmentation experiment, we use a variant of this dataset produced by damaging the same set of Documerica images with damage overlays as provided by DeepRemaster~\cite{iizuka2019deepremaster}.
These datasets are used to retrain restoration models where relevant, as well as to train a segmentation model to detect artefact damage in our real artefact damage evaluation set. The data is available at \URL{www.doi.org/10.6084/m9.figshare.21815844}. 

\subsection{Comparison against existing damage simulations\label{subsec:segmentation-comparison}}

We train two U-Net~\cite{ronneberger2015u} segmentation networks to detect artefact damage in real damaged film scans -- one using damage generated by our model, 
and one using the damage provided by DeepRemaster~\cite{iizuka2019deepremaster}.
We train both for 20 epochs on $256\times256$ crops of the corresponding damaged training set with learning rate $10^{-3}$ and a batch size of 16.
We compare the predictions from the two segmentation networks directly to those from the artefact segmentation module of BOPB~\cite{wan2020bringing} as our second baseline.

\begin{table}[h]
\begin{tabular}{@{}llll@{}}
\toprule
\textbf{Segmentation model training} & \textbf{IoU~$\uparrow$} & \textbf{F1~$\uparrow$} & \textbf{EMD~$\downarrow$} \\ \midrule
\textit{Our damage}                                                     & \round{3}{0.256437}           & \round{3}{0.408197}                        & \round{3}{0.007351}                                            \\
\textit{DeepRemaster~\cite{iizuka2019deepremaster} damage}                                                     & \round{3}{0.003217}           & \round{3}{0.006414}                        & \round{3}{0.016007}                                           \\
\textit{BOPB~\cite{wan2020bringing} pre-trained module}     
        & \round{3}{0.009613}            & \round{3}{0.019042}               & \round{3}{0.016382}                                                     \\
 \bottomrule
\end{tabular}
\caption{Comparison between the predicted segmentations for the evaluated models trained on different simulated damage.}
\label{tab:segmentation-quantitative-summary}
\vspace{-10pt}
\end{table}

\paragraph{Results.}
Segmentations obtained from the U-Net trained on our synthetically damaged data are of much higher quality than those obtained from the same model trained on the damage from DeepRemaster, as well as those obtained from the segmentation module of BOPB~\cite{wan2020bringing}, as shown in Figure~\ref{fig:qualitative-comparison-segmentations}. Quantitatively, the model trained on our artefacts again outperforms BOPB and DeepRemaster, achieving lower EMD and higher IoU and F1 scores, summarised in Table~\ref{tab:segmentation-quantitative-summary}.

\subsection{Comparison of restoration models}\label{subsec:restoration-comparison}

We compare a diverse set of approaches to artefact restoration.
First, we select three methods that directly perform damage restoration:
\begin{itemize}
    \item \textit{Bringing Old Photos Back to Life} (BOPB) \cite{wan2020bringing}, an approach specifically targeting analogue damage, for which the authors have provided pre-trained weights. We compare variants using our segmentation model vs.~theirs, and with fully-convolutional vs.~patch-wise processing.

    \item A restoration U-Net trained with perceptual loss \cite{visapp22}, another approach which specialises in film artefact removal, for which the pre-trained weights are available. We also test a variant re-trained on crops of our synthetic 4K dataset, instead of the $256\times256$ downsampled images used in the original. Lastly, we test a variant re-trained on the same crops of our synthetic dataset, damaged with overlays from DeepRemaster~\cite{iizuka2019deepremaster}.

    \item Adobe Photoshop's \textit{Dust \& Scratch filter}, which is a commonly-used off-the-shelf solution for artefact restoration.
    
\end{itemize}
Second, we select three state-of-the-art methods for image inpainting. In order to apply these to the restoration task, we also train a model for segmenting artefacts (see below), which is used to determine which pixels must be inpainted:
\begin{itemize}
    
    \item \textit{LaMa} \cite{suvorov2022resolution}, a state-of-the-art approach for high resolution irregular hole inpainting, with a pre-trained model available.
    
    \item \textit{Stable Diffusion} \cite{rombach2021highresolution}, popular for its state-of-the-art text-to-image generation abilities, also supports image inpainting.
    
    \item \textit{RePaint} \cite{lugmayr2022repaint}, another diffusion approach focusing on inpainting, proposing a resampling technique which conditions the inpainting on a pre-trained diffusion model.
\end{itemize}
Finally, we consider two blind inpainting approaches:
\begin{itemize}
    \item \textit{Blind Visual Motif Removal From a Single Image} (BVMR) \cite{hertz2019blind}, which simultaneously detects the pixels representing the visual motif to be inpainted, and synthesises new content for the affected pixels. The provided pre-trained model is for semi-transparent emoji watermark removal, which we re-train for the artefact restoration task.
    \item \textit{Restormer} \cite{Zamir_2022_CVPR}, an approach which achieves state-of-the-art results in the tasks of blind denoising and deraining, and is applicable to large resolution inputs. Out of several models provided, we find that the closest to the task of film artefact removal is the one trained for deraining.
\end{itemize}




\paragraph{Pre-segmentation of artefacts for inpainting methods.}
LaMa, RePaint and Stable Diffusion rely on masks to be supplied for the inpainting task. As they cannot identify artefacts automatically, they cannot be applied to the film restoration task out-of-the-box.
To remedy this, we employ the segmentation U-Net trained on our synthetically damaged data (described in Section~\ref{subsec:segmentation-comparison}) to predict damage masks on the test set of authentically damaged data.
When evaluating LaMa, Stable Diffusion, and RePaint on damage restoration, we first pass the images through the trained segmentation U-Net, then inpaint the regions it indicates are damaged (i.e.~we use the output of the segmentation model as the mask input for the inpainting model).
In addition, while Wan et al.'s \cite{wan2020bringing} method predicts its own damage masks, we also evaluate a variant using the masks from our segmentation model (shown in Section~\ref{subsec:segmentation-comparison} to produce artefact segmentations of superior quality).

\paragraph{Processing high resolution scans.}
Most of the models are designed to operate on $256\times256$ pixel inputs.
We adapt these to process the 4K image scans in our test set without downsampling them first.
During inference, we process the images in our dataset in patches of size $256\times256$, and we stitch back the predictions using 50\% overlap and a smooth blending function, as proposed by Pielawski \& W{\"a}hlby \cite{pielawski2020introducing}. For the inpainting models which require masks, we increase efficiency by only processing the patches for which the corresponding mask indicates the presence of artefacts; if no artefacts have been detected in the patch, we simply copy the input patch to the output tensor to be stitched with the rest of the restored image.
This does not apply to the Photoshop Dust \& Scratch filter, nor to LaMa, both of which can natively process 4K images.






\setlength{\tabcolsep}{0.5pt}
\begin{figure*}[hbtp]
  \begin{subfigure}[t]{.325\textwidth}
    \centering
        \begin{tikzpicture}[spy using outlines={circle,blue,magnification=3,size=1.8cm, connect spies}]
        \node {\includegraphics[width=\linewidth]{figures/damaged_input.jpg}};
        \spy on (-1.63,1.75) in node [left] at (1.8,1.20);
        \end{tikzpicture}
    \caption{\textbf{Input}: 4K film scan with authentic damage}
  \end{subfigure}
    \hfill
  \begin{subfigure}[t]{.325\textwidth}
    \centering
        \begin{tikzpicture}[spy using outlines={circle,yellow,magnification=3,size=1.8cm, connect spies}]
        \node {\includegraphics[width=\linewidth]{figures/our_mask.png}};
        \spy on (-1.63,1.75) in node [left] at (1.8,1.20);
        \end{tikzpicture}
    \caption{\textbf{Artefact Segmentation}: prediction from U-Net trained on synthetically damaged data.}
  \end{subfigure}
    \hfill
  \begin{subfigure}[t]{.325\textwidth}
    \centering
        \begin{tikzpicture}[spy using outlines={circle,yellow,magnification=3,size=1.8cm, connect spies}]
        \node {\includegraphics[width=\linewidth]{figures/bopbtl_mask.png}};
        \spy on (-1.63,1.75) in node [left] at (1.8,1.20);
        \end{tikzpicture}
    \caption{\textbf{Segmentation by BOPB~\cite{wan2020bringing}}.}
  \end{subfigure}
  
   \medskip
  
  \begin{subfigure}[t]{.325\textwidth}
    \centering
        \begin{tikzpicture}[spy using outlines={circle,blue,magnification=3,size=1.8cm, connect spies}]
        \node {\includegraphics[width=\linewidth]{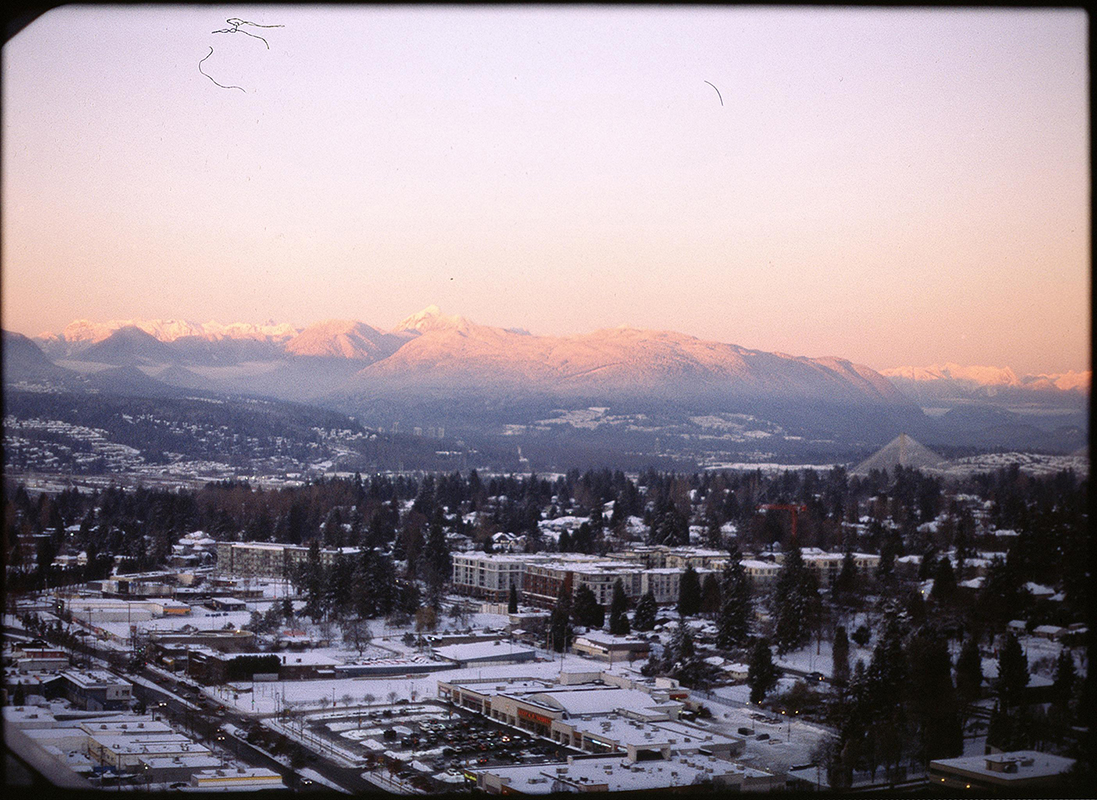}};
        \spy on (-1.63,1.75) in node [left] at (1.8,1.20);
        \end{tikzpicture}
    \caption{\textbf{Restoration by U-Net + perceptual loss~\cite{visapp22}}: using originally provided model weights.}
  \end{subfigure}
    \hfill
  \begin{subfigure}[t]{.325\textwidth}
    \centering
        \begin{tikzpicture}[spy using outlines={circle,blue,magnification=3,size=1.8cm, connect spies}]
        \node {\includegraphics[width=\linewidth]{figures/restoreation_bopbtl_our_masks_no_patches.jpg}};
        \spy on (-1.63,1.75) in node [left] at (1.8,1.20);
        \end{tikzpicture}
    \caption{\textbf{Restoration by BOPB~\cite{wan2020bringing}}: using our segmentation.}
  \end{subfigure}
     \hfill  
  \begin{subfigure}[t]{.325\textwidth}
    \centering
        \begin{tikzpicture}[spy using outlines={circle,blue,magnification=3,size=1.8cm, connect spies}]
        \node {\includegraphics[width=\linewidth]{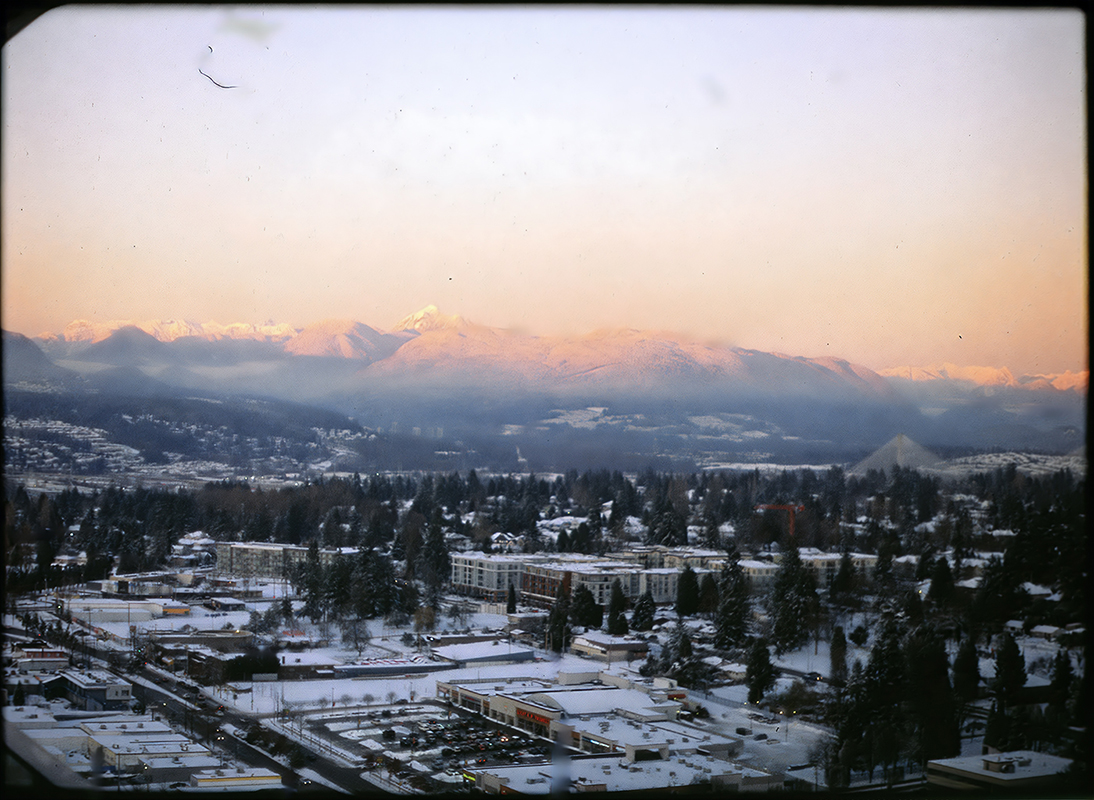}};
        \spy on (-1.63,1.75) in node [left] at (1.8,1.20);
        \end{tikzpicture}
    \caption{\textbf{Restoration by BOPB~\cite{wan2020bringing}}: using their segmentation.}
  \end{subfigure}
   \medskip
  
  \begin{subfigure}[t]{.325\textwidth}
    \centering
        \begin{tikzpicture}[spy using outlines={circle,blue,magnification=3,size=1.8cm, connect spies}]
        \node {\includegraphics[width=\linewidth]{figures/restoration_unet_retrained.jpg}};
        \spy on (-1.63,1.75) in node [left] at (1.8,1.20);
        \end{tikzpicture}
    \caption{\textbf{Restoration by U-Net + perceptual loss~\cite{visapp22}}: retrained on our synthetic damage.}
  \end{subfigure}
    \hfill
  \begin{subfigure}[t]{.325\textwidth}
    \centering
        \begin{tikzpicture}[spy using outlines={circle,blue,magnification=3,size=1.8cm, connect spies}]
        \node {\includegraphics[width=\linewidth]{figures/restoration_lama.jpg}};
        \spy on (-1.63,1.75) in node [left] at (1.8,1.20);
        \end{tikzpicture}
    \caption{\textbf{Restoration by LaMa~\cite{suvorov2022resolution}}: best performing model, using our segmentation.}
  \end{subfigure}
    \hfill
  \begin{subfigure}[t]{.325\textwidth}
    \centering
        \begin{tikzpicture}[spy using outlines={circle,blue,magnification=3,size=1.8cm, connect spies}]
        \node {\includegraphics[width=\linewidth]{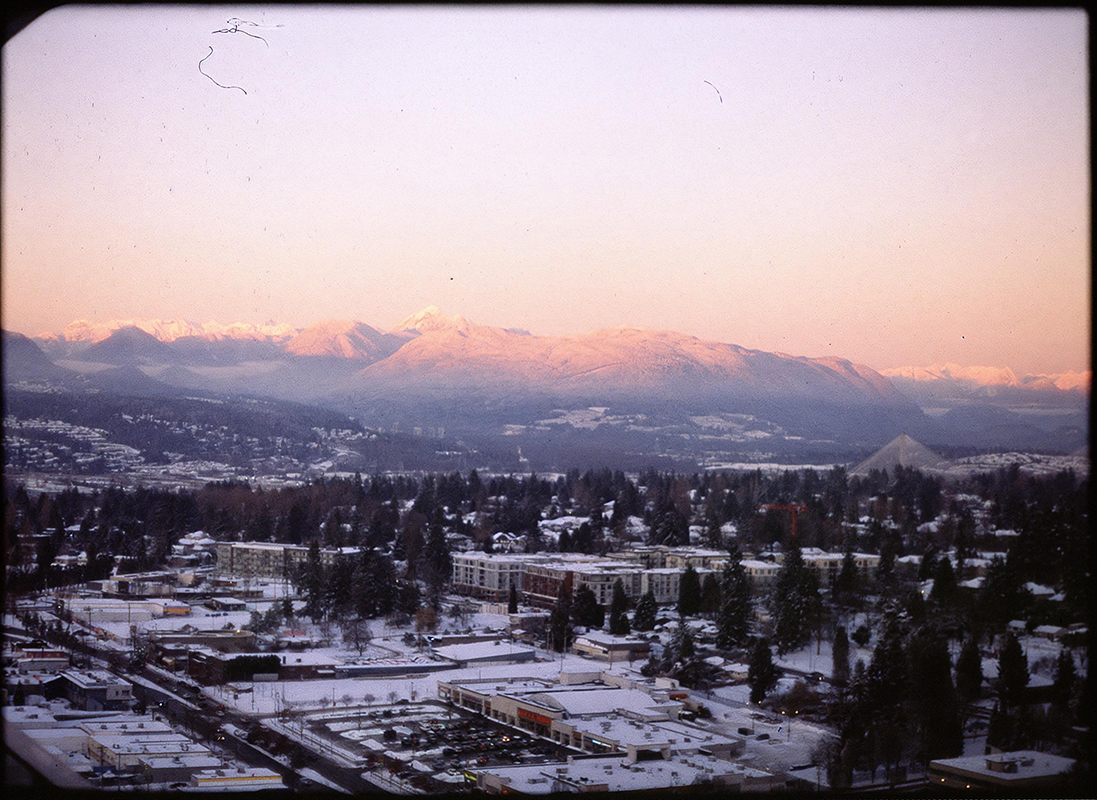}};
        \spy on (-1.63,1.75) in node [left] at (1.8,1.20);
        \end{tikzpicture}
    \caption{\textbf{Restoration by Stable Diffusion~\cite{rombach2021highresolution}}: using our segmentation.}
  \end{subfigure}
 
    \medskip
  
  \begin{subfigure}[t]{.325\textwidth}
    \centering
        \begin{tikzpicture}[spy using outlines={circle,blue,magnification=3,size=1.8cm, connect spies}]
        \node {\includegraphics[width=\linewidth]{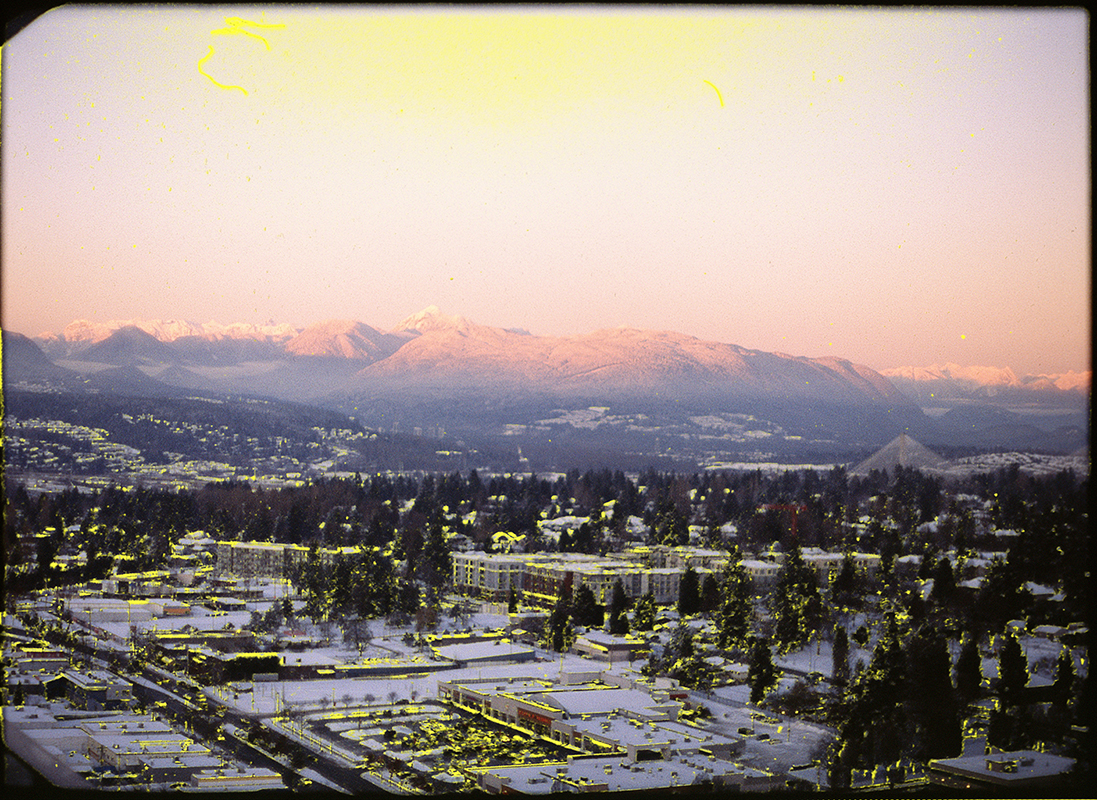}};
        \spy on (-1.63,1.75) in node [left] at (1.8,1.20);
        \end{tikzpicture}
    \caption{\textbf{Restoration by BVMR~\cite{visapp22}}: retrained on our synthetic damage.}
  \end{subfigure}
    \hfill
  \begin{subfigure}[t]{.325\textwidth}
    \centering
        \begin{tikzpicture}[spy using outlines={circle,blue,magnification=3,size=1.8cm, connect spies}]
        \node {\includegraphics[width=\linewidth]{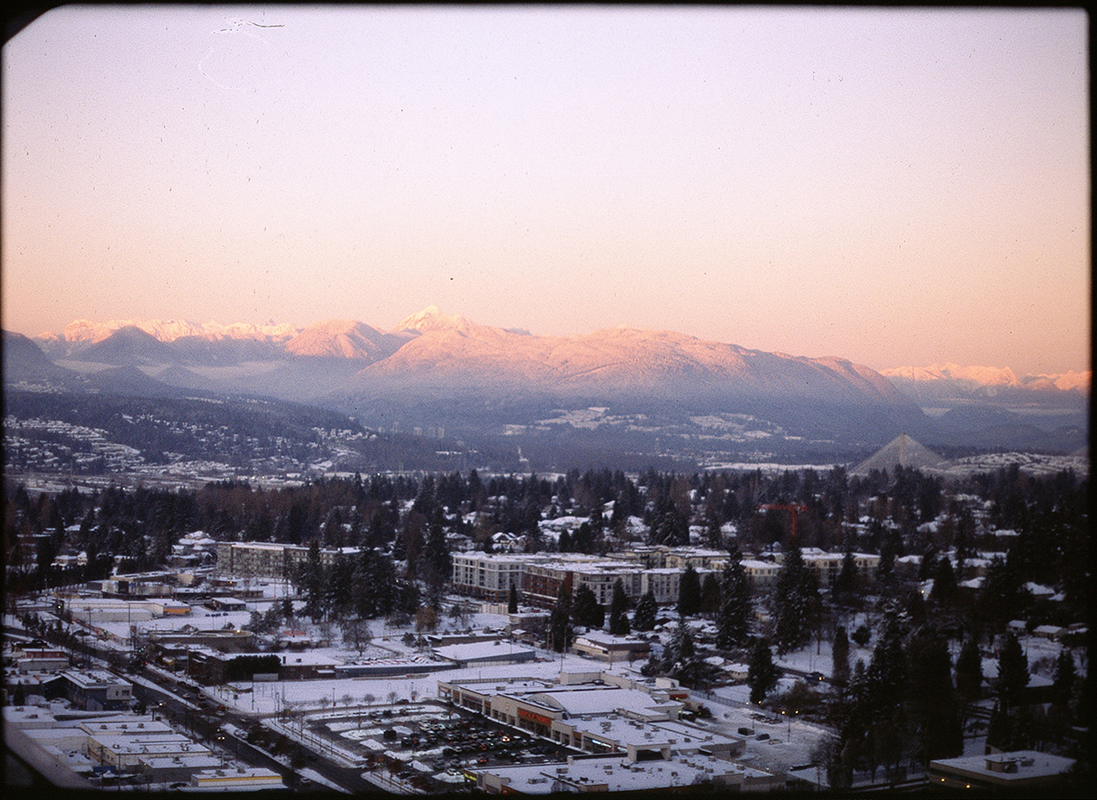}};
        \spy on (-1.63,1.75) in node [left] at (1.8,1.20);
        \end{tikzpicture}
    \caption{\textbf{Restoration by RePaint \cite{lugmayr2022repaint}}: using our segmentation.}
  \end{subfigure}
    \hfill
  \begin{subfigure}[t]{.325\textwidth}
    \centering
        \begin{tikzpicture}[spy using outlines={circle,blue,magnification=3,size=1.8cm, connect spies}]
        \node {\includegraphics[width=\linewidth]{figures/restoration_gt.jpg}};
        \spy on (-1.63,1.75) in node [left] at (1.8,1.20);
        \end{tikzpicture}
    \caption{\textbf{Ground Truth}: manually restored by human expert.}
  \end{subfigure}
  \caption{Input and ground truth from our authentic artefact damage dataset, along with chosen restorations. }
    \label{fig:qualitative-comparison-extended}
\end{figure*}

\begin{figure*}[hbtp]
  \begin{subfigure}[t]{.325\textwidth}
    \centering
        \begin{tikzpicture}[spy using outlines={circle,blue,magnification=3,size=1.8cm, connect spies}]
        \node {\includegraphics[width=\linewidth]{figures/input_cinestill50d_half_1.jpg}};
        \spy on (-1.7,-0.3) in node [left] at (1.8,1.1);
        \end{tikzpicture}
    \caption{\textbf{Input}: 4K film scan with authentic damage}
  \end{subfigure}
    \hfill
  \begin{subfigure}[t]{.325\textwidth}
    \centering
        \begin{tikzpicture}[spy using outlines={circle,yellow,magnification=3,size=1.8cm, connect spies}]
        \node {\includegraphics[width=\linewidth]{figures/cinestill50d_half_1_mask.png}};
        \spy on (-1.7,-0.3) in node [left] at (1.8,1.1);
        \end{tikzpicture}
    \caption{\textbf{Artefact Segmentation}: prediction from U-Net trained on synthetically damaged data.}
  \end{subfigure}
    \hfill
  \begin{subfigure}[t]{.325\textwidth}
    \centering
        \begin{tikzpicture}[spy using outlines={circle,yellow,magnification=3,size=1.8cm, connect spies}]
        \node {\includegraphics[width=\linewidth]{figures/bopb_mask_cinestill50d_half_1.png}};
        \spy on (-1.7,-0.3) in node [left] at (1.8,1.1);
        \end{tikzpicture}
    \caption{\textbf{Segmentation by BOPB~\cite{wan2020bringing}}.}
  \end{subfigure}
  
   \medskip
  
  \begin{subfigure}[t]{.325\textwidth}
    \centering
        \begin{tikzpicture}[spy using outlines={circle,blue,magnification=3,size=1.8cm, connect spies}]
        \node {\includegraphics[width=\linewidth]{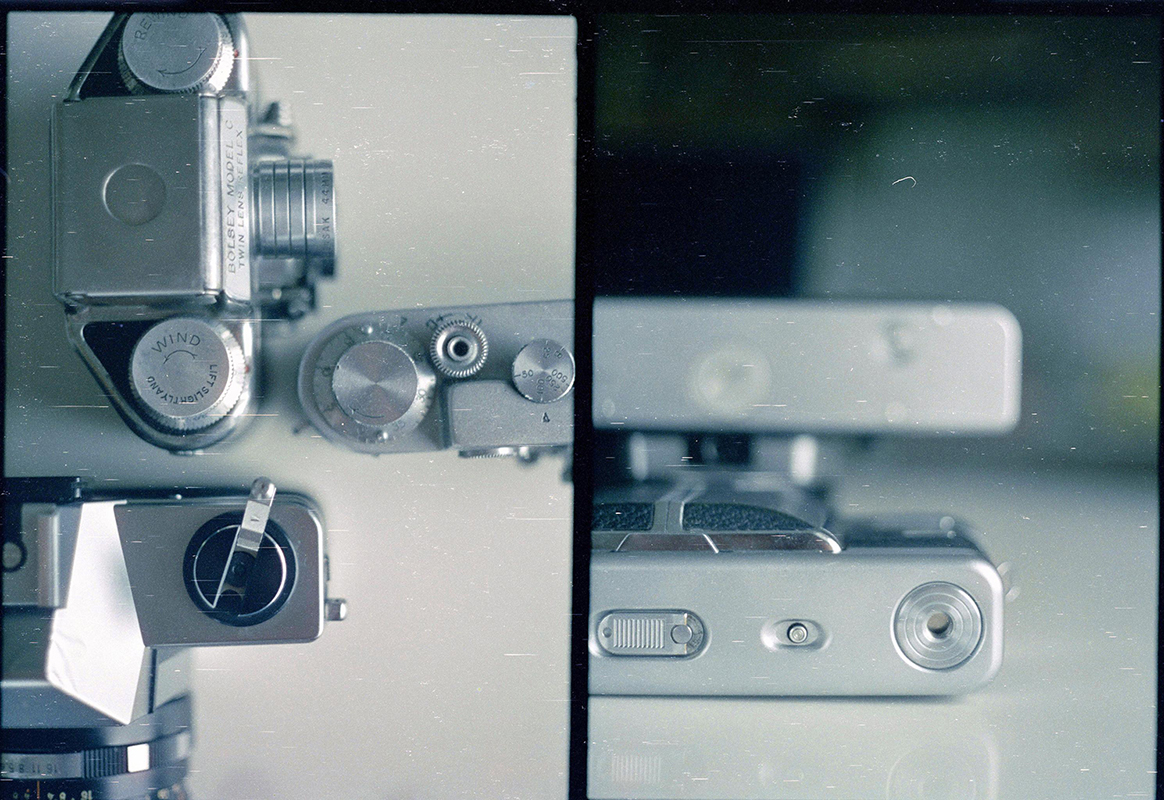}};
        \spy on (-1.7,-0.3) in node [left] at (1.8,1.1);
        \end{tikzpicture}
    \caption{\textbf{Restoration by U-Net + perceptual loss~\cite{visapp22}}: using originally provided model weights.}
  \end{subfigure}
    \hfill
  \begin{subfigure}[t]{.325\textwidth}
    \centering
        \begin{tikzpicture}[spy using outlines={circle,blue,magnification=3,size=1.8cm, connect spies}]
        \node {\includegraphics[width=\linewidth]{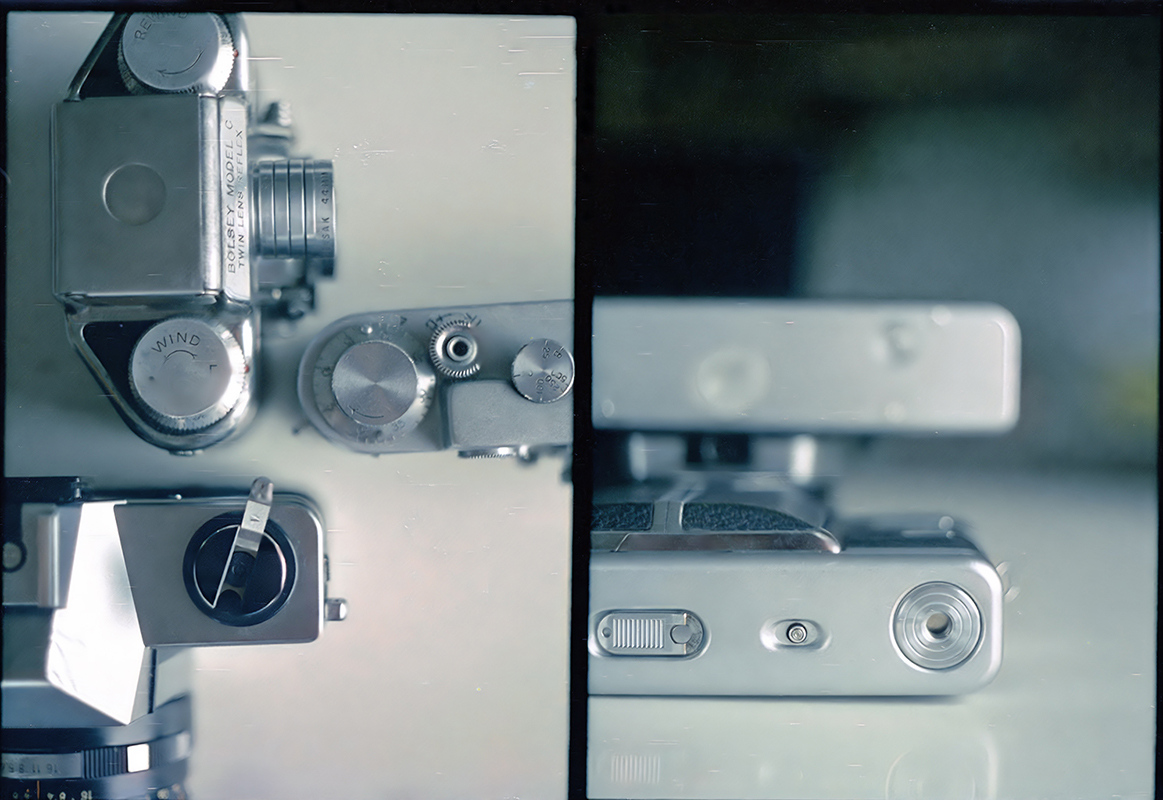}};
        \spy on (-1.7,-0.3) in node [left] at (1.8,1.1);
        \end{tikzpicture}
    \caption{\textbf{Restoration by BOPB~\cite{wan2020bringing}}: using our segmentation.}
  \end{subfigure}
    \hfill
  \begin{subfigure}[t]{.325\textwidth}
    \centering
        \begin{tikzpicture}[spy using outlines={circle,blue,magnification=3,size=1.8cm, connect spies}]
        \node {\includegraphics[width=\linewidth]{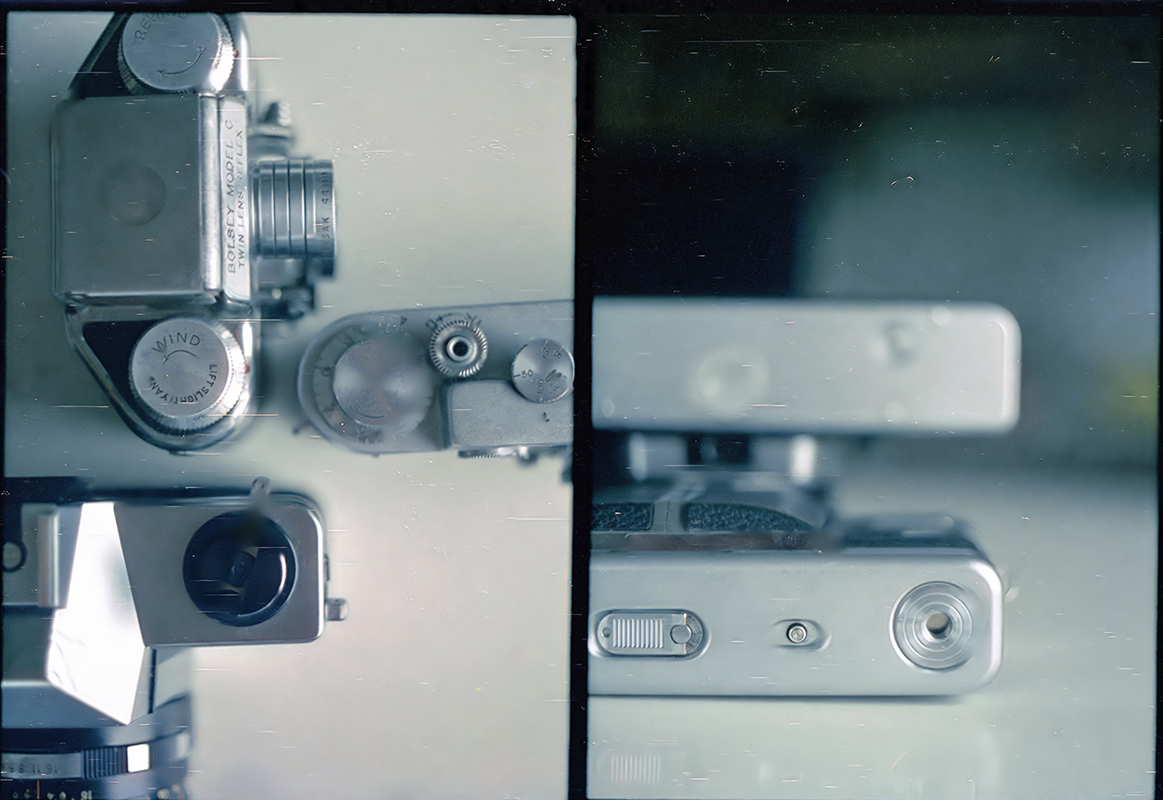}};
        \spy on (-1.7,-0.3) in node [left] at (1.8,1.1);
        \end{tikzpicture}
    \caption{\textbf{Restoration by BOPB~\cite{wan2020bringing}}: using their segmentation.}
  \end{subfigure}
  
   \medskip
  
  \begin{subfigure}[t]{.325\textwidth}
    \centering
        \begin{tikzpicture}[spy using outlines={circle,blue,magnification=3,size=1.8cm, connect spies}]
        \node {\includegraphics[width=\linewidth]{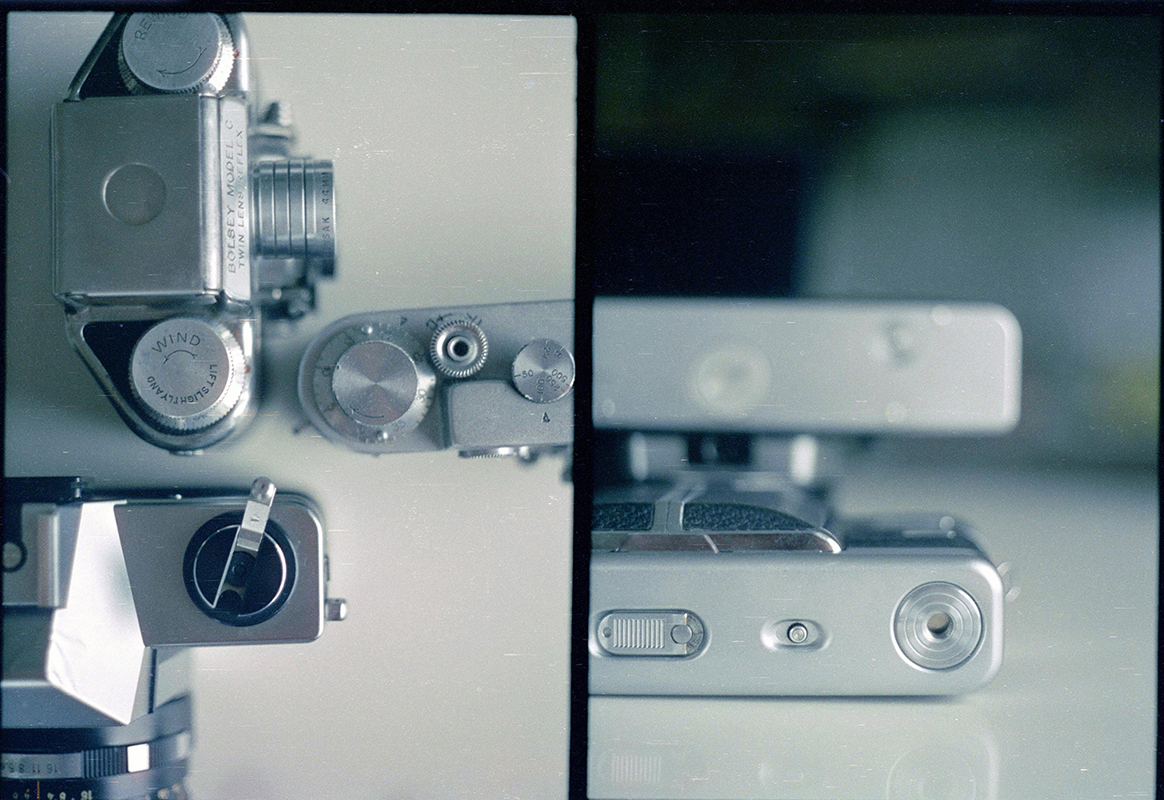}};
        \spy on (-1.7,-0.3) in node [left] at (1.8,1.1);
        \end{tikzpicture}
    \caption{\textbf{Restoration by U-Net + perceptual loss~\cite{visapp22}}: retrained on our synthetic damage.}
  \end{subfigure}
    \hfill
  \begin{subfigure}[t]{.325\textwidth}
    \centering
        \begin{tikzpicture}[spy using outlines={circle,blue,magnification=3,size=1.8cm, connect spies}]
        \node {\includegraphics[width=\linewidth]{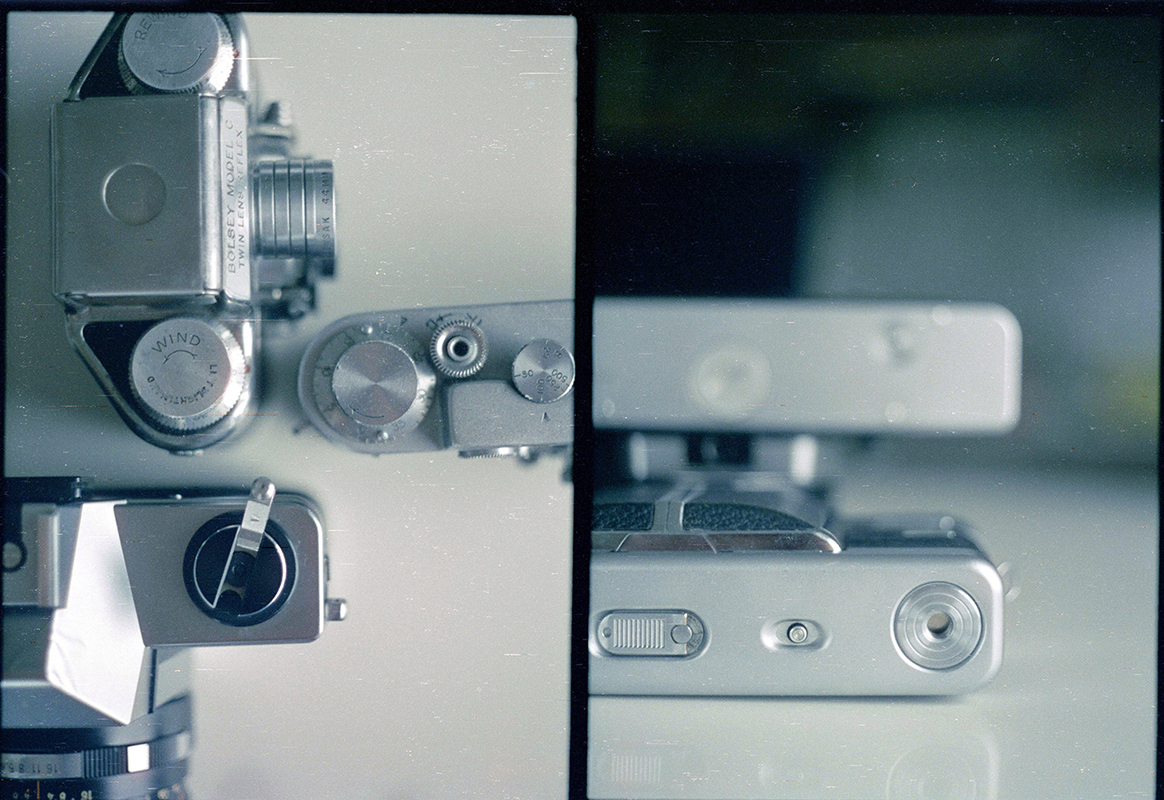}};
        \spy on (-1.7,-0.3) in node [left] at (1.8,1.1);
        \end{tikzpicture}
    \caption{\textbf{Restoration by LaMa~\cite{suvorov2022resolution}}: best performing model, using our segmentation.}
  \end{subfigure}
    \hfill
  \begin{subfigure}[t]{.325\textwidth}
    \centering
        \begin{tikzpicture}[spy using outlines={circle,blue,magnification=3,size=1.8cm, connect spies}]
        \node {\includegraphics[width=\linewidth]{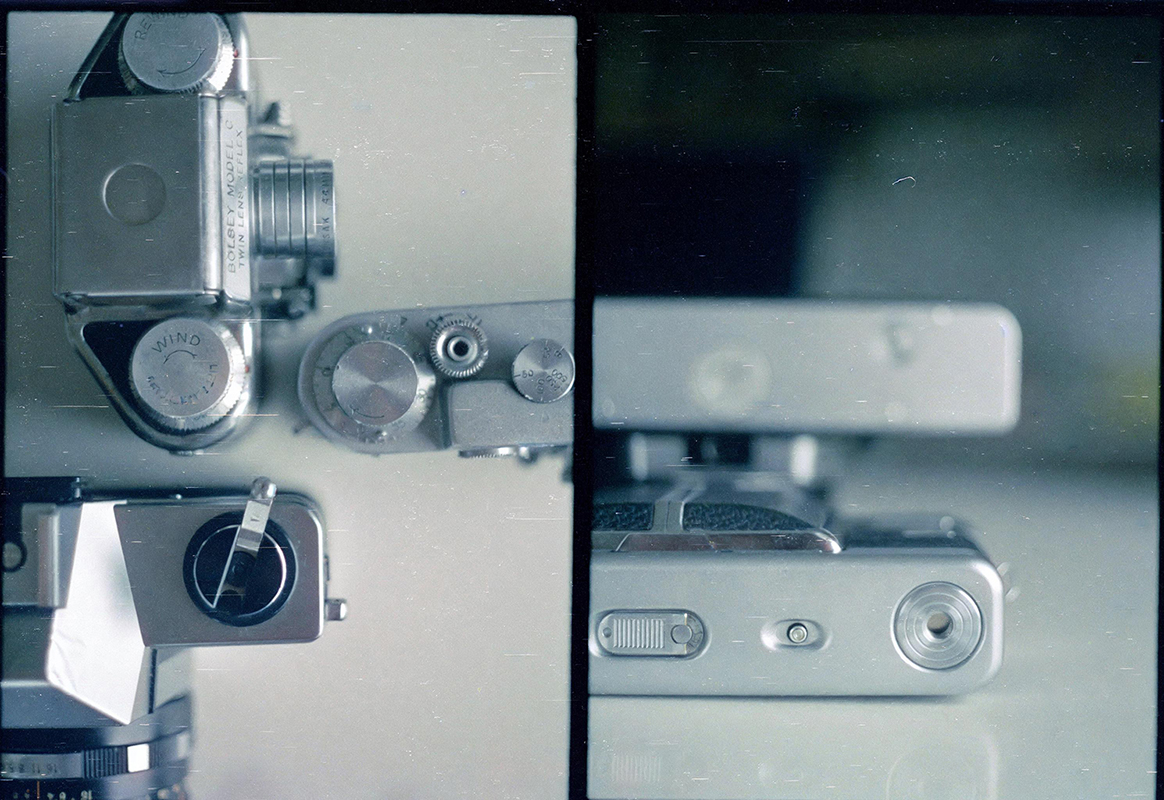}};
        \spy on (-1.7,-0.3) in node [left] at (1.8,1.1);
        \end{tikzpicture}
    \caption{\textbf{Restoration by Stable Diffusion~\cite{rombach2021highresolution}}: using our segmentation.}
  \end{subfigure}
 
    \medskip
  
  \begin{subfigure}[t]{.325\textwidth}
    \centering
        \begin{tikzpicture}[spy using outlines={circle,blue,magnification=3,size=1.8cm, connect spies}]
        \node {\includegraphics[width=\linewidth]{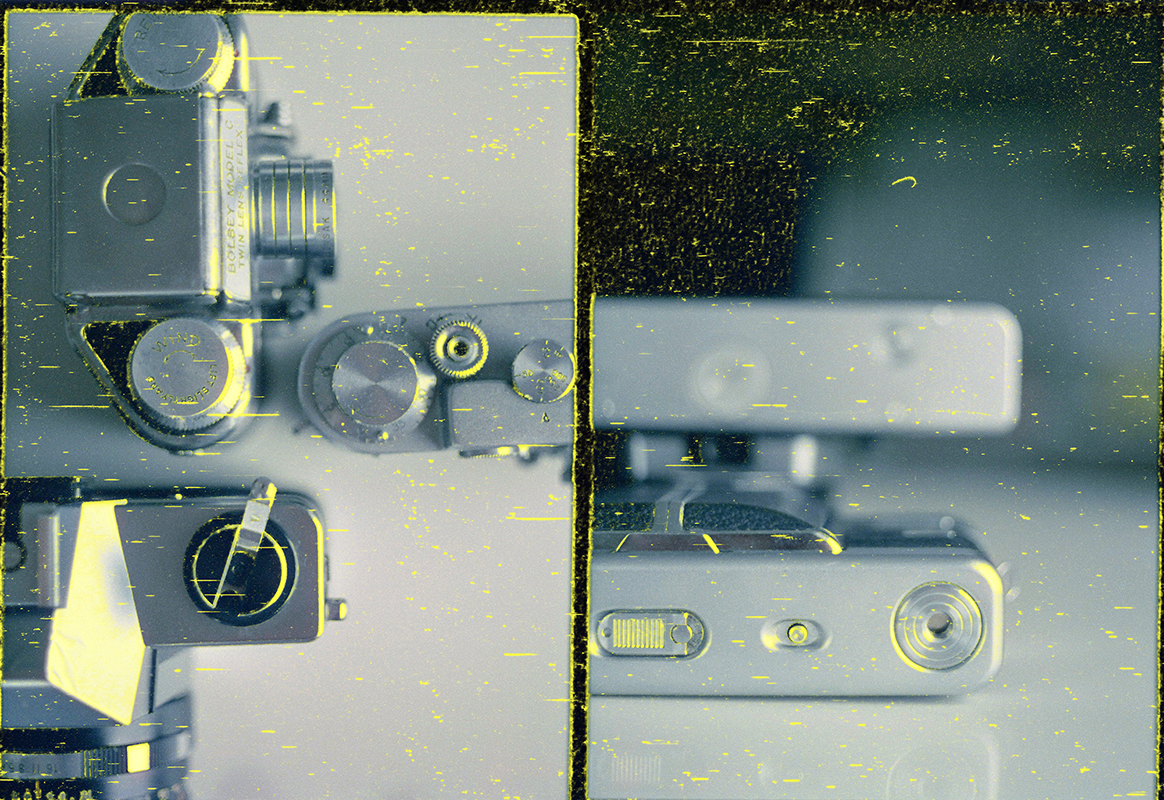}};
        \spy on (-1.7,-0.3) in node [left] at (1.8,1.1);
        \end{tikzpicture}
    \caption{\textbf{Restoration by BVMR~\cite{visapp22}}: retrained on our synthetic damage.}
  \end{subfigure}
    \hfill
  \begin{subfigure}[t]{.325\textwidth}
    \centering
        \begin{tikzpicture}[spy using outlines={circle,blue,magnification=3,size=1.8cm, connect spies}]
        \node {\includegraphics[width=\linewidth]{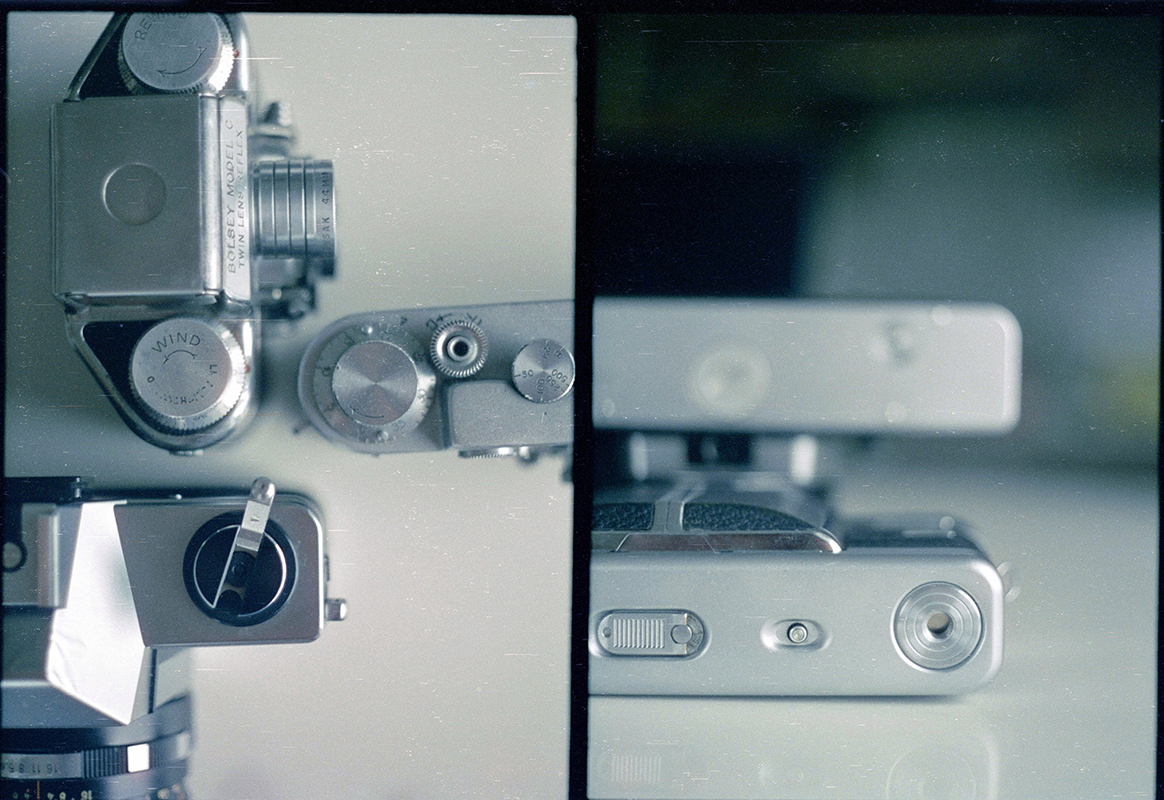}};
        \spy on (-1.7,-0.3) in node [left] at (1.8,1.1);
        \end{tikzpicture}
    \caption{\textbf{Restoration by RePaint \cite{lugmayr2022repaint}}: using our segmentation.}
  \end{subfigure}
    \hfill
  \begin{subfigure}[t]{.325\textwidth}
    \centering
        \begin{tikzpicture}[spy using outlines={circle,blue,magnification=3,size=1.8cm, connect spies}]
        \node {\includegraphics[width=\linewidth]{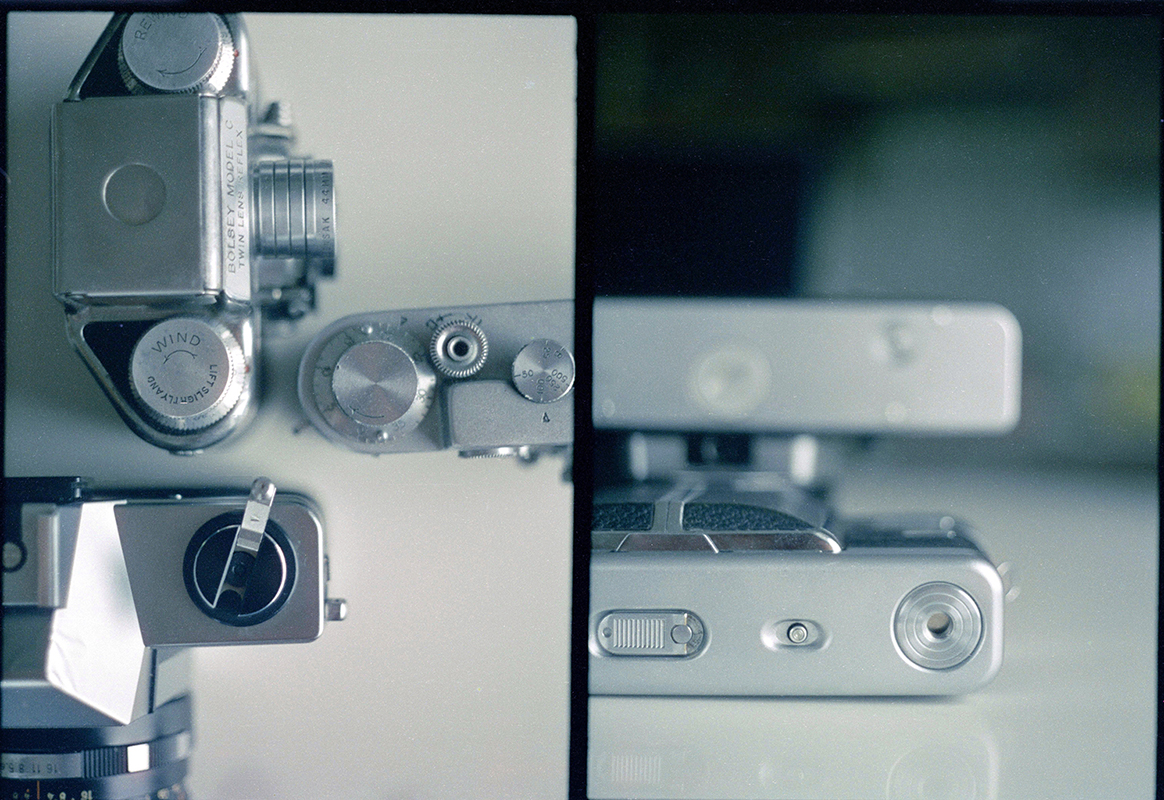}};
        \spy on (-1.7,-0.3) in node [left] at (1.8,1.1);
        \end{tikzpicture}
    \caption{\textbf{Ground Truth}: manually restored by human expert.}
  \end{subfigure}
  \caption{Input and ground truth from our authentic artefact damage dataset, along with chosen restorations. }
    \label{fig:qualitative-comparison-extended2}
\end{figure*}


\begin{figure*}[hbtp]
  \begin{subfigure}[t]{.325\textwidth}
    \centering
        \begin{tikzpicture}[spy using outlines={circle,blue,magnification=3,size=1.8cm, connect spies}]
        \node {\includegraphics[width=\linewidth]{figures/input_cinestill800t_half_1.jpg}};
        \spy on (2.25,-0.1) in node [left] at (1.8,1.1);
        \end{tikzpicture}
    \caption{\textbf{Input}: 4K film scan with authentic damage}
  \end{subfigure}
    \hfill
  \begin{subfigure}[t]{.325\textwidth}
    \centering
        \begin{tikzpicture}[spy using outlines={circle,yellow,magnification=3,size=1.8cm, connect spies}]
        \node {\includegraphics[width=\linewidth]{figures/cinestill800t_half_1_mask.png}};
        \spy on (2.25,-0.1) in node [left] at (1.8,1.1);
        \end{tikzpicture}
    \caption{\textbf{Artefact Segmentation}: prediction from U-Net trained on synthetically damaged data.}
  \end{subfigure}
    \hfill
  \begin{subfigure}[t]{.325\textwidth}
    \centering
        \begin{tikzpicture}[spy using outlines={circle,yellow,magnification=3,size=1.8cm, connect spies}]
        \node {\includegraphics[width=\linewidth]{figures/bopb_mask_cinestill800t_half_1.png}};
        \spy on (2.25,-0.1) in node [left] at (1.8,1.1);
        \end{tikzpicture}
    \caption{\textbf{Segmentation by BOPB~\cite{wan2020bringing}}.}
  \end{subfigure}
  
   \medskip
  
  \begin{subfigure}[t]{.325\textwidth}
    \centering
        \begin{tikzpicture}[spy using outlines={circle,blue,magnification=3,size=1.8cm, connect spies}]
        \node {\includegraphics[width=\linewidth]{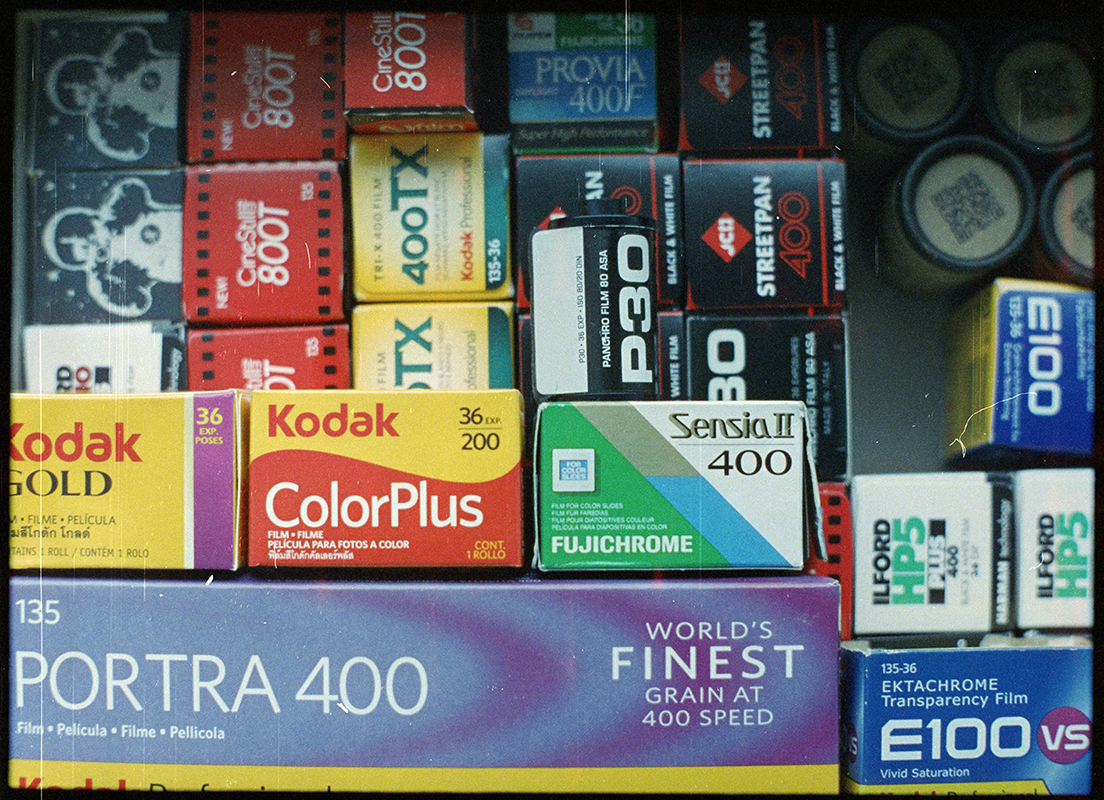}};
        \spy on (2.25,-0.1) in node [left] at (1.8,1.1);
        \end{tikzpicture}
    \caption{\textbf{Restoration by U-Net + perceptual loss~\cite{visapp22}}: using originally provided model weights.}
  \end{subfigure}
    \hfill
  \begin{subfigure}[t]{.325\textwidth}
    \centering
        \begin{tikzpicture}[spy using outlines={circle,blue,magnification=3,size=1.8cm, connect spies}]
        \node {\includegraphics[width=\linewidth]{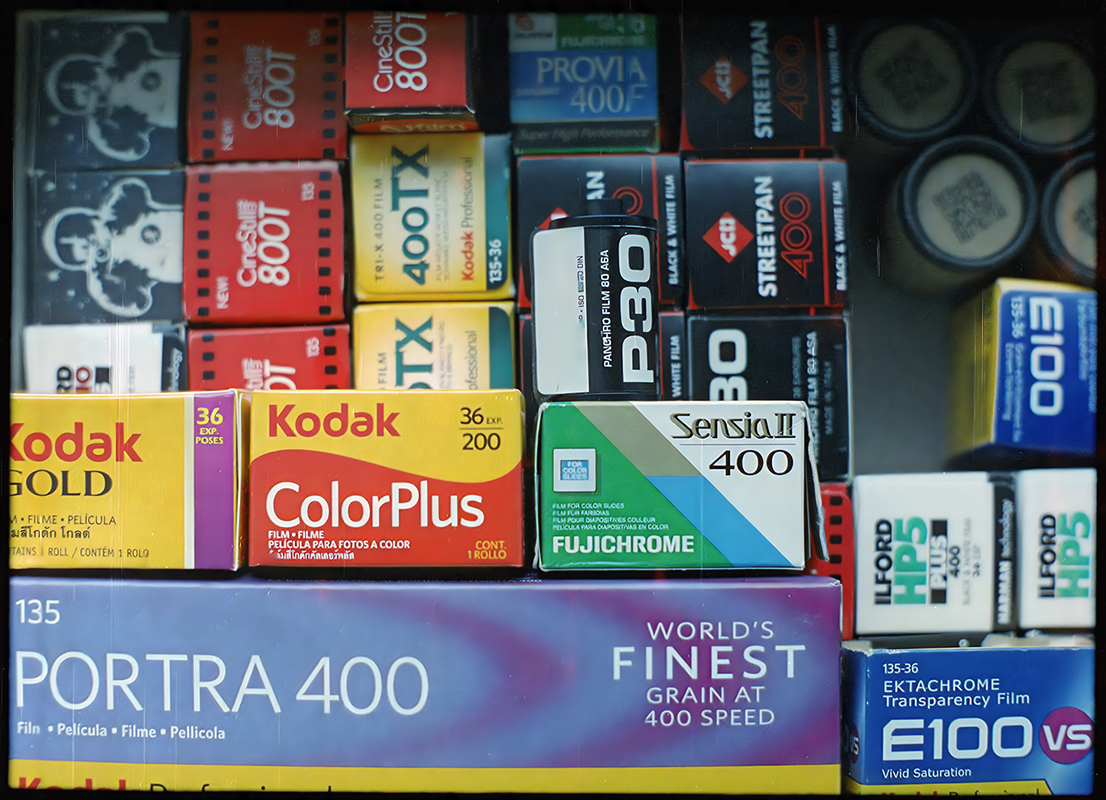}};
        \spy on (2.25,-0.1) in node [left] at (1.8,1.1);
        \end{tikzpicture}
    \caption{\textbf{Restoration by BOPB~\cite{wan2020bringing}}: using our segmentation.}
  \end{subfigure}
    \hfill
  \begin{subfigure}[t]{.325\textwidth}
    \centering
        \begin{tikzpicture}[spy using outlines={circle,blue,magnification=3,size=1.8cm, connect spies}]
        \node {\includegraphics[width=\linewidth]{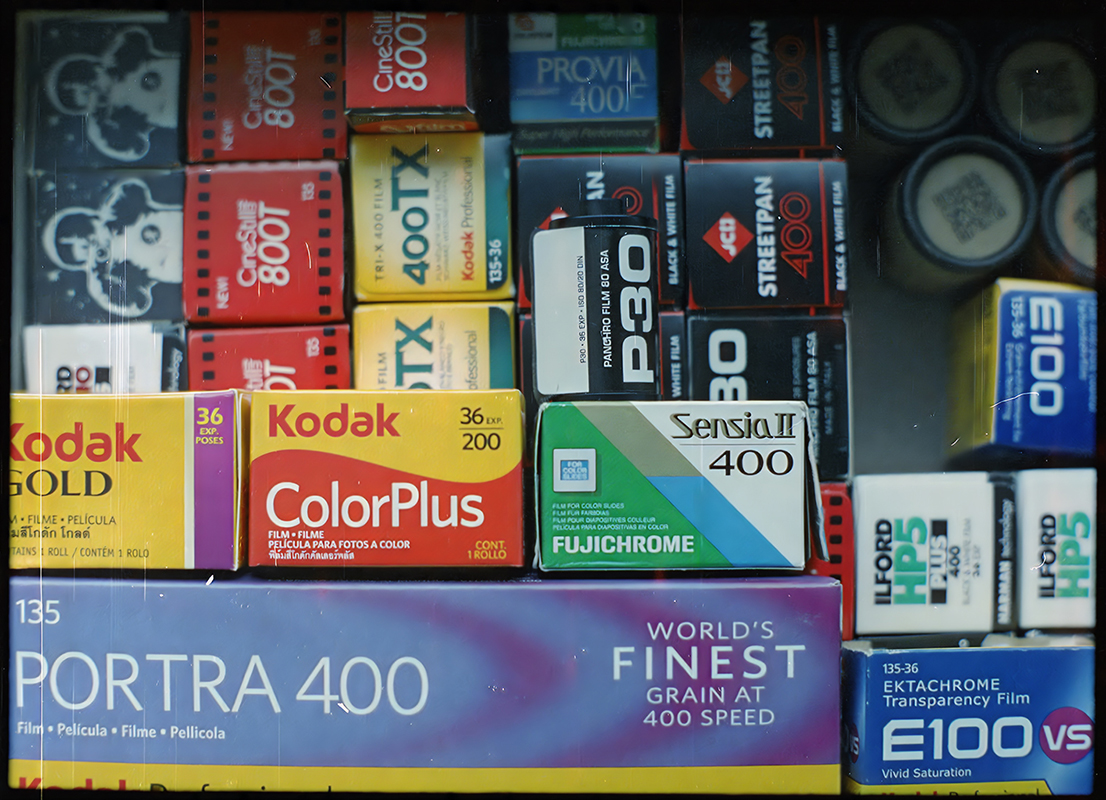}};
        \spy on (2.25,-0.1) in node [left] at (1.8,1.1);
        \end{tikzpicture}
    \caption{\textbf{Restoration by BOPB~\cite{wan2020bringing}}: using their segmentation.}
  \end{subfigure}
  
   \medskip
  
  \begin{subfigure}[t]{.325\textwidth}
    \centering
        \begin{tikzpicture}[spy using outlines={circle,blue,magnification=3,size=1.8cm, connect spies}]
        \node {\includegraphics[width=\linewidth]{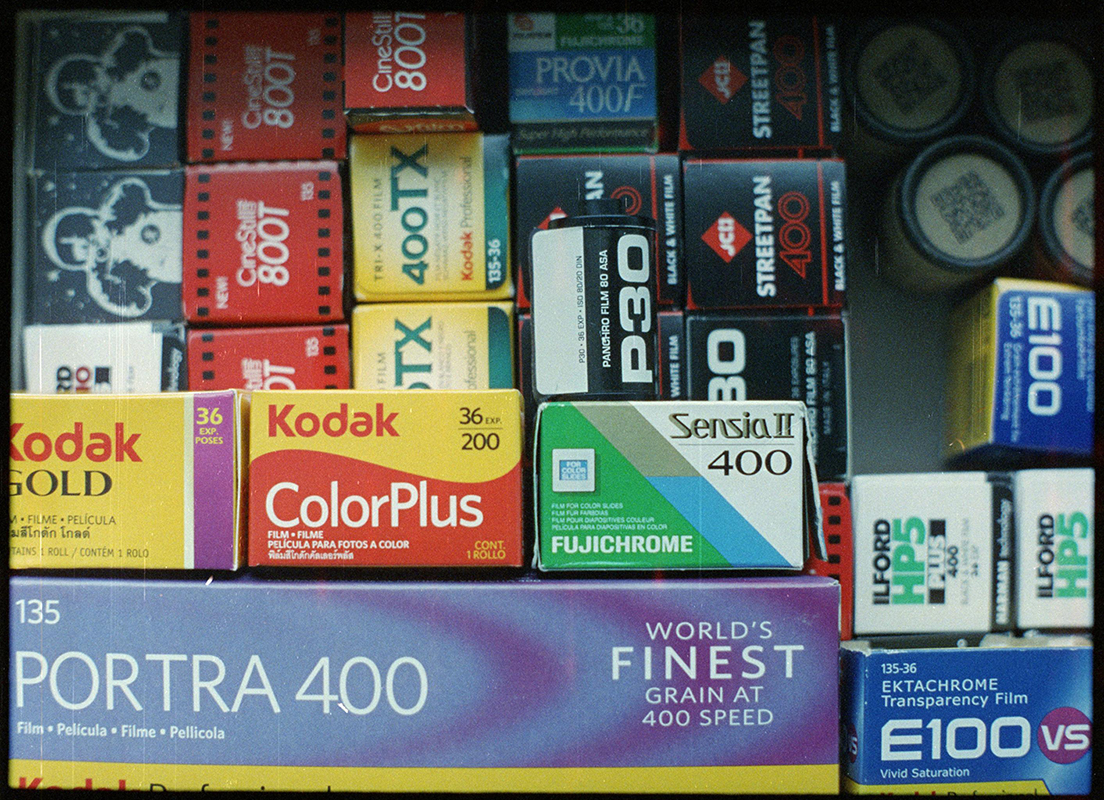}};
        \spy on (2.25,-0.1) in node [left] at (1.8,1.1);
        \end{tikzpicture}
    \caption{\textbf{Restoration by U-Net + perceptual loss~\cite{visapp22}}: retrained on our synthetic damage.}
  \end{subfigure}
    \hfill
  \begin{subfigure}[t]{.325\textwidth}
    \centering
        \begin{tikzpicture}[spy using outlines={circle,blue,magnification=3,size=1.8cm, connect spies}]
        \node {\includegraphics[width=\linewidth]{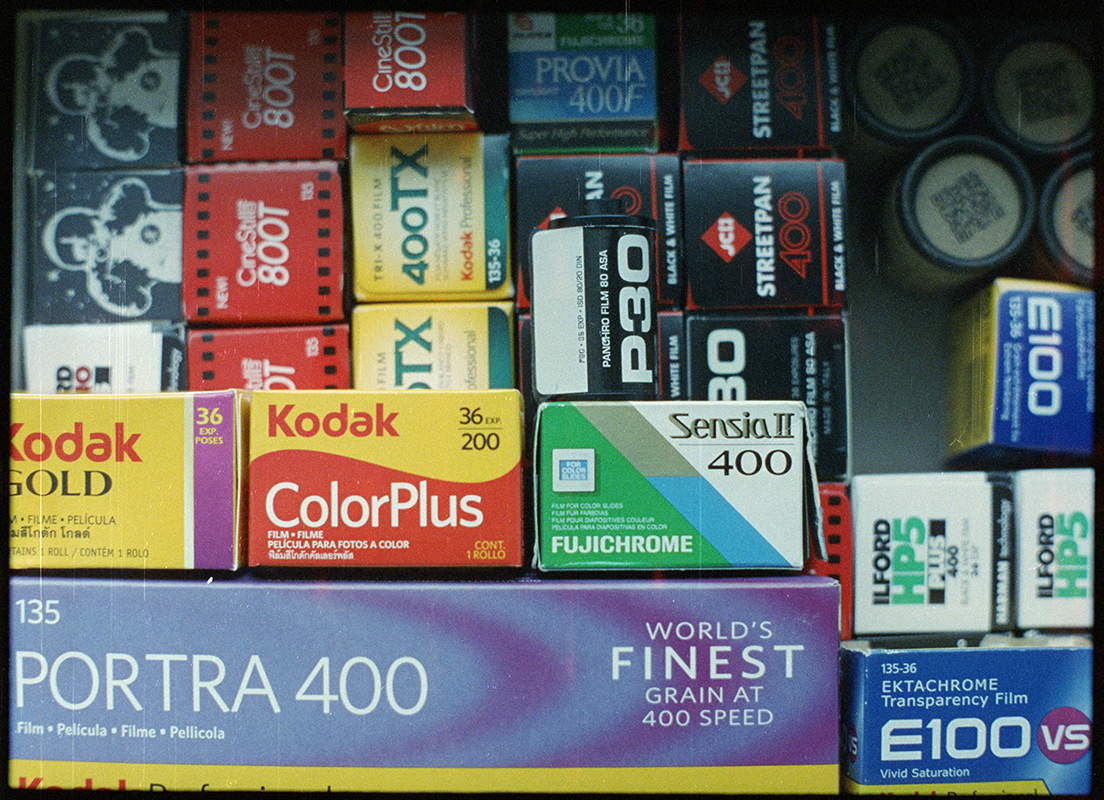}};
        \spy on (2.25,-0.1) in node [left] at (1.8,1.1);
        \end{tikzpicture}
    \caption{\textbf{Restoration by LaMa~\cite{suvorov2022resolution}}: best performing model, using our segmentation.}
  \end{subfigure}
    \hfill
  \begin{subfigure}[t]{.325\textwidth}
    \centering
        \begin{tikzpicture}[spy using outlines={circle,blue,magnification=3,size=1.8cm, connect spies}]
        \node {\includegraphics[width=\linewidth]{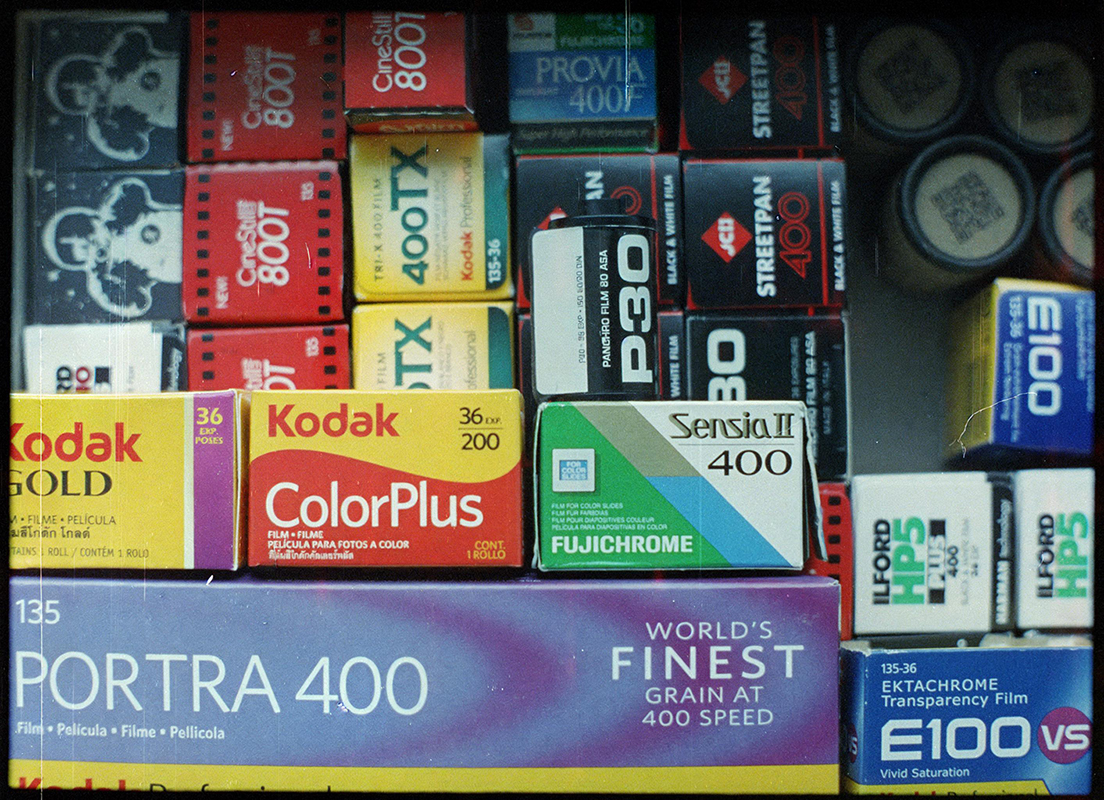}};
        \spy on (2.25,-0.1) in node [left] at (1.8,1.1);
        \end{tikzpicture}
    \caption{\textbf{Restoration by Stable Diffusion~\cite{rombach2021highresolution}}: using our segmentation.}
  \end{subfigure}
 
    \medskip
  
  \begin{subfigure}[t]{.325\textwidth}
    \centering
        \begin{tikzpicture}[spy using outlines={circle,blue,magnification=3,size=1.8cm, connect spies}]
        \node {\includegraphics[width=\linewidth]{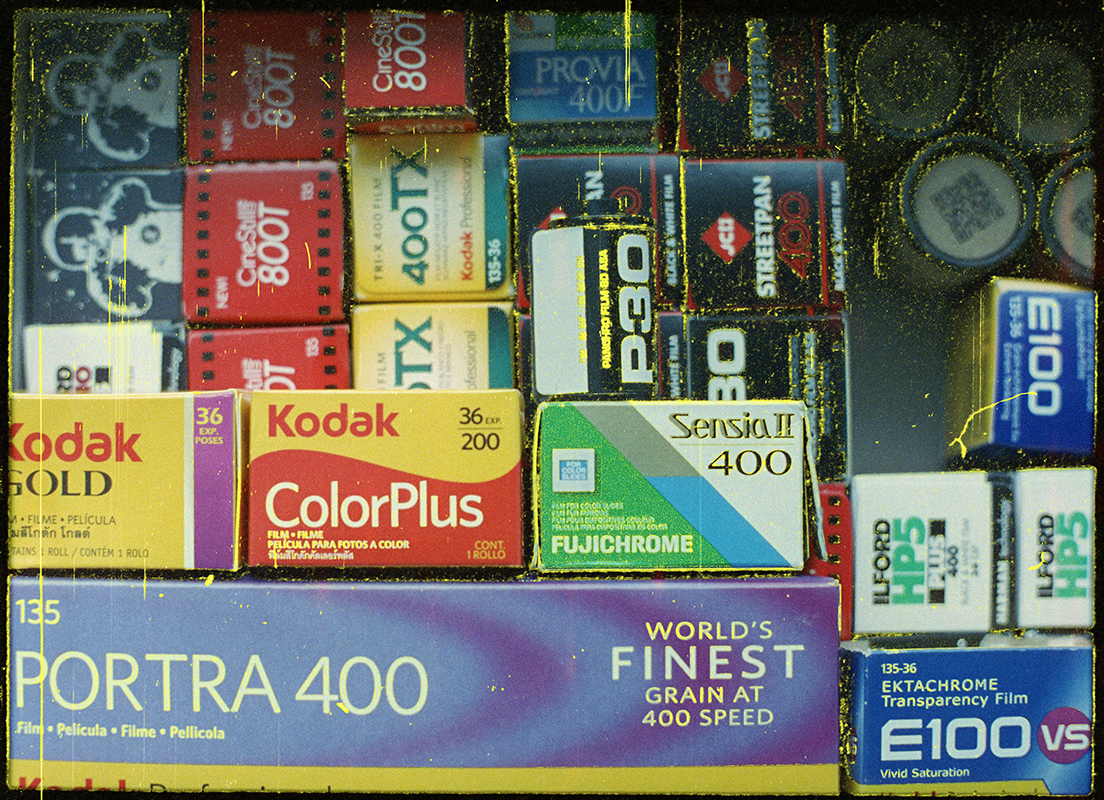}};
        \spy on (2.25,-0.1) in node [left] at (1.8,1.1);
        \end{tikzpicture}
    \caption{\textbf{Restoration by BVMR~\cite{visapp22}}: retrained on our synthetic damage.}
  \end{subfigure}
    \hfill
  \begin{subfigure}[t]{.325\textwidth}
    \centering
        \begin{tikzpicture}[spy using outlines={circle,blue,magnification=3,size=1.8cm, connect spies}]
        \node {\includegraphics[width=\linewidth]{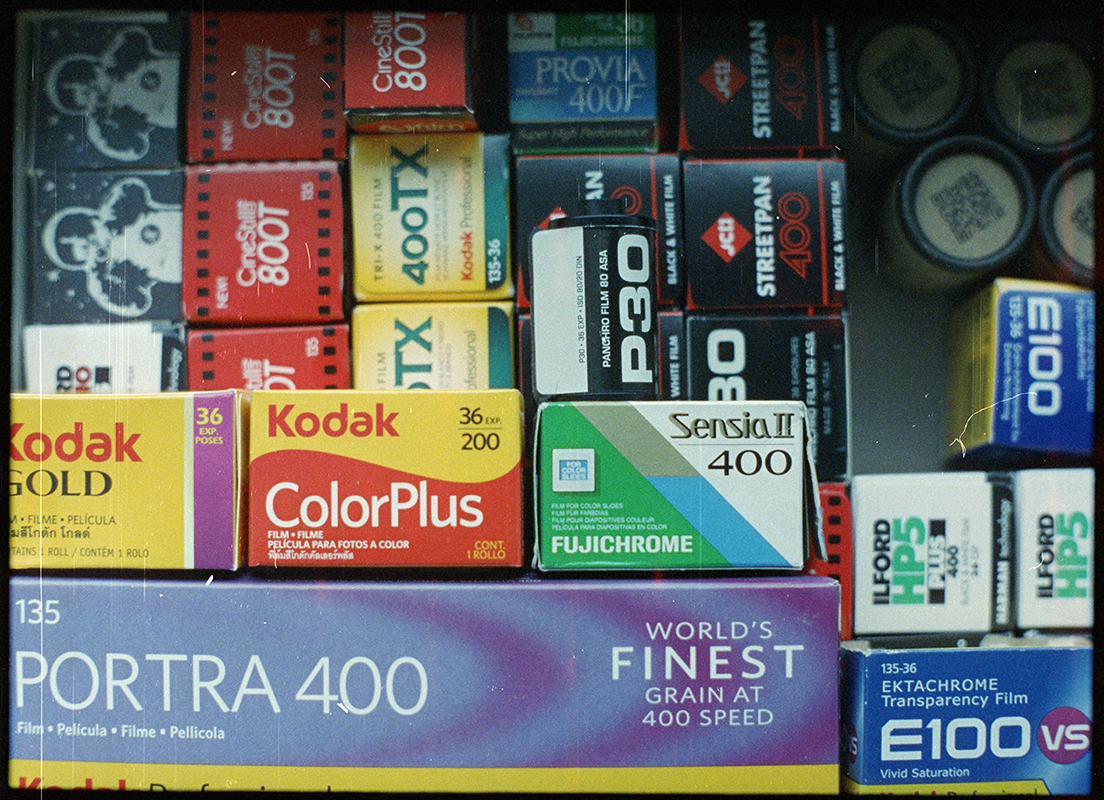}};
        \spy on (2.25,-0.1) in node [left] at (1.8,1.1);
        \end{tikzpicture}
    \caption{\textbf{Restoration by RePaint \cite{lugmayr2022repaint}}: using our segmentation.}
  \end{subfigure}
    \hfill
  \begin{subfigure}[t]{.325\textwidth}
    \centering
        \begin{tikzpicture}[spy using outlines={circle,blue,magnification=3,size=1.8cm, connect spies}]
        \node {\includegraphics[width=\linewidth]{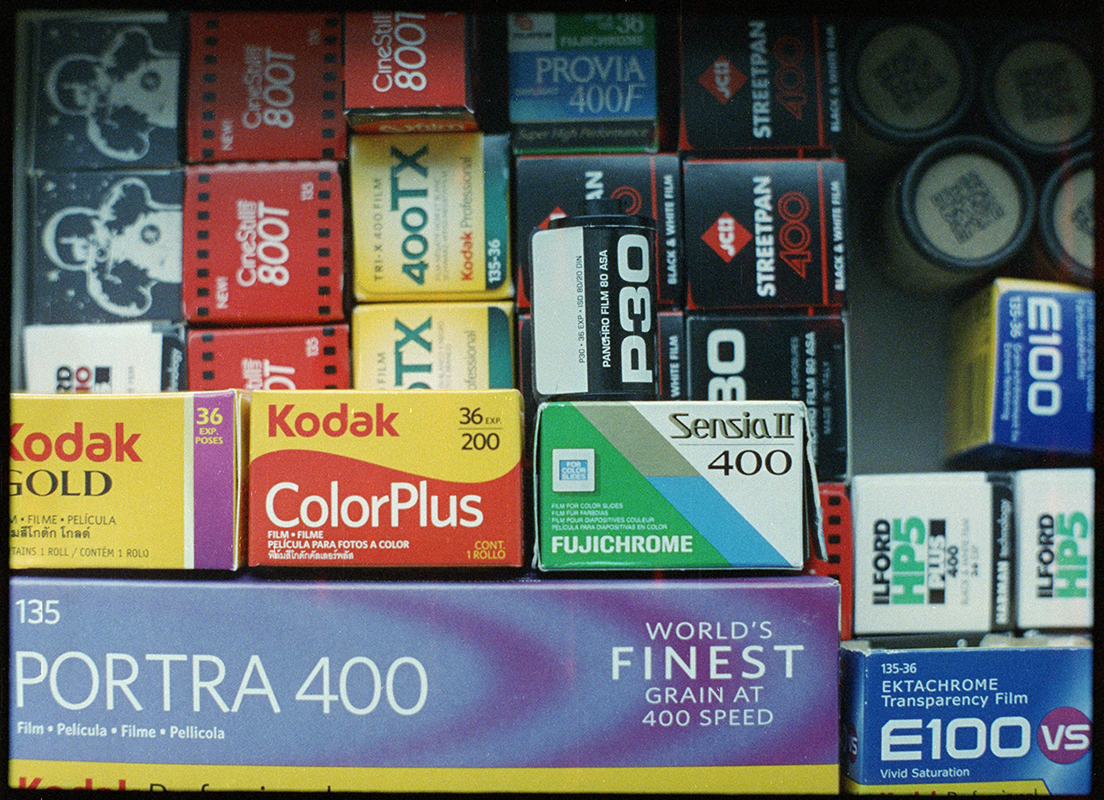}};
        \spy on (2.25,-0.1) in node [left] at (1.8,1.1);
        \end{tikzpicture}
    \caption{\textbf{Ground Truth}: manually restored by human expert.}
  \end{subfigure}
  \caption{Input and ground truth from our authentic artefact damage dataset, along with chosen restorations. }
    \label{fig:qualitative-comparison-extended4}
\end{figure*}

\setlength{\tabcolsep}{6pt}

\begin{table*}[]
\resizebox{\linewidth}{!}{%
\begin{tabular}{@{}lllllllll@{}}
\toprule
\multirow{2}{*}{\textbf{Method}} & \multirow{2}{*}{\textbf{Notes}} & \multirow{2}{*}{\textbf{Patch-wise}}                               & \multicolumn{2}{c}{\textbf{PSNR (dB)} $\uparrow$} & \multicolumn{2}{c}{\textbf{SSIM} $\uparrow$} & \multicolumn{2}{c}{\textbf{LPIPS} $\downarrow$} \\ \cmidrule(l){4-5}\cmidrule(lr){6-7}  \cmidrule(l){8-9}
                                  &  &               & Mean        & Std     & Mean        & Std     & Mean        & Std      \\ \midrule
BOPB~\cite{wan2020bringing} &
    & Yes            
    & \round{1}{14.779}      & \round{2}{3.137547}   
    & \round{3}{0.756268}    & \round{3}{0.06292}    
    & \round{3}{0.560866}    & \round{3}{0.060192}    \\
BOPB~\cite{wan2020bringing}  &    & No           
    & \round{1}{24.00027}   & \round{2}{2.798892}   
    & \round{3}{0.904336}    & \round{3}{0.048154}   
    & \round{3}{0.431766}    & \round{3}{0.06167}     \\
BOPB~\cite{wan2020bringing} & w/ our masks 
    & Yes                                               
    & \round{1}{14.912402}   & \round{2}{3.6684} 
    & \round{3}{0.76927}     & \round{3}{0.067617}   
    & \round{3}{0.555698}    & \round{3}{0.066783}    \\
BOPB~~\cite{wan2020bringing} & w/ our masks 
    & No 
    & \round{1}{24.353391}   & \round{2}{2.341881}   
    & \round{3}{0.917082}    & \round{3}{0.043026}   
    & \round{3}{0.424936}    & \round{3}{0.424936}    \\
\textbf{LaMa~\cite{suvorov2022resolution}}  & \textbf{w/ our masks}
    & \textbf{No}           
    & $\mathbf{39.6}$   
    & $\mathbf{5.24}$   
    & $\mathbf{0.991}$
    & $\mathbf{0.012}$   
    & $\mathbf{0.042}$    & $\mathbf{0.038}$    \\
Photoshop Dust \& Scratch Filter    &
    & No        
    & \round{1}{28.888434}   & \round{2}{3.09088}    
    & \round{3}{0.96485}     & \round{3}{0.02187}    
    & \round{3}{0.276773}    & \round{3}{0.087693}    \\
U-Net + perceptual loss~\cite{visapp22}    & trained on downsampled frames   
    & Yes                    
    & \round{1}{31.827841}   & \round{2}{2.62107}    
    & \round{3}{0.978925}    & \round{3}{0.014816}   
    & \round{3}{0.224448}    & \round{3}{0.046726}    \\
U-Net + perceptual loss~\cite{visapp22}    & re-trained on DeepRemaster damage~\cite{iizuka2019deepremaster}   
    & Yes                   
    & \round{1}{27.477950}   & \round{2}{1.358982}    
    & \round{3}{0.945480}    & \round{3}{0.015137}   
    & \round{3}{0.248534}    & \round{3}{0.043398}    \\
U-Net + perceptual loss~\cite{visapp22} & re-trained on our synthetic damage  
    & Yes  
    & $33.0$   & $2.46$   
    & $0.984$    & $0.011$   
    & $0.215$    & $0.044$    \\
Stable Diffusion ~\cite{rombach2021highresolution}   &   w/ our masks
    & Yes                
    & \round{1}{28.649723}   & \round{2}{2.427813}   
    & \round{3}{0.958839}    & \round{3}{0.019639}   
    & \round{3}{0.305305}    & \round{3}{0.039742}    \\
BVMR~\cite{hertz2019blind}          & re-trained on artefact damage  & Yes                
    & \round{1}{17.779836}   & \round{2}{4.084066}   
    & \round{3}{0.798220}    & \round{3}{0.108210}   
    & \round{3}{0.391832}    & \round{3}{0.153276}    \\
Restormer~\cite{Zamir_2022_CVPR}    & de-raining model
    & Yes                
    & \round{1}{32.029198}   & \round{2}{2.515785}   
    & \round{3}{0.977807}    & \round{3}{0.014820}   
    & \round{3}{0.238450}    & \round{3}{0.043950}    \\
\bottomrule
\end{tabular}%
}
\caption{Quantitative evaluation of restoration results on our high resolution dataset of authentic film damage.}
\label{tab:evaluation-results}
\end{table*}

\subsection{Restoration results and discussion}

Table \ref{tab:evaluation-results} summarises our quantitative results, showing PSNR, SSIM and LPIPS metrics with respect to the ground-truth professionally-restored image set. All three metrics largely agree, with LaMa and the U-Net retrained with our damage as the best and second best performing models respectively. We can directly observe the substantial improvement that re-training the simple U-Net restoration model \cite{visapp22} with our simulated damage brings to its restoration performance. On the other hand, re-training the same model with the damage from DeepRemaster~\cite{iizuka2019deepremaster} results in deteriorated restoration performance.

BOPB works especially poorly in patch-wise mode as it introduces color distribution shift in each patch, leading to a very distorted result and correspondingly low scores. 
We attribute this to the fact that BOPB aims to remove film grain and modify color to match modern digital photographs, in addition to removing artefacts. This means it performs poorly at our specific task of damage removal (and is thus unsuitable for applications where the qualities of analogue photographs are desirable).
%
%
Interestingly, the standard ``Dust and Scratches filter'' in Photoshop exceeds the performance of BOPB (patchwise and fully-convolutional modes), Stable Diffusion and BVMR.
Restormer and the best U-Net model have comparable performance, but LaMa exceeds the performance of all of the competitive models on all three metrics, particularly as measured by LPIPS.
This is in spite of LaMa being trained with inpainting masks that are significantly different in size and shape to typical analogue damage artefacts.
Note that we do not include results for RePaint in Table~\ref{tab:evaluation-results}.
This model requires over 12 minutes for the restoration of a single $256\times256$ pixel patch on a Tesla P100 GPU.
Given that one 4K image from our test set is split into around 200 overlapping patches, about 40 hours are needed to process just one image. While this is clearly impractical in a real-life restoration scenario, we process a subset of 10 images from our test set for qualitative evaluation.

We present qualitative results in Figures~\ref{fig:qualitative-comparison-extended}, \ref{fig:qualitative-comparison-extended2} and \ref{fig:qualitative-comparison-extended4}, along with additional examples in the supplementary material, Figures S2--4.
BOPB performs significantly better when we substitute its predicted artefact segmentation masks (which are strongly under-segmented) with ours.
However, it still modifies colors and smoothes the film grain, significantly changing the character of images. 
Even when guided by our segmentation masks, Stable Diffusion fails to inpaint the damage. Similarly, BMVR, even when re-trained using our synthetic damage, struggles with the same issue as BOPB and injects additional damage.
In general, we find that methods based on inpainting which do not modify regions away from the artefacts perform much better than BOPB, as they are less likely to introduce new damage or distortions.
In line with our observations from the quantitative evaluation, retraining the U-Net model \cite{visapp22} on our improved synthetic damage results in higher quality restorations.
Visually, LaMa performs slightly better than \cite{visapp22}; however both methods fail to inpaint part of the upper hair in Figure~\ref{fig:qualitative-comparison}.
RePaint's performance, guided by our segmentation masks, is comparable to that of LaMa and \cite{visapp22}. Despite producing restorations of relatively high visual quality, the model is limited in practical usefulness by the amount of time it takes to process images.


\section{Conclusions}
Automated restoration of mechanical damage in scanned film is a challenging task. Fully automated restoration would be transformative in improving the quality of imaging in the enormous archives of extant film held around the world. However, conventional inpainting and restoration processes are unsuitable for professional-quality restoration. All of the machine learning approaches we tested for film restoration perform well below the level required to be competitive with professional hand restoration. Models trained on low-resolution patches performed very poorly when applied patch-wise, but even state-of-the-art models like LaMa applied at native resolution were unable to adequately inpaint artefacts without introducing undue distortion. Many of the algorithms would also be computationally impractical for processing large image collections at high resolution.

Progress towards film restoration that can operate truly automatically at realistic scan resolutions requires better models of the film damage process. To that end, we have presented both a sophisticated statistical model for synthesising large quantities of realistic mechanical artefacts. Our extensive human validation of the synthetic artefact model suggests that even experts cannot reliably distinguish between our synthetic damage models and real damage at any level of zoom. Furthermore, we have demonstrated that training with damage generated by our model leads to a significant improvement in both artefact detection and end-to-end restoration tasks when tested on authentic damaged images. We have also published a high-quality baseline dataset to drive progress in restoration research. 

Overall, we conclude that there remains significant work to achieve acceptable automatic restoration quality. We see the use of high-quality damage simulators, multi-scale approaches that can incorporate wide image context while operating at high resolution, and evaluation with challenging full-resolution image benchmarks as important directions to achieve these goals.

\section*{Acknowledgements}
This work was supported by the Engineering and Physical Sciences Research Council [grant number EP/R513222/1].

\bibliographystyle{eg-alpha-doi} 
\input{body.bbl}

\newpage
\onecolumn
\appendix

\end{document}

%% file: body.bbl
\newcommand{\etalchar}[1]{$^{#1}$}

%% file: hcdbnqjtcjmvfxrwcqhvxzfkhynqrfnr 2/body.bbl
\newcommand{\etalchar}[1]{$^{#1}$}
\begin{thebibliography}{\uppercase{WYW{\etalchar{*}}18}}

\bibitem[ADMG18]{achlioptas2018learning}
\textsc{Achlioptas P., Diamanti O., Mitliagkas I., Guibas L.}:
\newblock Learning representations and generative models for 3d point clouds.
\newblock In \emph{International Conference on Machine Learning} (2018), PMLR,
  pp.~40--49.

\bibitem[BKC17]{badrinarayanan2017segnet}
\textsc{Badrinarayanan V., Kendall A., Cipolla R.}:
\newblock Segnet: A deep convolutional encoder-decoder architecture for image
  segmentation.
\newblock \emph{IEEE Transactions on Pattern Analysis and Machine Intelligence
  39}, 12 (2017), 2481--2495.

\bibitem[BZI22]{bakhtiarnia2022efficient}
\textsc{Bakhtiarnia A., Zhang Q., Iosifidis A.}:
\newblock Efficient high-resolution deep learning: A survey.
\newblock \emph{arXiv preprint arXiv:2207.13050} (2022).

\bibitem[CCZS22]{chen2022simple}
\textsc{Chen L., Chu X., Zhang X., Sun J.}:
\newblock Simple baselines for image restoration.
\newblock \emph{arXiv preprint arXiv:2204.04676} (2022).

\bibitem[Cha19]{chambah2019}
\textsc{Chambah M.}:
\newblock {Digital film restoration and image quality}.
\newblock In \emph{{ICA-BELGIUM Colour Symposium}} (Ghent, Belgium, 2019).
\newblock URL: \url{https://hal.archives-ouvertes.fr/hal-02998573}.

\bibitem[CSJH05]{chambah2005}
\textsc{Chambah M., Saint-Jean C., Helt F.}:
\newblock Image quality evaluation in the field of digital film restoration.
\newblock \emph{Proceedings of SPIE - The International Society for Optical
  Engineering 5668} (01 2005).
\newblock \href {https://doi.org/10.1117/12.586738}
  {\path{doi:10.1117/12.586738}}.

\bibitem[CSJHR06]{chambah2006}
\textsc{Chambah M., Saint-Jean C., Helt F., Rizzi A.}:
\newblock {Further image quality assessment in digital film restoration}.
\newblock In \emph{Image Quality and System Performance III} (2006), Cui L.~C.,
  Miyake Y., (Eds.), vol.~6059, International Society for Optics and Photonics,
  SPIE, p.~60590S.
\newblock URL: \url{https://doi.org/10.1117/12.642414}, \href
  {https://doi.org/10.1117/12.642414} {\path{doi:10.1117/12.642414}}.

\bibitem[CSL{\etalchar{*}}17]{cai17tvc}
\textsc{Cai N., Su Z., Lin Z., Wang H., Yang Z., Ling B. W.-K.}:
\newblock Blind inpainting using the fully convolutional neural network.
\newblock \emph{The Visual Computer 33} (2017), 249--261.

\bibitem[FSG17]{fan2017point}
\textsc{Fan H., Su H., Guibas L.~J.}:
\newblock A point set generation network for 3d object reconstruction from a
  single image.
\newblock In \emph{Proceedings of the IEEE Conference on Computer Vision and
  Pattern Recognition} (2017), pp.~605--613.

\bibitem[Gan08]{gann-film-08}
\textsc{Gann R.~G.}:
\newblock Image scanner and method for detecting a defect in an image to be
  scanned, Apr. 2008.
\newblock (Google Patents Entry)
  \httpsAddr{//patents.google.com/patent/US7355159B2/}.

\bibitem[HFH{\etalchar{*}}19]{hertz2019blind}
\textsc{Hertz A., Fogel S., Hanocka R., Giryes R., Cohen-Or D.}:
\newblock Blind visual motif removal from a single image.
\newblock In \emph{Proceedings of the IEEE/CVF Conference on Computer Vision
  and Pattern Recognition} (2019), pp.~6858--6867.

\bibitem[HRCE18]{ham18}
\textsc{Ham C., Raj A., Cartillier V., Essa I.}:
\newblock Variational image inpainting.
\newblock In \emph{NeurIPS 2018 Workshops} (2018).

\bibitem[ISS19]{iizuka2019deepremaster}
\textsc{Iizuka S., Simo-Serra E.}:
\newblock {DeepRemaster: Temporal Source-Reference Attention Networks for
  Comprehensive Video Enhancement}.
\newblock \emph{ACM Transactions on Graphics (Proc. of SIGGRAPH ASIA) 38}, 6
  (2019), 1--13.

\bibitem[ISW22]{visapp22}
\textsc{Ivanova. D., Siebert. J., Williamson. J.}:
\newblock Perceptual loss based approach for analogue film restoration.
\newblock In \emph{Proceedings of the 17th International Joint Conference on
  Computer Vision, Imaging and Computer Graphics Theory and Applications -
  Volume 4: VISAPP,} (2022), INSTICC, SciTePress, pp.~126--135.
\newblock \href {https://doi.org/10.5220/0010829300003124}
  {\path{doi:10.5220/0010829300003124}}.

\bibitem[IZZE17]{isola2017image}
\textsc{Isola P., Zhu J.-Y., Zhou T., Efros A.~A.}:
\newblock Image-to-image translation with conditional adversarial networks.
\newblock In \emph{Proceedings of the IEEE conference on computer vision and
  pattern recognition} (2017), pp.~1125--1134.

\bibitem[JAFF16]{johnson2016perceptual}
\textsc{Johnson J., Alahi A., Fei-Fei L.}:
\newblock Perceptual losses for real-time style transfer and super-resolution.
\newblock In \emph{European Conference on Computer Vision} (2016), Springer,
  pp.~694--711.

\bibitem[KWK21]{kumar2021colorization}
\textsc{Kumar M., Weissenborn D., Kalchbrenner N.}:
\newblock Colorization transformer.
\newblock \emph{arXiv preprint arXiv:2102.04432} (2021).

\bibitem[LCS{\etalchar{*}}21]{liang2021swinir}
\textsc{Liang J., Cao J., Sun G., Zhang K., Van~Gool L., Timofte R.}:
\newblock Swinir: Image restoration using swin transformer.
\newblock In \emph{Proceedings of the IEEE/CVF International Conference on
  Computer Vision} (2021), pp.~1833--1844.

\bibitem[LDR{\etalchar{*}}22]{lugmayr2022repaint}
\textsc{Lugmayr A., Danelljan M., Romero A., Yu F., Timofte R., Van~Gool L.}:
\newblock Repaint: Inpainting using denoising diffusion probabilistic models.
\newblock In \emph{Proceedings of the IEEE/CVF Conference on Computer Vision
  and Pattern Recognition} (2022), pp.~11461--11471.

\bibitem[LMH{\etalchar{*}}18]{pmlr-v80-lehtinen18a}
\textsc{Lehtinen J., Munkberg J., Hasselgren J., Laine S., Karras T., Aittala
  M., Aila T.}:
\newblock {N}oise2{N}oise: Learning image restoration without clean data.
\newblock In \emph{Proceedings of the 35th International Conference on Machine
  Learning} (10--15 Jul 2018), Dy J., Krause A., (Eds.), vol.~80 of
  \emph{Proceedings of Machine Learning Research}, PMLR, pp.~2965--2974.
\newblock URL: \url{https://proceedings.mlr.press/v80/lehtinen18a.html}.

\bibitem[LRS{\etalchar{*}}18]{liu2018image}
\textsc{Liu G., Reda F.~A., Shih K.~J., Wang T.-C., Tao A., Catanzaro B.}:
\newblock Image inpainting for irregular holes using partial convolutions.
\newblock In \emph{Proceedings of the European conference on computer vision
  (ECCV)} (2018), pp.~85--100.

\bibitem[MBP{\etalchar{*}}22]{minaee-segmentation-review-22}
\textsc{Minaee S., Boykov Y., Porikli F., Plaza A., Kehtarnavaz N., Terzopoulos
  D.}:
\newblock Image segmentation using deep learning: A survey.
\newblock \emph{IEEE Transactions on Pattern Analysis \& Machine Intelligence
  44}, 07 (jul 2022), 3523--3542.
\newblock \href {https://doi.org/10.1109/TPAMI.2021.3059968}
  {\path{doi:10.1109/TPAMI.2021.3059968}}.

\bibitem[Mir20]{mironicua2020generative}
\textsc{Mironic{\u{a}} I.}:
\newblock A generative adversarial approach with residual learning for dust and
  scratches artifacts removal.
\newblock In \emph{Proceedings of the 2nd Workshop on Structuring and
  Understanding of Multimedia heritAge Contents} (2020), pp.~15--22.

\bibitem[Per85]{perlin}
\textsc{Perlin K.}:
\newblock An image synthesizer.
\newblock In \emph{Proceedings of the 12th Annual Conference on Computer
  Graphics and Interactive Techniques} (New York, NY, USA, 1985), SIGGRAPH '85,
  Association for Computing Machinery, p.~287–296.
\newblock URL: \url{https://doi.org/10.1145/325334.325247}, \href
  {https://doi.org/10.1145/325334.325247} {\path{doi:10.1145/325334.325247}}.

\bibitem[PLXL21]{peng2021generating}
\textsc{Peng J., Liu D., Xu S., Li H.}:
\newblock Generating diverse structure for image inpainting with hierarchical
  vq-vae.
\newblock In \emph{Proceedings of the IEEE/CVF Conference on Computer Vision
  and Pattern Recognition} (2021), pp.~10775--10784.

\bibitem[PW20]{pielawski2020introducing}
\textsc{Pielawski N., W{\"a}hlby C.}:
\newblock Introducing hann windows for reducing edge-effects in patch-based
  image segmentation.
\newblock \emph{PloS one 15}, 3 (2020), e0229839.

\bibitem[RBL{\etalchar{*}}21]{rombach2021highresolution}
\textsc{Rombach R., Blattmann A., Lorenz D., Esser P., Ommer B.}:
\newblock High-resolution image synthesis with latent diffusion models, 2021.
\newblock \href {http://arxiv.org/abs/2112.10752} {\path{arXiv:2112.10752}}.

\bibitem[RFB15]{ronneberger2015u}
\textsc{Ronneberger O., Fischer P., Brox T.}:
\newblock U-net: Convolutional networks for biomedical image segmentation.
\newblock In \emph{International Conference on Medical image computing and
  computer-assisted intervention} (2015), Springer, pp.~234--241.

\bibitem[RZH{\etalchar{*}}19]{ren2019progressive}
\textsc{Ren D., Zuo W., Hu Q., Zhu P., Meng D.}:
\newblock Progressive image deraining networks: A better and simpler baseline.
\newblock In \emph{Proceedings of the IEEE/CVF Conference on Computer Vision
  and Pattern Recognition} (2019), pp.~3937--3946.

\bibitem[SBB{\etalchar{*}}99]{stavely-film-99}
\textsc{Stavely D.~J., Bloom D.~M., Battles A.~E., Campbell D.~K., E. O.
  R.~H.}:
\newblock Film scanner with dust and scratch correction by use of dark-field
  illumination, Oct. 1999.
\newblock (Google Patents Entry)
  \httpsAddr{//patents.google.com/patent/US5969372A/}.

\bibitem[SCC{\etalchar{*}}22]{saharia2022palette}
\textsc{Saharia C., Chan W., Chang H., Lee C., Ho J., Salimans T., Fleet D.,
  Norouzi M.}:
\newblock Palette: Image-to-image diffusion models.
\newblock In \emph{ACM SIGGRAPH 2022 Conference Proceedings} (2022), pp.~1--10.

\bibitem[SLM{\etalchar{*}}22]{suvorov2022resolution}
\textsc{Suvorov R., Logacheva E., Mashikhin A., Remizova A., Ashukha A.,
  Silvestrov A., Kong N., Goka H., Park K., Lempitsky V.}:
\newblock Resolution-robust large mask inpainting with fourier convolutions.
\newblock In \emph{Proceedings of the IEEE/CVF Winter Conference on
  Applications of Computer Vision} (2022), pp.~2149--2159.

\bibitem[SMF19]{strubel:hal-02369128}
\textsc{Strubel D., Marc B., Fofi D.}:
\newblock {Deep learning approach for artefacts correction on photographic
  films}.
\newblock In \emph{{Fourteenth International Conference on Quality Control by
  Artificial Vision}} (Mulhouse, France, May 2019), Cudel, Bazeille C., Verrier
  S., N, (Eds.), vol.~11172 of \emph{Proceedings of SPIE}, {SPIE-INT SOC
  OPTICAL ENGINEERING, 1000 20TH ST, PO BOX 10, BELLINGHAM, WA 98227-0010 USA},
  p.~35.
\newblock URL:
  \url{https://hal-univ-bourgogne.archives-ouvertes.fr/hal-02369128}, \href
  {https://doi.org/10.1117/12.2521421} {\path{doi:10.1117/12.2521421}}.

\bibitem[SMW85]{sanz-85}
\textsc{Sanz J. L.~C., Merkle F., Wong K.~Y.}:
\newblock Automated digital visual inspection with dark-field microscopy.
\newblock \emph{J. Opt. Soc. Am. A 2}, 11 (Nov 1985), 1857--1862.
\newblock URL:
  \url{http://opg.optica.org/josaa/abstract.cfm?URI=josaa-2-11-1857}, \href
  {https://doi.org/10.1364/JOSAA.2.001857} {\path{doi:10.1364/JOSAA.2.001857}}.

\bibitem[{The}77]{documerica}
\textsc{{The US National Archives}}:
\newblock Documerica project, 1971-1977.
\newblock (Digitised Flickr Collection),
  \url{https://www.flickr.com/photos/usnationalarchives/collections/72157620729903309/}.

\bibitem[TM21]{talebi2021learning}
\textsc{Talebi H., Milanfar P.}:
\newblock Learning to resize images for computer vision tasks.
\newblock In \emph{Proceedings of the IEEE/CVF International Conference on
  Computer Vision} (2021), pp.~497--506.

\bibitem[WBSS04]{ssim}
\textsc{Wang Z., Bovik A., Sheikh H., Simoncelli E.}:
\newblock Image quality assessment: from error visibility to structural
  similarity.
\newblock \emph{IEEE Transactions on Image Processing 13}, 4 (2004), 600--612.
\newblock \href {https://doi.org/10.1109/TIP.2003.819861}
  {\path{doi:10.1109/TIP.2003.819861}}.

\bibitem[WCTJ20]{wang20eccv}
\textsc{Wang Y., Chen Y.-C., Tao X., Jia J.}:
\newblock {VCNet}: A robust approach to blind image inpainting.
\newblock In \emph{European Conference on Computer Vision (ECCV)} (2020).

\bibitem[WYW{\etalchar{*}}18]{wang2018esrgan}
\textsc{Wang X., Yu K., Wu S., Gu J., Liu Y., Dong C., Qiao Y., Change~Loy C.}:
\newblock Esrgan: Enhanced super-resolution generative adversarial networks.
\newblock In \emph{Proceedings of the European conference on computer vision
  (ECCV) workshops} (2018).

\bibitem[WZC{\etalchar{*}}20]{wan2020bringing}
\textsc{Wan Z., Zhang B., Chen D., Zhang P., Chen D., Liao J., Wen F.}:
\newblock Bringing old photos back to life.
\newblock In \emph{proceedings of the IEEE/CVF conference on computer vision
  and pattern recognition} (2020), pp.~2747--2757.

\bibitem[WZCL22]{wan2022bringing}
\textsc{Wan Z., Zhang B., Chen D., Liao J.}:
\newblock Bringing old films back to life.
\newblock In \emph{Proceedings of the IEEE/CVF Conference on Computer Vision
  and Pattern Recognition} (2022), pp.~17694--17703.

\bibitem[YDLL18]{yu2018crafting}
\textsc{Yu K., Dong C., Lin L., Loy C.~C.}:
\newblock Crafting a toolchain for image restoration by deep reinforcement
  learning.
\newblock In \emph{Proceedings of the IEEE conference on computer vision and
  pattern recognition} (2018), pp.~2443--2452.

\bibitem[YLY{\etalchar{*}}18]{yu2018generative}
\textsc{Yu J., Lin Z., Yang J., Shen X., Lu X., Huang T.~S.}:
\newblock Generative image inpainting with contextual attention.
\newblock In \emph{Proceedings of the IEEE conference on computer vision and
  pattern recognition} (2018), pp.~5505--5514.

\bibitem[ZAK{\etalchar{*}}21]{zamir2021multi}
\textsc{Zamir S.~W., Arora A., Khan S., Hayat M., Khan F.~S., Yang M.-H., Shao
  L.}:
\newblock Multi-stage progressive image restoration.
\newblock In \emph{Proceedings of the IEEE/CVF conference on computer vision
  and pattern recognition} (2021), pp.~14821--14831.

\bibitem[ZAK{\etalchar{*}}22]{Zamir_2022_CVPR}
\textsc{Zamir S.~W., Arora A., Khan S., Hayat M., Khan F.~S., Yang M.-H.}:
\newblock Restormer: Efficient transformer for high-resolution image
  restoration.
\newblock In \emph{Proceedings of the IEEE/CVF Conference on Computer Vision
  and Pattern Recognition (CVPR)} (June 2022), pp.~5728--5739.

\bibitem[ZIE16]{zhang2016colorful}
\textsc{Zhang R., Isola P., Efros A.~A.}:
\newblock Colorful image colorization.
\newblock In \emph{European conference on computer vision} (2016), Springer,
  pp.~649--666.

\bibitem[ZIE{\etalchar{*}}18]{zhang18cvpr}
\textsc{Zhang R., Isola P., Efros A.~A., Shechtman E., Wang O.}:
\newblock The unreasonable effectiveness of deep features as a perceptual
  metric.
\newblock In \emph{CVPR} (2018).

\end{thebibliography}
